\documentclass[runningheads]{llncs}
\usepackage[T1]{fontenc}
\usepackage{graphicx}
\usepackage{booktabs}
\usepackage[misc]{ifsym}

\usepackage{mwe}
\usepackage{amsmath}
\usepackage{xcolor}
\usepackage{bm}

\usepackage{subfig}
\begin{document}

\title{EDN: A Novel Edge-Dependent Noise Model \\ for Graph Data}

\titlerunning{EDN: A Novel Edge-Dependent Noise Model for Graph Data}

\author{Pintu Kumar\inst{1} \Letter \orcidID{0000-0001-6111-1458} \and
Nandyala Hemachandra\inst{1} \orcidID{0000-0003-2917-1551
}
}
\tocauthor{Pintu Kumar, Nandyala Hemachandra}
\toctitle{EDN: A Novel Edge-Dependent Noise Model for Graph Data}

\authorrunning{P. Kumar and N. Hemachandra}

\institute{Indian Institute of Technology Bombay,  India \email{\{pintuk,nh\}@iitb.ac.in}
}

\maketitle              
\footnote{This work has been accepted for research track at \textit{European Conference on Machine Learning and Principles and Practice of Knowledge Discovery in Databases} (ECML PKDD 2025).}

\begin{abstract}
An important structural feature of a graph is its set of edges, as it captures the relationships among the nodes (the graph's topology). Existing node label noise models like Symmetric Label Noise (SLN) and Class Conditional Noise (CCN) disregard this important node relationship in graph data; and the Edge-Dependent Noise (EDN) model addresses this limitation. EDN posits that in real-world scenarios, label noise may be influenced by the connections between nodes. We explore three variants of EDN.  A crucial notion that relates nodes and edges in a graph is the degree of a node; we show that in all three variants, the probability of a node's label corruption is dependent on its degree. Additionally, we compare the dependence of these probabilities on node degree across different variants. We performed experiments on popular graph datasets using 5 different GNN architectures and 8 noise robust algorithms for graph data. The results demonstrate that 2 variants of EDN lead to greater performance degradation in both Graph Neural Networks (GNNs) and existing noise-robust algorithms, as compared to traditional node label noise models. We statistically verify this by posing a suitable hypothesis-testing problem. This emphasizes the importance of incorporating EDN when evaluating noise robust algorithms for graphs, to enhance the reliability of graph-based learning in noisy environments. Link to code: \url{https://github.com/pintu-dot/edn}

\end{abstract}

\section{Introduction}
Graph Neural Networks (GNNs) have shown good performance on the graph node classification task \cite{Xiao2021GraphNN,Zhou2018GraphNN}. GNNs assume that the available labels for training data are clean and noise-free, which may not be the case when working with real-world data \cite{Ju2024ASO}. Labels of real-world data are prone to noise, and noise can creep into data for many reasons, like expensive labelling, lack of expertise, human weariness, erroneous devices, adversaries changing labels, insufficient information
 to provide labels, etc \cite{tut_nh,Ju2024ASO}. Hence, effectively learning for graph data in the presence of label noise has gained attention from the community 
\cite{Yuan2023LearningOGcgnn,NT2019LearningGN,Du2021NoiserobustGL,Zhu2024RobustNC,dai2021nrgnn,qian2023robust,Li2024ContrastiveLOcrgnn}.

 One of the main reasons behind GNN's superior performance compared to traditional multilayer perceptron is that GNNs incorporate structural information during learning. Structural information is an integral part of graph data. However, all current work on noise-robust graph learning uses one of the following: 1. Symmetric Label noise, 2. Pairwise noise / Class-Conditional noise, 3. Instance-Dependent noise. These noise models were originally proposed for i.i.d. data and not for graph data, and hence, they assume that label noise is independent of the structure of the node.

Consider a graph where nodes represent users in an online discussion forum. An edge between two nodes captures the interaction between users on a platform, such as a reply, comment, or question-answer exchange. Every user is assigned one of two labels \emph{\{helpful, not helpful\}}. In such a graph, label noise can creep in two scenarios: \textbf{1.} Two helpful users who interacted with each other disagreed with each other or, due to some misunderstanding, couldn't convey their point of view, and hence, labelled each other as \textit{not helpful}; \textbf{2.} Similarly, two unhelpful users might get incorrectly labelled as \textit{helpful} because of a rare good discussion or due to colluding. In such graphs, noise dependent on just one node is not useful; rather, noise should be passed through edges. The labels of nodes on both sides of the edge should be changed together. We refer to this approach of adding noise to node labels as an edge-dependent noise model (EDN). In this work, we study three variants of edge-dependent noise and their impact on learnability. 

The main contributions of the paper are as follows: \textbf{1.} Propose three variants of EDN, in which noise is passed through edges. \textbf{2.} For these noise models, we derive closed-form expressions for the probabilities of label change in terms of node degree. 
\textbf{3.}  We analytically compare these probabilities as a function of node degree. \textbf{4.} We perform detailed experiments to check the behaviour of existing GNN architectures and noise-robust graph learning algorithms in the presence of EDN, using 5 different GNN architectures and 8 noise-robust graph learning algorithms. \textbf{5.} Based on confidence intervals of test accuracies for various noise models on many datasets, we observe that two variants of EDN at many noise levels substantially degrade the performance as compared to the existing noise label models. \textbf{6.} We pose this as a suitable hypothesis test problem to statistically verify our observations, and we conclude the same.

\section{Related Work - Existing Node Label Noise Models}
Let $\mathcal{G=(V,E)}$ be a graph, where $\mathcal{V}$ denotes the set of vertices and $\mathcal{E\subseteq V\times V}$ denotes the set of edges. Each node $v_i \in \mathcal{V}$, have an associated label $y_i \in \{1,2,\cdots,K\}$. In models for label flipping, we can associate a discrete-time Markov chain  \cite{Norris_1997} on the state space of labels $\{1,2, \cdots, K\}$ in which each flip in the label corresponds to a state transition of the Markov Chain. We give details of the existing methods for adding noise to node labels; in each case, we identify the transition probability matrix of the associated Markov chain.

\subsection{Symmetric Label Noise}
Symmetric Label Noise (SLN) \cite{wang2024noisygl,tut_nh} assumes that the label of a node is changed with some fixed probability $\rho$ (and hence retained with probability $1-\rho$). Also, the probability of a label being reassigned to each of the other classes is the same, which is $\rho / (K-1)$. Mathematically, if $y$ and $y'$ denote true and noisy label respectively, then $P(y'=n|y=m)=\frac{\rho}{K-1}$, where $n,m \in \{1,2,\ldots,K\}$ and $m\neq n$. Transition probability matrix for SLN $(Q_{sln})$ is given by

\begin{equation}
Q_{sln} =
\begin{bmatrix}
1- \rho & \frac{\rho}{K-1} & \frac{\rho}{K-1} & \ldots & \frac{\rho}{K-1} \\ 
\frac{\rho}{K-1} & 1-\rho & \frac{\rho}{K-1} & \ldots & \frac{\rho}{K-1} \\ 
\vdots & \ddots & \ddots & \ddots & \vdots \\ 
\frac{\rho}{K-1} & \ldots & \ddots & 1-\rho & \frac{\rho}{K-1} \\ 
\frac{\rho}{K-1} & \frac{\rho}{K-1} & \ldots & \frac{\rho}{K-1} & 1-\rho
\end{bmatrix}
\label{eq:sln_matrix}
\end{equation}

\subsection{Class Conditional Noise}
In Class Conditional Noise (CCN) \cite{wang2024noisygl,tut_nh}, the probability with which the label is changed depends on both $y$ and $y'$. The probability of a node of class $m$ being reassigned to class $n$ is given by $\rho_{mn}$ ($P(y'=n|y=m)=\rho_{mn}$), where $m\neq n$. So, a node with label $m$ is flipped with probability $\rho_m=\sum_{i=1,i\neq m}^K\rho_{mi}$ and the label is retained with the probability $1-\rho_m$. The transition probability matrix $(Q_{ccn})$ is given by 

$$
Q_{ccn} = 
\begin{bmatrix}
1- \rho_1\ \ & \rho_{12} &\ \ \rho_{13}  & \ldots & \ \ \ \rho_{1K} \\ 
 \rho_{21}& 1-\rho_2 \ \  & \ \ \rho_{23}  \ \ &  & \ \ \ \rho_{2K}\\ 
\vdots &  & \ \ \ddots & \ \ \ddots &\ \ \  \vdots\\ 
 \rho_{(K-1)1} &  &  & \ \ \ 1-\rho_{K-1} & \ \ \ \rho_{(K-1)K} \\ 
\rho_{K1}  &\rho_{K2}   & \dots &\ \ \ \rho_{K(K-1)}   &\ \ \ 1- \rho_K 
\end{bmatrix}
$$

\subsubsection{Pairwise Noise:}
Pairwise Noise (PWN) \cite{wang2024noisygl} is a special class of CCN. The motivation behind Pairwise Noise is that one is more likely to mislabel two similar classes. For Pairwise Noise $\rho_1=\rho_2=\ldots=\rho_K=\rho$,  and the label is flipped to the next label (with probability $\rho$). The transition probability matrix $Q_{pwn}$ is given by 

\begin{equation}
Q_{pwn}=
\begin{bmatrix}
1- \rho\ \ & \rho &\ \ 0  & \ldots & \ \ \ 0 \\ 
 0& 1-\rho \ \  & \ \ \rho \ \ &  & \ \ \ 0\\ 
\vdots &  & \ \ \ddots & \ \ \ddots &\ \ \  \vdots\\ 
 0&  &  & \ \ \ 1-\rho & \ \ \ \rho\\ 
\rho &0  & \dots &\ \ \ 0  &\ \ \ 1- \rho 
\end{bmatrix}
\label{eqn:pair_matrix}
\end{equation}

Many label noise robust algorithms for graphs have been proposed to tackle existing node label noise. DGNN \cite{NT2019LearningGN} employs backward loss correction. PIGNN \cite{Du2021NoiserobustGL} leverages pairwise interactions (PI) between nodes for noise-resistant learning. RNCGLN \cite{Zhu2024RobustNC} uses pseudo-labeling within a self-training framework to correct noisy labels. NRGNN \cite{dai2021nrgnn} connects unlabeled nodes with high feature similarity to labelled nodes for better pseudo-labelling. RTGNN \cite{qian2023robust} enhances information flow by bridging labelled and unlabeled nodes while employing dual GNNs for noise mitigation. CRGNN \cite{Li2024ContrastiveLOcrgnn} combines contrastive learning and dynamic cross-entropy loss to encourage robust feature learning. CGNN \cite{Yuan2023LearningOGcgnn} integrates graph contrastive learning and a sample selection strategy based on the homophily assumption to filter noisy labels. DeGLIF \cite{DeGLIf} uses the influence function to identify and relabel noisy nodes in the graph. In our work, we would check how these algorithms perform in the presence of structure-dependent noise like EDN.

Some works, such as \cite{Chen2024ADEdgeDropAE,Rong2019DropEdgeTD}, introduce structural noise into the graph by dropping edges. In contrast, our work focuses on node label noise propagated through the edges, rather than modifying the graph structure itself. Next, \cite{Liu2019AUF} and \cite{Zhang2020AdversarialLA} explore structurally motivated adversarial or noisy label settings on graphs. However, they primarily focus on label flips and structural changes that mislead GNN's training, whereas our EDN model introduces a fundamentally different approach where label noise arises due to edge connectivity, making it structure-dependent and degree-sensitive.

\section{ A Novel Edge-Dependent Noise Model (EDN)}  

Assume $\mathcal{G=(V,E)}$, is an undirected graph, having \textit{$m$ nodes}. $\mathcal{X}=\{x_1,\ldots,x_m\}$ and $\mathcal{Y}=\{y_1,\ldots ,y_m\}$  are the set of feature vectors and the set of true labels associated with corresponding nodes, respectively. We propose a node-label noise model for graphs (called edge-dependent noise model (EDN)) where the noisy labels of connected nodes are correlated as these noisy labels depend on the edge connecting them. Similar to existing noise models, edge-dependent noise models inject noise into the graph in two steps: 1. Selecting nodes whose labels should be changed, 2. Deciding new labels for selected nodes. What differentiates EDN from existing label noise models is that in EDN, the selection of nodes for label change depends on their structural information.

\subsection{Selecting nodes to change their labels} In the proposed EDN model, we sample each edge with \textbf{fixed probability} $\bm{\rho}$; these sampled edges are called noisy edges. 
These noisy edges suggest that the labels of nodes on both sides of the noisy edge should be changed. For a node with degree 1, the label is changed if the edge incident to that node is noisy. For nodes with a degree $n>1$, incident edges may have conflicting opinions on changing the label of the node. Based on how these opinions are aggregated, we have three variants:  

\subsubsection{Majority vote (MV):} The label of a node $v$ is changed if more than or equal to half of the edges incident to $v$ are noisy. If the degree of $v$ is $deg(v)$, the probability of a node getting selected to change its label is hence given by $q(deg(v),\rho)$
\begin{equation}
\label{q}
    q(deg(v),\rho)=\sum_{i=\lceil \frac{deg(v)}{2}\rceil}^{deg(v)} {deg(v) \choose i}\rho^i(1-\rho)^{deg(v)-i}.\end{equation}

\subsubsection{Veto power (Veto):} The label of a node $v$ is changed if at least one of the incident edges to $v$ is noisy. The probability of the label of a node $v$ with degree $deg(v)$ getting changed is hence given by  $r(deg(v),\rho)$
  \begin{equation}
  \label{r}
  r(deg(v),\rho)=1-{deg(v) \choose 0}(1-\rho)^{deg(v)}=1-(1-\rho)^{deg(v)}.\end{equation}

\subsubsection{Sequential flipping (seq):} 
  In this variant, the label of a node sequentially evolves. 
For a node $v$, we consider all its \textit{noisy incident edges}. The first noisy incident edge changes the label to a new class. The second noisy edge changes the label from this new class to another (with a possibility of reverting to the original label). This process continues for all noisy edges. The probability with which a node $v$ with degree $deg(v)$ is flipped depends on how new labels are assigned when observing a noisy edge and is discussed in Section \ref{sec:assign}.

\subsection{Assigning noisy labels to selected nodes}
\label{sec:assign}

For Majority vote and Veto power, after selecting nodes whose labels are to be changed, we use SLN and Pairwise noise \cite{wang2024noisygl,tut_nh} to assign a new label. The difference between the existing and the EDN variants is that existing models have the same probability of selecting every node, whereas, in our noise model, the \textit{probability of selecting a node is dependent on the degree of the node.} This is demonstrated by Equation \ref{q},\ref{r},\ref{s_sln} and \ref{s_pwn}. Existing noise models have the same transition probability matrix for all nodes, whereas EDN has different transition probability matrices for nodes with different degrees. In sequential flipping, the assignment of a new label due to a noisy edge follows SLN and Pairwise Noise. Both these sub-variants lead to different probabilities of flipping a node. 

\subsubsection{Sequential flipping + SLN:} Each edge incident to node $v$ is noisy with probability $\rho$. So, due to a single edge, the label of node $v$ is unchanged with probability $1-\rho$, and the label changes to a different class with probability $\rho$.
In the case of sequential flipping $+$ SLN, when a noisy edge alters the label, the new label is selected according to the SLN model. This means that each possible class is equally likely, with a probability of $\frac{\rho}{K-1}$. The transition probability matrix associated with the noise caused by a single edge is given by $Q_{SLN}$ from Equation \ref{eq:sln_matrix}. If $v$ has degree $n$, then the transition probability matrix for Sequential flipping $+$ SLN is $Q_{SLN}^n$ (relabelling n times, every time starting with a new label). $Q_{SLN}$ is a symmetric matrix, using the diagonalization property of symmetric matrix (Spectral theorem) \cite{strang2000linear}, we derive $Q_{SLN}^n$ as follows:

\begin{equation*}
\begin{aligned}
    Q_{SLN}^{n}=\frac{1}{K}\begin{bmatrix}
1\ \ & 1 &\ \ 1  & \ldots & \ \ \ 1 \\ 
 1& 1 \ \  & \ \ 1  \ \ &  & \ \ \ 1\\ 
\vdots &  & \ \ \ddots & \ \ \ddots &\ \ \  \vdots\\ 
 1 &  &  & \ \ \ 1 & \ \ \ 1 \\ 
1  &1   & \dots &\ \ \ 1   &\ \ \ 1 
\end{bmatrix}+ \left(1-\frac{K\rho}{K-1}\right)^{n} 
\begin{bmatrix}
\frac{K-1}{K} & -\frac{1}{K} & -\frac{1}{K}  & \ldots &  -\frac{1}{K} \\ 
 -\frac{1}{K}& \frac{K-1}{K}   &  -\frac{1}{K}  &  & -\frac{1}{K}\\ 
\vdots &  & \ddots &\ddots &  \vdots\\ 
 -\frac{1}{K} &  &  & \frac{K-1}{K} &  -\frac{1}{K} \\ 
-\frac{1}{K}  &-\frac{1}{K}   & \dots & -\frac{1}{K}   & \frac{k-1}{K} 
\end{bmatrix}
\end{aligned}
\end{equation*}
Detailed derivation for $Q_{SLN}^n$ is available in Appendix A. Using $Q_{SLN}^n$, for a node $v$ starting with the true label $y$, the probability of the label being changed to a specific class is given by: 
$$s\_sc(deg(v),\rho)=\frac{1}{K} \left( 1-\left(1-\frac{K\rho}{K-1}\right )^{deg(v)}\right).$$ 

For the node $v$, when using Sequential flipping + SLN model, the probability of its label being flipped is $(K-1)\times s\_sc(n)$ and is hence given by 
\begin{equation}
\label{s_sln}
s_{sln}(deg(v),\rho)=\frac{K-1}{K} \left( 1-\left(1-\frac{K\rho}{K-1}\right )^{deg(v)}\right).\end{equation}

\subsubsection{Sequential flipping + PWN:} In sequential flipping with PWN, label reassignment due to a single edge follows the pairwise noise model. The corresponding transition probability matrix is given by $Q_{pwn}$ from Equation \ref{eqn:pair_matrix}. If $v$ has degree $n$, then the transition probability matrix for Sequential flipping with the pairwise noise is $Q_{pwn}^n$ (relabelling $n$ times, every time starting with a new label). Observe that in $Q_{pwn}$, each row is a rightward cyclic shift of the previous row, which makes it a circulant matrix \cite{CircDavis,strang2000linear}. We use the eigendecomposition of the circulant matrix \cite{CircDavis,strang2000linear} to obtain $Q_{pwn}^n$. Since $Q_{pwn}^n$ is a product of circulant matrices, it remains circulant \cite{CircDavis,strang2000linear}, meaning the entire matrix is characterised by its first row. The first row of $Q_{pwn}^n$ is given by:
$$Q_{pwn}^n[0,j]=\sum_{m=0}^n{n\choose m} \rho^m  (1-\rho)^{n-m} \delta_{m-j\mod K}.$$
Detailed derivation for $Q_{pwn}^n$ is in the Appendix B. For the node $v$, when using the Sequential flipping + PWN model, the probability of its label being flipped is
\begin{equation}
\label{s_pwn}
    s_{pwn}(deg(v),\rho)=\sum_{j=1}^{K-1}\sum_{m=1}^{deg(v)}{deg(v)\choose m} \rho^m  (1-\rho)^{deg(v)-m} \delta_{m-j\mod K}
\end{equation}
\begin{figure}[!h]%
    \centering
    \subfloat{{\includegraphics[width=0.4\linewidth]{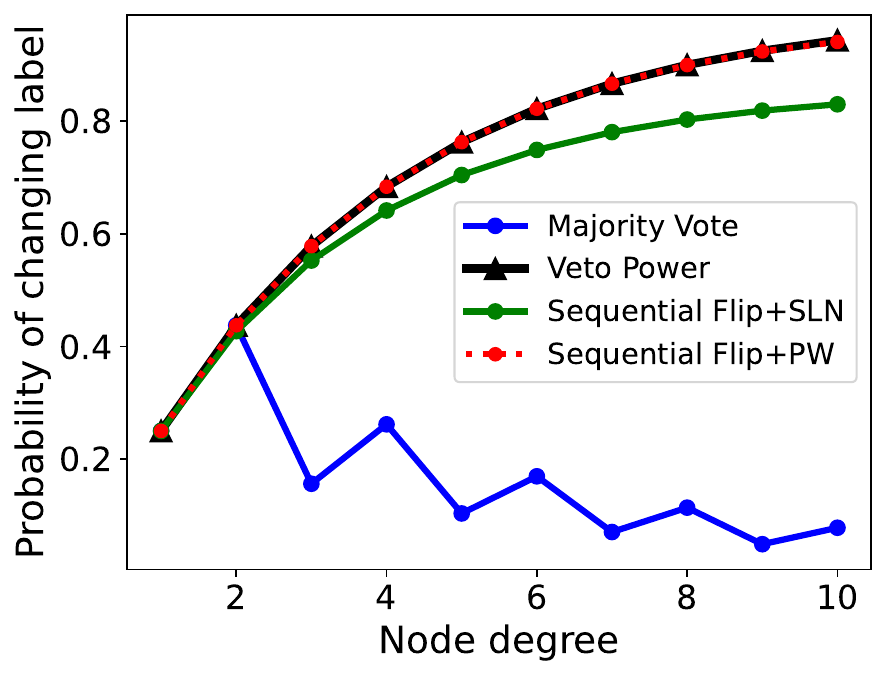} }}%
    \qquad
    \subfloat{{\includegraphics[width=0.4\linewidth]{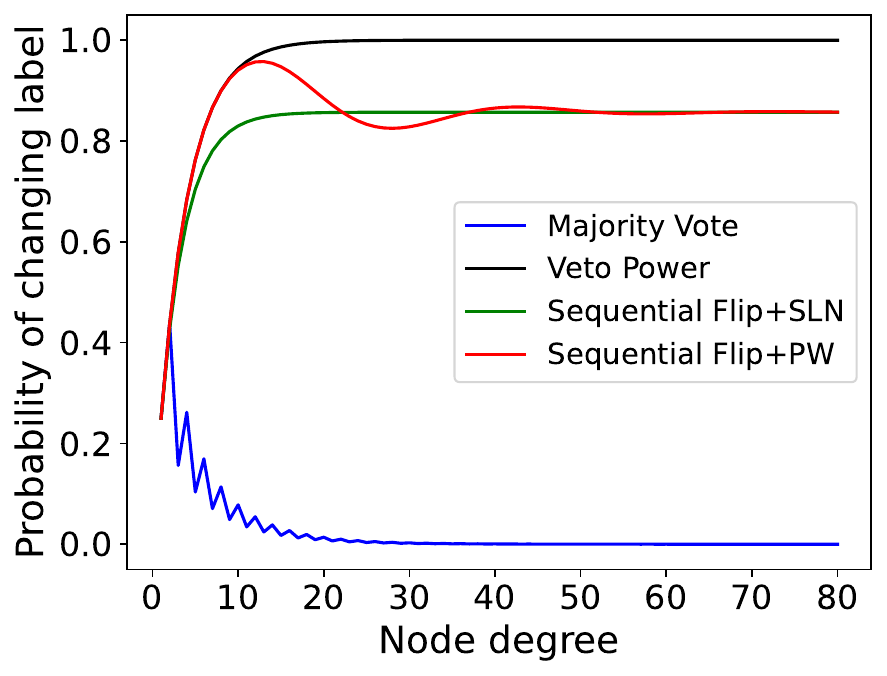} }}%
      \caption{Node degree vs probability of label change for three variants of EDN. We use $\rho=0.25$ (the probability of an edge being noisy) and use $K=7$  for sequential flipping.}
    \label{fig:3var}
\end{figure}

In all three variants, the probability of a node's label changing \textit{depends on its degree.} The relationship between node degree and this probability is illustrated in Fig. \ref{fig:3var}. We observe that for a fixed $\rho$, in the majority vote variant, the probability $q(deg(v),\rho)$ decreases with slight fluctuations as the node degree increases. In the other two variants, both the probabilities $r(deg(v),\rho)$ and $s_{sln}(deg(v),\rho)$ monotonically increase and saturate at $1$ and $\frac{K-1}{K}$ respectively. We summarize this as Theorem \ref{thm:thm1}, with its proof provided in Appendix C:

\begin{theorem}
\label{thm:thm1}
    Three variants of EDN satisfy the following properties:
    \begin{enumerate}
    \item For a fixed $\rho$, $r(deg(v),\rho)$ is an increasing function of $deg(v)$. Also, $s(deg(v),\rho)$ is an increasing function of $deg(v)$ for $ \rho<\frac{K-1}{K}$. 
        \item $r(deg(v),\rho)\geq q(deg(v),\rho) \  \forall ~ deg(v)$ and fixed $\rho$.
        \item If $\rho<\frac{K-1}{K}$, then $s_{sln}(deg(v),\rho)<\frac{K-1}{K}$ and $s_{sln}(n)=\frac{K-1}{K}$ iff $\rho=\frac{K-1}{K}$.
    \end{enumerate}
\end{theorem}

The curve for $s_{pwn}$ initially follows veto power and then oscillates to converge to seq+SLN. For $deg(v)\leq K-1$, we have $s_{pwn}(deg(v),\rho)=r(deg(v),\rho)$. In Figure \ref{fig:3var}, we observe that $s_{pwn}$ forms a cycle of alternating increasing and decreasing phases. For a node $v_i$, without loss of generality, assume that its label $y_i=1$. If $deg(v)=1$, its label gets changed to class 2 with some probability, but never to any other class. If $deg(v)=2$, $v$, moving from $Q_{pwn}$ to $Q_{pwn}^2$, if the node has been already assigned label 2, it can not return to the original label and can either remain in 2 or get reassigned to class 3. If $v_i$ was not assigned label 2, it now has the probability $\rho$ of getting assigned to class 2. This implies $s_{pwn}(2,\rho)>s_{pwn}(1,\rho)$; meaning, $s_{pwn}$ starts with an increasing phase. If $deg(v)=K-1$, then it can be reassigned to any class, and $s_{pwn}$ is an increasing function of node degree up to this point. At $deg(v)=K$, if $v_i$ was reassigned to class $K$ by its first $K-1$ edges, then there is a small probability it gets reassigned to label 1. When this probability exceeds that of class 1 reassigned to class 2 (which occurs at $n=13$ for $K=7$), then it starts to decrease. and the decreasing phase will continue until another cycle is completed again, explaining the alternating pattern.

Recall $Q_{pwn}$ is the transition probability matrix of a discrete-time Markov chain. This Markov chain is aperiodic and has only one communicating class, and  $\pi=[1/K,1/K,\ldots,1/K]$ is the unique stationary distribution for $Q_{pwn}$. Hence, $Q_{pwn}^n$ converges to a matrix $Q^*$, where every row of $Q^*$ equals $\pi$ \cite{Norris_1997}. So, $s_{pwn}(deg(v),\rho)$ converges to $(K-1)\times\frac{1}{K}$, this value is same as the upper bound for $s_{sln}$. We summarize this discussion about $s_{pwn}$ as:

\begin{theorem}
    $s_{pwn}(deg(v),\rho)$, the probability with which the label of node $v$ is changed in presence of Sequential flipping+PWN model, satisfies the following:
    \begin{enumerate}
        \item $s_{pwn}$ forms a cycle of alternating increasing and decreasing phases. It initially increases, followed by a period of decrease, and this pattern continues.
        \item For any fixed $\rho$ we have $s_{pwn}(deg(v),\rho)=r(deg(v),\rho)$, for $deg(v)\leq K-1$.
        \item $s_{pwn}(deg(v),\rho)$ converges to $\frac{K-1}{K}$ as $deg(v) \to \infty$.
    \end{enumerate}
\end{theorem}

\section{Experiments and Results}
\subsection{Datasets and their splits}
We test the impact of EDN on existing GNNs and Noise-robust algorithms. using   Citeseer \cite{Yang2016RevisitingSL}, Cora \cite{Yang2016RevisitingSL}, and Amazon photo \cite{shchur2018pitfalls},  with splits similar to DeGLIF \cite{DeGLIf}. Details about dataset statistics are in Table \ref{tab:dat_stat}. These datasets were selected as they vary in node count, feature dimensions, and average degree. 
\begin{table*}[!h]
\caption{Dataset Statistics}\label{tab:dat_stat}
\centering
\begin{tabular}{ l   l   l   l   l}
\hline
Dataset &  \# Nodes \ \ \ \ & \# Edges \ \ \ \ & Feature dim\ \ \   & \# Classes\\
\hline
CiteSeer & 3,327 & 9,104 & 3,703 & 6 \\
Cora & 2,708 & 10,556 & 1,433 & 7\\
Amazon Photo & 7,650 & 238,162 & 745 & 8\\
\hline
\end{tabular}
\end{table*}

\textbf{Datasets split details:}  We use split similar to 
\cite{DeGLIf}. For the Cora dataset, we use 172 nodes per class for training, 500 nodes for validation, and 1000 nodes for testing. For the Citeseer dataset, we use 250 randomly sampled nodes per class for training, 500 nodes for validation, and 1000 nodes for testing. For the Amazon Photo dataset, we use 54 nodes per class for training, 500 nodes in total for validation, and the rest of the nodes for testing. All datasets have been fetched from the PyTorch Geometric library, with feature normalisation being true.

\subsection{Injecting EDN noise}
The same value of $\rho$ can lead to different levels of noise using different variants of EDN noise models. Also, as variants of EDN are degree-dependent, graphs with different degree distributions can have different noise levels for the same $\rho$. Let $d(n)$ represent the degree distribution of a graph, then the expected noise level in the graph is given by $\sum_{i=1}^{max degree} l(i)d(i)$ where $l$ is $q,r$ or $s$. To make a fair comparison with the existing noise model and among different variants of EDN, we choose different $\rho$ for each variant and each dataset so that the expected noise level in the graph is the same. A ready-to-refer value of $\rho$ for different datasets, corresponding to overall noise levels in the graph ranging from 5\% to 50\% in increments of 5\%, is available in Table 6 of the supplementary material.

\subsection{Experimental Setup}
We add different types of noise to data, where the noise level is between 5\% and 50\% in increments of 5\%. GCN, GraphSAGE, GAT and Graph Transformer have been implemented using Pytorch Geometric using GCNConv, SAGEConv, GATConv and Graph Transformer respectively. For GCN and GraphSAGE we use 1 hidden layer of size 16 and relu activation. For GAT and Graph Transformer, we use 1 hidden layer with 8 heads of size 8. For GIN we use implementation by \cite{wang2024noisygl}. It is worth mentioning that the comparison is not between GNNs but between different noise models, so using slightly different architectures for different GNNs strengthens our experiments. We use implementation by \cite{wang2024noisygl} for all noise-robust algorithms except for DeGLIF for, which we use implementation by \cite{DeGLIf}. Each experiment is repeated 10 times, with mean $\pm$ standard deviation reported. Models are trained on a 24 GB Nvidia RTX 4090 GPU.

\begin{figure}[!ht]%
    \centering
    \subfloat{{\includegraphics[width=0.44\linewidth]{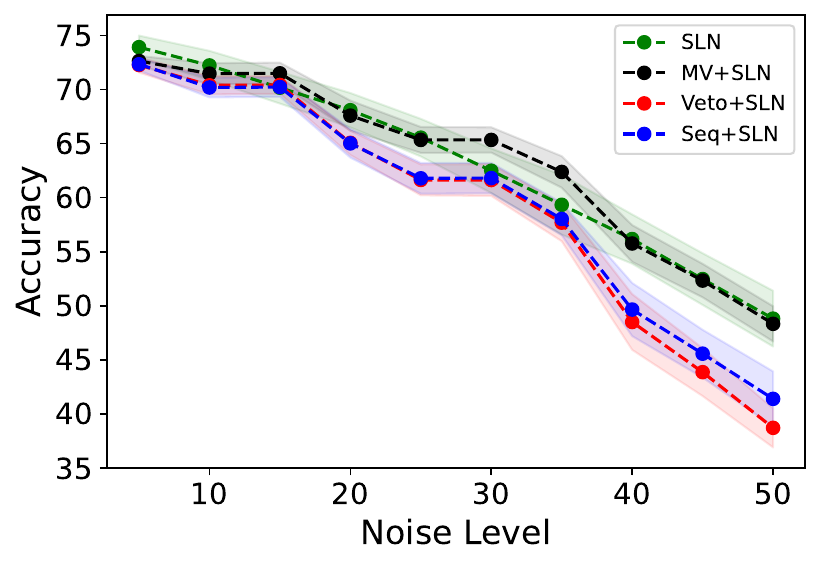} }}%
    \qquad
    \subfloat{{\includegraphics[width=0.44\linewidth]{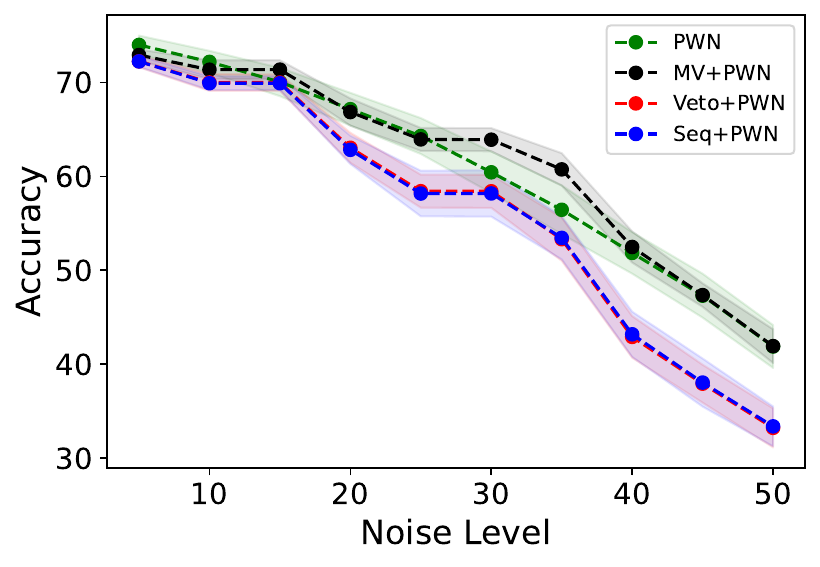} }}%
    
    \caption{ GCN accuracy at same noise level but different noise models on Citeseer dataset. }%
    \label{fig:binary_app}
\end{figure}

\subsection{Computational Results}
\label{subsec:comp_results}
In this section, we attempt to empirically understand the impact of EDN on existing graph learning algorithms. To do so we try to answer following questions:

\subsubsection{Q1. How does GCN architecture perform in the presence of EDN?} 

GCN \cite{kipf2016semi} is one of the widely used GNN architecture, it is also used as a backbone for many noise robust algorithms for graph \cite{dai2021nrgnn,qian2023robust,DeGLIf}. We test GCN under different noise variants of EDN, as well as SLN and PWN. Results for the Citesser dataset are reported in Fig \ref{fig:binary_app}. 
Results for all datasets at $5\%,25\%,$ and $45\%$ noise level is presented in Table \ref{tab:gnn}. Due to the large size of tables and lack of space, results for all datasets at all noise levels are provided in Tables 7,8,9 in the supplementary material. GCN performance in the presence of the Majority vote variant is comparable to existing noise models, SLN and PWN. GCN shows more degradation in performance when injected with Veto Power and Sequential flipping variants of EDN as compared to the existing node label noise model. At low noise levels (less than 15\%) the gap is low. However, at higher noise levels, veto power and sequential flipping gap widen, and they degrade GCN performance the most.

\begin{table}[!ht]
\caption{Comparison of noise model variants across GNN architecture. Reported values are accuracy$\pm$std of 10 repetitions. }
\label{tab:gnn}
\resizebox{\textwidth}{!}{
\begin{tabular}{cr|c|llllllll}
 & \begin{tabular}[c]{@{}c@{}}GNN \\ Architecture\end{tabular} & \multicolumn{1}{c|}{\begin{tabular}[c]{@{}c@{}}Noise \\ Level\end{tabular}} & \multicolumn{1}{l}{SLN\ \ \ \ \ \ \ \ \ \ \ } & \multicolumn{1}{c}{\begin{tabular}[l]{@{}l@{}}MV+ \\ SLN\ \ \ \ \ \ \ \ \ \ \ \end{tabular}} & \multicolumn{1}{c}{\begin{tabular}[l]{@{}l@{}}Veto+\\ SLN \ \ \ \ \ \ \ \ \ \ \end{tabular}} & \multicolumn{1}{c}{\begin{tabular}[l]{@{}l@{}}Seq+\\ SLN\ \ \ \ \ \ \ \ \ \ \end{tabular}} & \multicolumn{1}{l}{PWN\ \ \ \ \ \ \ \ \ \ } & \multicolumn{1}{c}{\begin{tabular}[l]{@{}l@{}}MV+\\ PWN\ \ \ \ \ \ \ \ \ \ \end{tabular}} & \multicolumn{1}{c}{\begin{tabular}[l]{@{}l@{}}Veto+\\ PWN\ \ \ \ \ \ \ \ \ \ \end{tabular}}  &  \multicolumn{1}{c}{\begin{tabular}[l]{@{}l@{}}Seq+\\ PWN\end{tabular}} \\ \hline
 & GCN & 5\% & 73.92$\pm$1.1 & 72.64$\pm$0.5 & 72.27$\pm$0.7 & 72.34$\pm$0.6 & 73.99$\pm$1 & 72.91$\pm$0.6 & 72.24$\pm$0.6 & 72.23$\pm$0.58 \\
 &  & 25\% & 65.55$\pm$1.74 & 65.34$\pm$1.20 & 61.64$\pm$1.43 & 61.79$\pm$1.42 & 64.29$\pm$1.91 & 63.93$\pm$1.22 & 58.42$\pm$1.76 & 58.17$\pm$2.43 \\
 &  & 45\% & 52.47$\pm$2.39 & 52.33$\pm$1.55 & 43.86$\pm$2.21 & 45.57$\pm$2.21 & 47.30$\pm$2.34 & 47.36$\pm$1.25 & 37.93$\pm$2.02 & 38.05$\pm$2.56 \\
C & GIN & 5\% & 71.92$\pm$4.50 & 69.52$\pm$1.93 & 68.59$\pm$2.76 & 67.76$\pm$3.35 & 67.99$\pm$3.27 & 68.56$\pm$3.01 & 68.78$\pm$2.22 & 68.58$\pm$1.95 \\
I &  & 25\% & 60.12$\pm$5.71 & 63.39$\pm$3.58 & 54.51$\pm$4.01 & 55.42$\pm$6.26 & 54.22$\pm$9.29 & 61.47$\pm$3.51 & 49.34$\pm$7.72 & 49.98$\pm$4.16 \\
T &  & 45\% & 42.29$\pm$8.80 & 47.87$\pm$7.42 & 37.98$\pm$5.40 & 39.28$\pm$5.53 & 39.19$\pm$3.55 & 46.07$\pm$6.53 & 34.25$\pm$5.86 & 33.50$\pm$1.65 \\
E & GraphSAGE & 5\% & 75.95$\pm$0.96 & 74.96$\pm$0.51 & 74.69$\pm$0.49 & 74.61$\pm$0.62 & 75.97$\pm$0.88 & 75.07$\pm$0.56 & 74.00$\pm$0.53 & 75.69$\pm$1.30 \\
S &  & 25\% & 70.96$\pm$1.49 & 70.70$\pm$1.07 & 68.42$\pm$1.36 & 68.99$\pm$0.91 & 68.93$\pm$1.92 & 68.33$\pm$1.07 & 63.27$\pm$1.43 & 65.70$\pm$2.16 \\
E &  & 45\% & 59.61$\pm$1.99 & 59.97$\pm$1.79 & 52.89$\pm$2.11 & 54.17$\pm$2.08 & 49.71$\pm$2.85 & 49.19$\pm$1.56 & 41.19$\pm$2.29 & 44.32$\pm$4.10 \\
 E& GAT & 5\% & 76.46$\pm$1.53 & 74.75$\pm$0.67 & 74.33$\pm$0.89 & 74.36$\pm$0.67 & 76.62$\pm$1.49 & 74.93$\pm$0.85 & 74.39$\pm$0.66 & 76.61$\pm$1.39 \\
R &  & 25\% & 74.59$\pm$1.61 & 73.06$\pm$1.04 & 71.81$\pm$1.06 & 72.16$\pm$1.39 & 73.31$\pm$1.84 & 71.32$\pm$1.50 & 68.13$\pm$1.45 & 71.42$\pm$2.64 \\
 &  & 45\% & 70.88$\pm$1.64 & 70.45$\pm$1.81 & 66.60$\pm$2.33 & 67.51$\pm$1.87 & 55.00$\pm$3.75 & 53.93$\pm$2.64 & 43.41$\pm$3.18 & 45.97$\pm$6.50 \\
 & Graph Transformer & 5\% & 76.28$\pm$0.86 & 75.64$\pm$0.52 & 76.28$\pm$0.50 & 76.54$\pm$0.54 & 76.10$\pm$1.40 & 75.50$\pm$0.14 & 76.16$\pm$0.35 & 75.50$\pm$0.95 \\
 &  & 25\% & 70.24$\pm$1.63 & 70.36$\pm$0.60 & 68.84$\pm$0.68 & 68.68$\pm$0.86 & 68.86$\pm$1.03 & 67.48$\pm$1.12 & 64.94$\pm$0.68 & 65.45$\pm$2.00 \\
 &  & 45\% & 58.10$\pm$1.24 & 59.22$\pm$1.42 & 52.38$\pm$0.60 & 53.58$\pm$0.94 & 49.04$\pm$2.25 & 48.94$\pm$2.43 & 43.00$\pm$1.01 & 44.09$\pm$2.77\\
 \hline
 & GCN & 5\% & 84.73$\pm$0.94 & 85.00$\pm$0.44 & 85.19$\pm$0.74 & 85.09$\pm$0.73 & 84.27$\pm$0.98 & 85.36$\pm$0.48 & 85.36$\pm$0.50 & 84.58$\pm$0.54 \\
 &  & 25\% & 76.46$\pm$1.48 & 77.46$\pm$1.99 & 74.49$\pm$2.09 & 74.65$\pm$2.03 & 71.97$\pm$1.77 & 74.84$\pm$2.58 & 68.85$\pm$2.30 & 76.35$\pm$1.44 \\
 &  & 45\% & 62.29$\pm$2.29 & 63.16$\pm$2.94 & 57.12$\pm$2.54 & 58.06$\pm$2.54 & 50.52$\pm$2.76 & 52.86$\pm$3.87 & 44.28$\pm$2.62 & 50.18$\pm$3.92 \\
 & GIN  & 5\% & 80.95$\pm$2.01 & 81.11$\pm$0.88 & 80.45$\pm$2.63 & 82.11$\pm$2.60 & 80.35$\pm$2.38 & 83.34$\pm$1.55 & 80.77$\pm$2.21 & 79.36$\pm$1.99 \\
 &  & 25\% & 79.02$\pm$2.77 & 77.87$\pm$5.65 & 74.90$\pm$2.73 & 76.57$\pm$2.59 & 73.66$\pm$3.74 & 73.32$\pm$4.28 & 64.84$\pm$3.21 & 67.62$\pm$2.51 \\
C &  & 45\% & 71.02$\pm$4.38 & 75.17$\pm$2.36 & 66.79$\pm$9.06 & 65.95$\pm$11.8 & 48.40$\pm$5.01 & 53.90$\pm$4.37 & 42.55$\pm$10.1 & 48.36$\pm$3.33 \\
O & GraphSAGE & 5\% & 83.10$\pm$1.19 & 83.38$\pm$0.68 & 83.37$\pm$1.18 & 83.41$\pm$1.09 & 82.84$\pm$1.31 & 83.56$\pm$0.46 & 83.22$\pm$1.13 & 84.43$\pm$0.81 \\
R &  & 25\% & 70.72$\pm$1.85 & 72.39$\pm$2.37 & 68.83$\pm$1.91 & 69.02$\pm$2.12 & 67.69$\pm$2.50 & 69.64$\pm$2.76 & 66.53$\pm$1.93 & 71.6$\pm$2.48 \\
A &  & 45\% & 55.27$\pm$2.50 & 53.93$\pm$3.23 & 51.14$\pm$2.74 & 51.35$\pm$2.62 & 47.20$\pm$3.15 & 49.40$\pm$3.33 & 46.21$\pm$2.12 & 47.77$\pm$5.17 \\
 & GAT & 5\% & 79.50$\pm$1.80 & 80.05$\pm$1.02 & 80.39$\pm$1.14 & 80.45$\pm$1.20 & 79.35$\pm$1.67 & 79.43$\pm$1.51 & 79.45$\pm$1.37 & 79.06$\pm$1.5 \\
 &  & 25\% & 70.33$\pm$2.16 & 69.58$\pm$2.69 & 70.77$\pm$2.57 & 70.48$\pm$1.44 & 65.87$\pm$3.48 & 65.64$\pm$2.34 & 64.80$\pm$2.47 & 66.09$\pm$2.37 \\
 &  & 45\% & 57.46$\pm$3.68 & 57.14$\pm$3.53 & 55.49$\pm$2.94 & 55.56$\pm$3.14 & 44.76$\pm$3.42 & 46.71$\pm$3.98 & 43.82$\pm$3.07 & 45.25$\pm$5.46 \\
  & Graph Transformer & 5\% & 84.42$\pm$0.84 & 84.78$\pm$0.29 & 84.70$\pm$0.41 & 84.64$\pm$0.67 & 84.06$\pm$0.71 & 84.70$\pm$0.14 & 84.96$\pm$0.54 & 84.12$\pm$0.99 \\
 &  & 25\% & 75.30$\pm$0.91 & 76.56$\pm$0.70 & 77.18$\pm$0.67 & 77.50$\pm$0.87 & 72.12$\pm$1.05 & 72.18$\pm$1.08 & 72.88$\pm$0.95 & 70.33$\pm$2.01 \\
 &  & 45\% & 61.16$\pm$1.75 & 62.16$\pm$1.86 & 61.32$\pm$1.56 & 60.62$\pm$2.15 & 50.08$\pm$2.15 & 51.32$\pm$0.92 & 51.28$\pm$1.44 & 48.11$\pm$2.4\\
\hline
& GCN & 5\% & 86.78$\pm$1.43 & 83.65$\pm$6.28 & 85.05$\pm$4.56 & 83.55$\pm$6.18 & 86.66$\pm$1.46 & 84.65$\pm$5.59 & 85.35$\pm$4.21 & 84.85$\pm$2.89 \\
A &  & 25\% & 83.96$\pm$2.99 & 78.28$\pm$9.38 & 81.87$\pm$4.08 & 80.58$\pm$7.92 & 77.00$\pm$5.47 & 74.15$\pm$9.42 & 70.81$\pm$7.69 & 66.82$\pm$8.03 \\
M &  & 45\% & 73.45$\pm$5.67 & 75.36$\pm$8.82 & 66.92$\pm$11.98 & 67.85$\pm$11.36 & 44.15$\pm$4.85 & 49.57$\pm$7.92 & 40.74$\pm$2.65 & 40.03$\pm$5.61 \\
A & GIN & 5\% & 80.24$\pm$4.32 & 81.32$\pm$2.95 & 67.86$\pm$17.91 & 64.96$\pm$15.31 & 66.24$\pm$17.36 & 78.02$\pm$3.71 & 77.10$\pm$5.15 & 34.24$\pm$6.73 \\
Z &  & 25\% & 50.58$\pm$19.81 & 67.24$\pm$16.88 & 35.04$\pm$2.90 & 33.62$\pm$8.99 & 50.16$\pm$10.96 & 69.38$\pm$14.52 & 41.78$\pm$10.62 & 33.94$\pm$7.74 \\
O &  & 45\% & 34.38$\pm$7.71 & 40.28$\pm$4.80 & 24.74$\pm$4.46 & 27.04$\pm$7.18 & 33.52$\pm$17.39 & 48.84$\pm$17.31 & 25.14$\pm$3.94 & 22.58$\pm$7.66 \\
N & GraphSAGE & 5\% & 90.56$\pm$0.86 & 90.41$\pm$0.61 & 90.29$\pm$0.82 & 90.20$\pm$0.64 & 90.39$\pm$1.01 & 90.64$\pm$0.50 & 90.29$\pm$0.79 & 89.93$\pm$1.16 \\
 &  & 25\% & 81.73$\pm$2.29 & 84.06$\pm$1.34 & 82.15$\pm$1.89 & 82.15$\pm$2.09 & 77.96$\pm$2.87 & 80.25$\pm$2.78 & 77.73$\pm$3.13 & 74.08$\pm$4.06 \\
P &  & 45\% & 65.88$\pm$3.47 & 68.72$\pm$2.34 & 61.36$\pm$4.87 & 62.66$\pm$4.42 & 50.78$\pm$4.01 & 55.60$\pm$4.85 & 47.78$\pm$4.23 & 43.61$\pm$6.1 \\
H & GAT & 5\% & 78.20$\pm$1.79 & 79.07$\pm$1.89 & 77.75$\pm$1.95 & 77.29$\pm$1.60 & 77.36$\pm$2.45 & 78.79$\pm$1.52 & 76.30$\pm$1.64 & 76.81$\pm$1.79 \\
O &  & 25\% & 63.20$\pm$3.03 & 66.08$\pm$2.52 & 63.13$\pm$2.32 & 62.52$\pm$2.51 & 61.62$\pm$3.64 & 63.45$\pm$3.42 & 61.21$\pm$1.97 & 58.14$\pm$3.82 \\
T &  & 45\% & 47.79$\pm$3.67 & 48.55$\pm$3.52 & 46.59$\pm$3.19 & 46.70$\pm$3.37 & 45.01$\pm$2.82 & 47.40$\pm$4.53 & 42.69$\pm$3.05 & 40.29$\pm$4.36 \\
O & Graph Transformer & 5\% & 85.57$\pm$0.83 & 80.87$\pm$8.55 & 84.13$\pm$1.88 & 84.25$\pm$1.36 & 85.48$\pm$0.95 & 85.80$\pm$0.72 & 84.32$\pm$1.22 & 84.94$\pm$0.61 \\
 &  & 25\% & 74.34$\pm$3.35 & 76.04$\pm$2.34 & 71.73$\pm$2.25 & 72.54$\pm$3.45 & 71.27$\pm$4.41 & 73.05$\pm$2.94 & 72.28$\pm$1.72 & 69.8$\pm$1.86 \\
 &  & 45\% & 58.77$\pm$2.13 & 58.33$\pm$4.13 & 57.54$\pm$2.01 & 57.76$\pm$1.11 & 54.38$\pm$2.85 & 49.03$\pm$6.71 & 54.25$\pm$3.26 & 49.8$\pm$4.64\\
\hline
\end{tabular}}
\end{table}

\begin{figure}[!h]
    \centering
    \includegraphics[width=0.9 \linewidth]{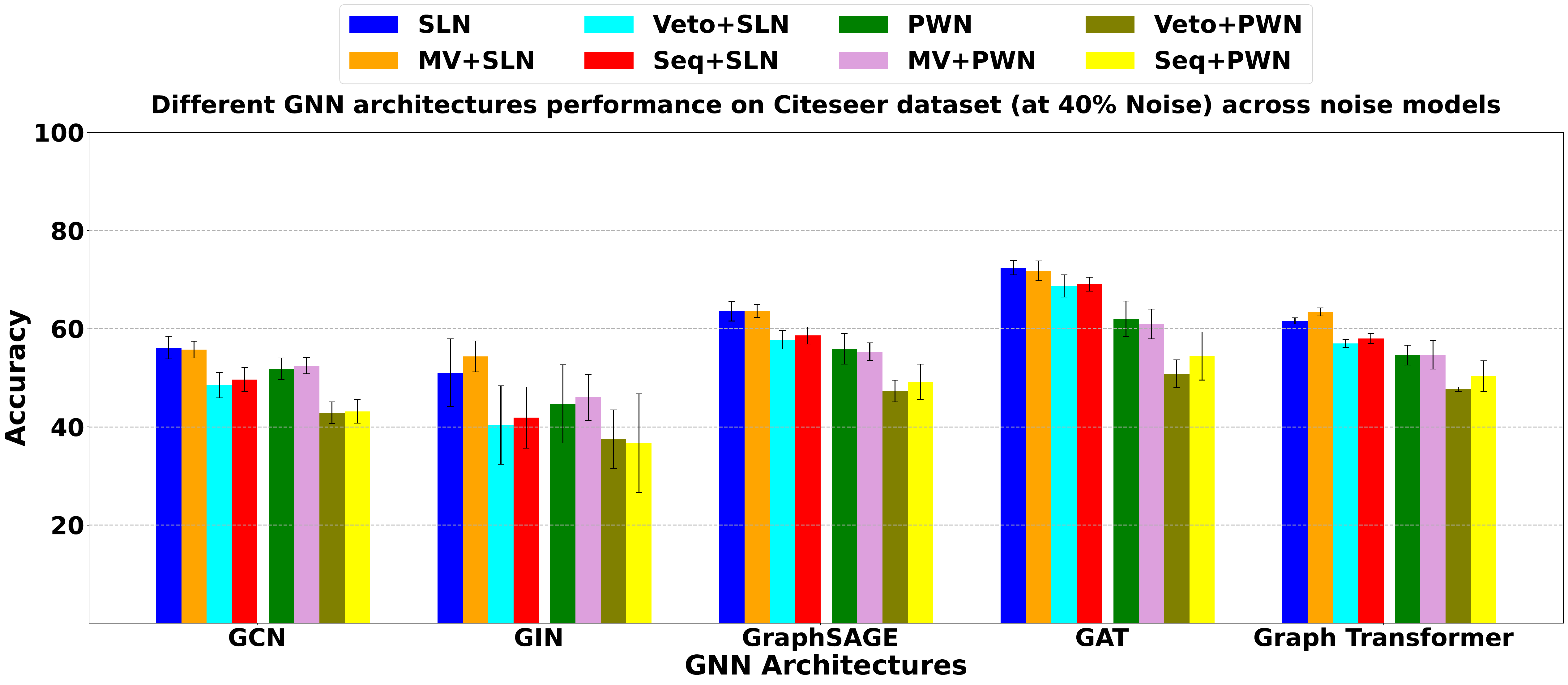}
    \caption{Comparison of noise model variants across GNN architectures. A cluster is for an architecture, and coloured bars show the accuracy of the corresponding noise type.}
    \label{fig:comparison_gnn}
\end{figure}

\subsubsection{Q2. How do other GNN architectures perform in the presence of EDN?} 

Maybe GCN is not robust enough for EDN, what about other GNN architectures, we test common GNN architectures GIN\cite{ginXu2018HowPA}, GraphSage\cite{sage}, GAT\cite{gatVelickovic2017GraphAN}, and Graph Transformer\cite{transforShi2020MaskedLP}. Result for Citeseer with 40\% noise is pictorially reported in Figure \ref{fig:comparison_gnn}. For all datasets at $5\%,25\%,$ and $45\%$ noise levels, result is presented in Table \ref{tab:gnn}. Detailed results for all datasets at all noise levels are in Tables 7,8,9. We observe that, similar to GCN, other GNN architectures' performance is also degraded more in the presence of Veto power and the Sequential flipping variant of EDN as compared to the existing noise model and majority vote variant.

\begin{figure}[!t]
    \centering
    \includegraphics[width=0.95\linewidth]{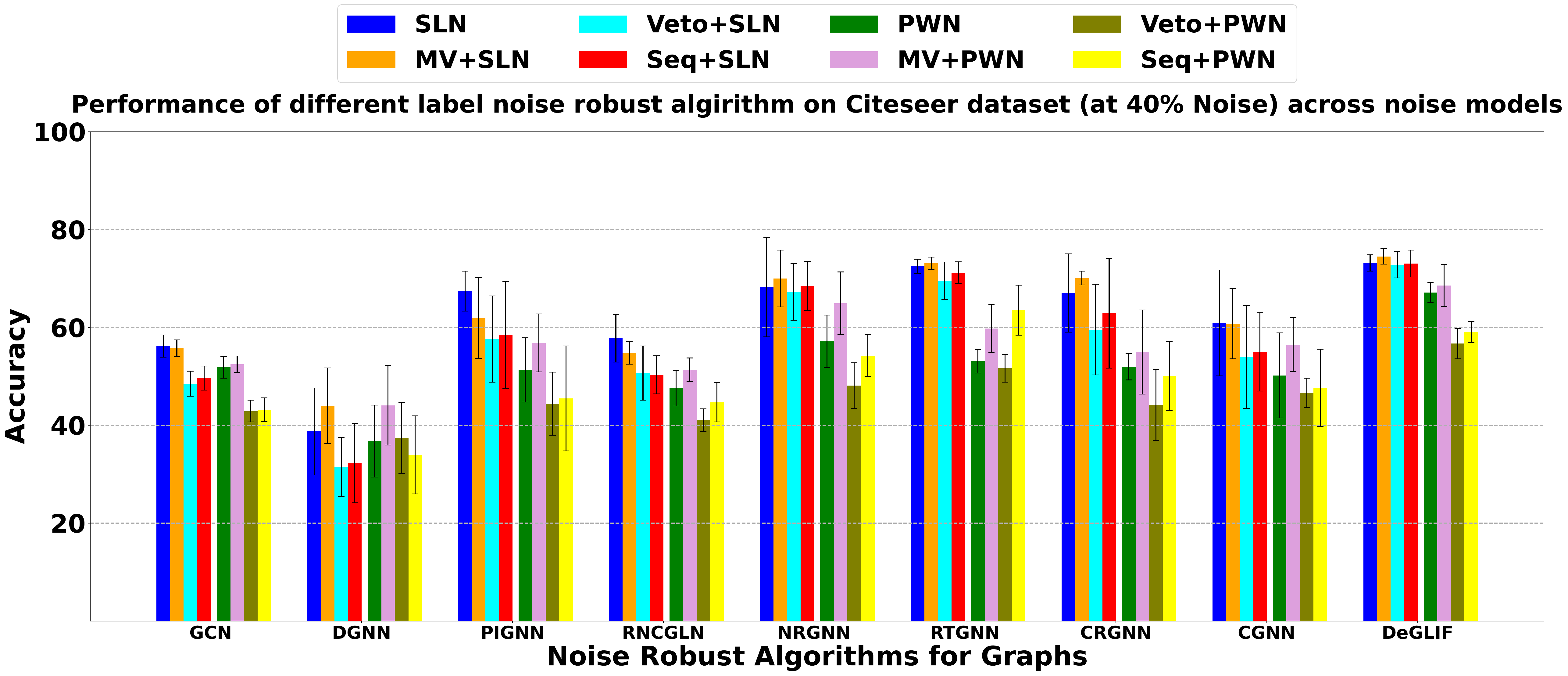}
    \caption{Comparison of noise model variants across graph label noise-robust algorithms. A cluster is for an algorithm, and coloured bars show the accuracy of a noise type.}
    \label{fig:nral}
\end{figure}

\subsubsection{Q3. How does the existing label noise robust algorithm perform in the presence of EDN?}

Next, we evaluate noise-robust algorithms DGNN \cite{NT2019LearningGN}, PIGNN \cite{Du2021NoiserobustGL}, RNCGLN \cite{Zhu2024RobustNC}, NRGNN \cite{dai2021nrgnn}, RTGNN \cite{qian2023robust}, CRGNN \cite{Li2024ContrastiveLOcrgnn}, CGNN \cite{Yuan2023LearningOGcgnn}, and DeGLIF \cite{DeGLIf} under different noise variants of EDN, as well as SLN and CCN. 
 Graphical representation of the result for the Citeseer dataset with 40\% noise is in Figure \ref{fig:nral}. For all datasets at $5\%,25\%,$ and $45\%$ noise level, result is presented in Tables \ref{tab:nrl1} and \ref{tab:nrl2}. Detailed results for all datasets are in Tables 10,11,12 in the supplementary material. At low noise levels, these algorithms give comparable results across all noise models. At higher noise levels, most algorithms (Fig. \ref{fig:nral}) show trends similar to GCN. Performance under the majority vote variant is comparable to or better than SLN or CCN. However, under the Veto Power and sequential variants, almost all algorithms struggle to learn as robustly as they do with SLN or CCN.

The Majority Vote (MV) algorithm penalizes low-degree nodes, making its performance decline similar to SLN or PW. On the other hand, Veto power and Sequential flipping have higher noise levels for higher degrees (Fig \ref{fig:comparison_gnn}). Variants of EDN leading to a greater performance drop suggest that the structure of nodes plays a critical role in how noise impacts performance, making such EDN variants valuable for robust evaluations.

\begin{table}[!h]
\caption{Comparison of noise model variants across graph label noise robust algorithms for Citeseer and Cora datasets. Reported values are accuracy$\pm$std of 10 repetitions.}
\label{tab:nrl1}
\resizebox{\textwidth}{!}{
\begin{tabular}{lr|c|llllllll}
 & \begin{tabular}[c]{@{}c@{}}Noise\\ Robust \\ Methods\end{tabular} & \multicolumn{1}{c|}{\begin{tabular}[c]{@{}c@{}}Noise \\ Level\end{tabular}} & \multicolumn{1}{l}{SLN\ \ \ \ \ \ \ \ \ \ \ } & \multicolumn{1}{c}{\begin{tabular}[l]{@{}l@{}}MV+ \\ SLN\ \ \ \ \ \ \ \ \ \ \ \end{tabular}} & \multicolumn{1}{c}{\begin{tabular}[l]{@{}l@{}}Veto+\\ SLN \ \ \ \ \ \ \ \ \ \ \end{tabular}} & \multicolumn{1}{c}{\begin{tabular}[l]{@{}l@{}}Seq+\\ SLN\ \ \ \ \ \ \ \ \ \ \end{tabular}} & \multicolumn{1}{l}{PWN\ \ \ \ \ \ \ \ \ \ } & \multicolumn{1}{c}{\begin{tabular}[l]{@{}l@{}}MV+\\ PWN\ \ \ \ \ \ \ \ \ \ \end{tabular}} & \multicolumn{1}{c}{\begin{tabular}[l]{@{}l@{}}Veto+\\ PWN\ \ \ \ \ \ \ \ \ \ \end{tabular}}  &  \multicolumn{1}{c}{\begin{tabular}[l]{@{}l@{}}Seq+\\ PWN\end{tabular}} \\ \hline
 & DGNN & 5\% & 66.46$\pm$2.84 & 65.15$\pm$2.37 & 62.17$\pm$2.41 & 60.28$\pm$3.58 & 64.04$\pm$2.22 & 66.06$\pm$1.90 & 61.92$\pm$8.79 & 58.12$\pm$13.30 \\
 &  & 25\% & 53.07$\pm$4.92 & 51.18$\pm$10.06 & 45.02$\pm$6.99 & 44.70$\pm$6.52 & 49.78$\pm$5.92 & 53.62$\pm$6.91 & 46.24$\pm$6.07 & 42.28$\pm$12.04 \\
 &  & 45\% & 41.89$\pm$6.73 & 43.71$\pm$4.72 & 29.48$\pm$6.39 & 27.27$\pm$9.29 & 32.62$\pm$6.84 & 44.71$\pm$5.86 & 32.83$\pm$4.41 & 34.20$\pm$7.13 \\
 & PIGNN  & 5\% & 76.58$\pm$2.04 & 72.83$\pm$3.60 & 71.34$\pm$5.36 & 71.61$\pm$5.36 & 74.02$\pm$2.20 & 73.60$\pm$1.74 & 71.63$\pm$5.38 & 73.14$\pm$2.17 \\
 & & 25\% & 71.61$\pm$3.67 & 69.62$\pm$3.46 & 66.79$\pm$6.75 & 66.11$\pm$8.15 & 66.19$\pm$5.57 & 68.11$\pm$4.49 & 62.79$\pm$4.89 & 63.62$\pm$7.72 \\
 &  & 45\% & 60.79$\pm$11.05 & 56.18$\pm$9.03 & 56.32$\pm$3.70 & 58.49$\pm$7.80 & 44.47$\pm$7.21 & 49.97$\pm$7.33 & 38.14$\pm$4.95 & 44.48$\pm$9.20 \\
 & RNCGLN & 5\% & 72.15$\pm$3.13 & 69.00$\pm$3.88 & 68.33$\pm$1.98 & 67.54$\pm$1.51 & 69.86$\pm$3.23 & 68.05$\pm$2.46 & 68.87$\pm$2.75 & 72.08$\pm$3.06 \\
 &  & 25\% & 65.09$\pm$4.12 & 62.51$\pm$2.37 & 64.76$\pm$4.86 & 62.41$\pm$4.41 & 58.22$\pm$2.80 & 61.01$\pm$2.82 & 56.50$\pm$2.01 & 58.88$\pm$3.38 \\
C &  & 45\% & 51.68$\pm$5.01 & 51.01$\pm$3.35 & 44.24$\pm$4.52 & 45.94$\pm$4.38 & 41.87$\pm$2.96 & 46.81$\pm$3.73 & 37.95$\pm$3.20 & 40.72$\pm$1.63 \\
I & RTGNN  & 5\% & 73.98$\pm$4.38 & 74.31$\pm$1.04 & 74.26$\pm$1.53 & 73.95$\pm$1.35 & 74.18$\pm$0.80 & 74.07$\pm$1.15 & 74.12$\pm$0.97 & 75.08$\pm$1.32 \\
T &  & 25\% & 72.47$\pm$1.78 & 72.81$\pm$1.55 & 72.82$\pm$1.97 & 71.95$\pm$3.23 & 65.87$\pm$2.81 & 71.07$\pm$2.65 & 66.09$\pm$2.46 & 70.22$\pm$3.38 \\
E &  & 45\% & 70.25$\pm$1.97 & 71.09$\pm$2.30 & 64.59$\pm$5.95 & 69.16$\pm$2.81 & 42.61$\pm$4.55 & 53.90$\pm$3.99 & 45.36$\pm$3.73 & 54.05$\pm$3.09 \\
S & NRGNN & 5\% & 75.76$\pm$0.99 & 74.19$\pm$1.47 & 74.22$\pm$1.19 & 74.14$\pm$1.15 & 71.69$\pm$3.75 & 73.56$\pm$2.35 & 74.53$\pm$1.63 & 76.58$\pm$3.30 \\
E &  & 25\% & 73.78$\pm$1.02 & 72.68$\pm$2.15 & 70.86$\pm$4.00 & 70.40$\pm$4.18 & 67.06$\pm$2.90 & 72.08$\pm$2.89 & 66.07$\pm$6.16 & 69.92$\pm$1.67 \\
E &  & 45\% & 70.80$\pm$2.43 & 70.73$\pm$5.18 & 65.17$\pm$6.89 & 64.09$\pm$8.24 & 46.98$\pm$3.99 & 55.79$\pm$5.34 & 40.30$\pm$5.00 & 49.60$\pm$10.83 \\
 R& CRGNN & 5\% & 76.34$\pm$2.41 & 74.74$\pm$1.86 & 74.64$\pm$1.79 & 74.24$\pm$2.41 & 75.09$\pm$1.07 & 74.78$\pm$2.31 & 74.30$\pm$1.48 & 74.88$\pm$3.26 \\
 &  & 25\% & 73.22$\pm$1.63 & 70.35$\pm$6.12 & 71.21$\pm$3.16 & 72.50$\pm$1.97 & 65.98$\pm$4.25 & 70.15$\pm$1.75 & 58.29$\pm$9.09 & 65.34$\pm$3.45 \\
 &  & 45\% & 64.84$\pm$4.24 & 64.16$\pm$8.81 & 57.43$\pm$7.86 & 63.37$\pm$5.93 & 45.19$\pm$3.63 & 52.46$\pm$3.34 & 40.44$\pm$3.00 & 42.56$\pm$6.28 \\
 & CGNN & 5\% & 77.80$\pm$0.83 & 74.20$\pm$1.74 & 73.69$\pm$1.85 & 73.11$\pm$3.40 & 72.67$\pm$4.49 & 73.51$\pm$2.24 & 74.21$\pm$1.59 & 76.60$\pm$4.10 \\
 &  & 25\% & 69.86$\pm$4.46 & 70.54$\pm$3.82 & 69.04$\pm$5.15 & 70.09$\pm$4.03 & 61.63$\pm$9.43 & 67.52$\pm$6.32 & 60.02$\pm$7.17 & 63.08$\pm$9.39 \\
 &  & 45\% & 58.62$\pm$8.03 & 55.96$\pm$11.66 & 50.57$\pm$11.89 & 50.62$\pm$10.95 & 44.70$\pm$6.39 & 50.49$\pm$6.26 & 41.45$\pm$3.50 & 46.02$\pm$4.30 \\
 & DeGLIF & 5\% & 77.58$\pm$1.10 & 77.36$\pm$1.24 & 77.64$\pm$1.71 & 77.50$\pm$1.80 & 77.92$\pm$1.24 & 77.34$\pm$1.41 & 77.64$\pm$1.42 & 77.74$\pm$1.2 \\
 &  & 25\% & 76.10$\pm$1.21 & 76.24$\pm$1.48 & 75.86$\pm$2.28 & 76.04$\pm$1.96 & 74.86$\pm$2.20 & 75.94$\pm$2.44 & 73.40$\pm$1.72 & 74.4$\pm$2.95 \\
 &  & 45\% & 70.58$\pm$1.44 & 72.38$\pm$2.06 & 71.18$\pm$2.99 & 71.98$\pm$2.90 & 59.48$\pm$3.76 & 62.02$\pm$3.25 & 47$\pm$5.34 & 52.48$\pm$8.28 \\ \hline
  & DGNN & 5\% & 78.32$\pm$4.81 & 82.68$\pm$2.81 & 79.80$\pm$1.65 & 78.08$\pm$4.58 & 74.66$\pm$9.15 & 82.50$\pm$0.95 & 78.38$\pm$7.58 & 79.18$\pm$1.75 \\
 &  & 25\% & 74.90$\pm$5.87 & 79.38$\pm$2.39 & 65.20$\pm$11.22 & 65.38$\pm$13.62 & 61.78$\pm$9.81 & 77.36$\pm$3.91 & 53.28$\pm$12.29 & 66.54$\pm$7.27 \\
 &  & 45\% & 62.34$\pm$12.99 & 60.08$\pm$15.94 & 40.26$\pm$8.56 & 40.10$\pm$8.99 & 46.64$\pm$11.84 & 45.02$\pm$11.65 & 30.94$\pm$6.26 & 52.36$\pm$8.54 \\
& PIGNN & 5\% & \multicolumn{1}{l}{80.19$\pm$3.16} & \multicolumn{1}{l}{81.48$\pm$2.16} & \multicolumn{1}{l}{81.81$\pm$2.22} & \multicolumn{1}{l}{81.71$\pm$2.15} & \multicolumn{1}{l}{80.05$\pm$3.50} & \multicolumn{1}{l}{81.58$\pm$1.97} & \multicolumn{1}{l}{81.79$\pm$1.88} & 81.48$\pm$2.84 \\
 &  & 25\% & \multicolumn{1}{l}{79.84$\pm$2.30} & \multicolumn{1}{l}{80.11$\pm$2.09} & \multicolumn{1}{l}{80.66$\pm$2.37} & \multicolumn{1}{l}{78.41$\pm$5.17} & \multicolumn{1}{l}{76.54$\pm$3.43} & \multicolumn{1}{l}{75.71$\pm$3.57} & \multicolumn{1}{l}{72.85$\pm$3.56} & 72.87$\pm$8.57 \\
 &  &  45\% & \multicolumn{1}{l}{75.68$\pm$2.69} & \multicolumn{1}{l}{75.74$\pm$2.92} & \multicolumn{1}{l}{74.95$\pm$6.39} & \multicolumn{1}{l}{74.61$\pm$5.72} & \multicolumn{1}{l}{49.64$\pm$10.92} & \multicolumn{1}{l}{46.89$\pm$9.64} & \multicolumn{1}{l}{47.60$\pm$7.82} & 57.05$\pm$10.22 \\
 & RNCGLN & 5\% & 83.82$\pm$5.00 & 84.06$\pm$4.05 & 85.32$\pm$2.00 & 87.04$\pm$2.01 & 83.78$\pm$2.71 & 82.08$\pm$3.42 & 88.32$\pm$2.02 & 81.38$\pm$0.61 \\
 &  & 25\% & 80.26$\pm$5.05 & 80.24$\pm$3.99 & 71.54$\pm$6.55 & 70.50$\pm$4.53 & 71.94$\pm$5.37 & 77.00$\pm$9.12 & 64.98$\pm$3.93 & 70.60$\pm$1.39 \\
 &  & 45\% & 56.52$\pm$5.79 & 61.22$\pm$6.57 & 51.68$\pm$5.66 & 55.40$\pm$5.21 & 52.40$\pm$8.32 & 63.90$\pm$8.12 & 42.58$\pm$2.93 & 49.72$\pm$0.79 \\
C & RTGNN & 5\% & \multicolumn{1}{l}{73.54$\pm$2.02} & \multicolumn{1}{l}{74.42$\pm$2.12} & \multicolumn{1}{l}{74.64$\pm$1.78} & \multicolumn{1}{l}{75.04$\pm$1.70} & \multicolumn{1}{l}{75.29$\pm$3.09} & \multicolumn{1}{l}{73.91$\pm$2.63} & \multicolumn{1}{l}{74.28$\pm$1.90} & 73.95$\pm$1.58 \\
O &  & 25\% & \multicolumn{1}{l}{75.67$\pm$3.00} & \multicolumn{1}{l}{72.20$\pm$3.70} & \multicolumn{1}{l}{74.12$\pm$3.52} & \multicolumn{1}{l}{73.16$\pm$5.07} & \multicolumn{1}{l}{70.44$\pm$3.29} & \multicolumn{1}{l}{62.89$\pm$6.02} & \multicolumn{1}{l}{55.39$\pm$9.03} & 68.75$\pm$4.56 \\
R &  & 45\% & \multicolumn{1}{l}{66.44$\pm$7.61} & \multicolumn{1}{l}{64.95$\pm$8.76} & \multicolumn{1}{l}{68.27$\pm$6.98} & \multicolumn{1}{l}{71.45$\pm$5.01} & \multicolumn{1}{l}{57.38$\pm$6.77} & \multicolumn{1}{l}{51.81$\pm$3.87} & \multicolumn{1}{l}{40.51$\pm$8.34} & 42.77$\pm$7.32 \\
A & NRGNN & 5\% & \multicolumn{1}{l}{75.13$\pm$3.77} & \multicolumn{1}{l}{75.47$\pm$1.73} & \multicolumn{1}{l}{75.65$\pm$2.62} & \multicolumn{1}{l}{73.19$\pm$2.98} & \multicolumn{1}{l}{75.40$\pm$2.21} & \multicolumn{1}{l}{74.81$\pm$1.82} & \multicolumn{1}{l}{76.12$\pm$2.36} & 74.99$\pm$2.20 \\
 &  & 25\% & \multicolumn{1}{l}{75.03$\pm$4.14} & \multicolumn{1}{l}{73.52$\pm$1.88} & \multicolumn{1}{l}{75.46$\pm$2.96} & \multicolumn{1}{l}{75.05$\pm$1.90} & \multicolumn{1}{l}{69.89$\pm$7.82} & \multicolumn{1}{l}{64.88$\pm$7.89} & \multicolumn{1}{l}{64.97$\pm$6.98} & 65.47$\pm$7.14 \\
 &  & 45\% & \multicolumn{1}{l}{71.63$\pm$7.77} & \multicolumn{1}{l}{65.53$\pm$9.80} & \multicolumn{1}{l}{69.96$\pm$7.19} & \multicolumn{1}{l}{70.07$\pm$5.05} & \multicolumn{1}{l}{50.58$\pm$13.34} & \multicolumn{1}{l}{40.34$\pm$9.91} & \multicolumn{1}{l}{40.91$\pm$10.40} & 49.96$\pm$6.07 \\
 & CRGNN & 5\% & 84.10$\pm$1.86 & 83.99$\pm$1.48 & 84.18$\pm$1.72 & 84.36$\pm$1.53 & 84.26$\pm$1.64 & 84.28$\pm$0.92 & 83.70$\pm$1.58 & 82.16$\pm$4.84 \\
 &  & 25\% & 78.25$\pm$1.88 & 76.40$\pm$2.57 & 76.43$\pm$5.53 & 77.58$\pm$2.64 & 75.71$\pm$2.56 & 74.05$\pm$2.41 & 67.48$\pm$4.64 & 69.94$\pm$6.77 \\
 &  & 45\% & 65.02$\pm$10.42 & 60.58$\pm$12.98 & 62.98$\pm$6.41 & 60.63$\pm$7.09 & 48.32$\pm$9.40 & 49.27$\pm$6.13 & 46.61$\pm$7.83 & 44.22$\pm$7.37 \\
 & CGNN &  5\% & 83.70$\pm$3.67 & 82.43$\pm$4.47 & 83.15$\pm$3.11 & 82.81$\pm$3.26 & 83.57$\pm$2.19 & 83.15$\pm$2.76 & 78.94$\pm$11.48 & 83.27$\pm$2.92 \\
 &  & 25\% & 79.46$\pm$3.45 & 74.93$\pm$5.98 & 77.06$\pm$4.52 & 76.29$\pm$4.45 & 75.82$\pm$3.25 & 70.34$\pm$9.60 & 70.54$\pm$3.38 & 71.37$\pm$10.88 \\
 &  & 45\% & 67.76$\pm$6.19 & 63.57$\pm$9.94 & 65.63$\pm$13.02 & 65.15$\pm$13.00 & 51.20$\pm$7.78 & 47.45$\pm$6.24 & 46.39$\pm$8.02 & 47.56$\pm$10.42 \\ 
 & DeGLIF & 5\% & 88.79$\pm$2.60 & 88.77$\pm$2.79 & 87.04$\pm$6.67 & 89.41$\pm$1.99 & 88.20$\pm$2.12 & 88.73$\pm$2.32 & 88.86$\pm$2.41 & 84.46$\pm$0.8 \\
 &  & 25\% & 87.33$\pm$2.74 & 85.71$\pm$6.37 & 86.17$\pm$3.31 & 85.68$\pm$5.04 & 86.63$\pm$1.84 & 87.60$\pm$2.01 & 73.27$\pm$16.10 & 80.74$\pm$1.15 \\
 &  & 45\% & 84.85$\pm$2.21 & 83.60$\pm$2.00 & 75.33$\pm$7.08 & 78.58$\pm$6.30 & 60.11$\pm$3.78 & 63.50$\pm$6.69 & 44.94$\pm$2.33 & 61.2$\pm$6.07 \\ \hline

\end{tabular}}
\end{table}

\begin{table}[!h]
\caption{Comparison of noise model variants across noise robust algorithms for graphs for the Amazon Photo dataset. Reported values are accuracy$\pm$std of 10 repetitions. }
\label{tab:nrl2}
\resizebox{\textwidth}{!}{
\begin{tabular}{lr|c|llllllll}
 & \begin{tabular}[c]{@{}c@{}}Noise\\ Robust \\ Methods\end{tabular} & \multicolumn{1}{c|}{\begin{tabular}[c]{@{}c@{}}Noise \\ Level\end{tabular}} & \multicolumn{1}{l}{SLN\ \ \ \ \ \ \ \ \ \ \ } & \multicolumn{1}{c}{\begin{tabular}[l]{@{}l@{}}MV+ \\ SLN\ \ \ \ \ \ \ \ \ \ \ \end{tabular}} & \multicolumn{1}{c}{\begin{tabular}[l]{@{}l@{}}Veto+\\ SLN \ \ \ \ \ \ \ \ \ \ \end{tabular}} & \multicolumn{1}{c}{\begin{tabular}[l]{@{}l@{}}Seq+\\ SLN\ \ \ \ \ \ \ \ \ \ \end{tabular}} & \multicolumn{1}{l}{PWN\ \ \ \ \ \ \ \ \ \ } & \multicolumn{1}{c}{\begin{tabular}[l]{@{}l@{}}MV+\\ PWN\ \ \ \ \ \ \ \ \ \ \end{tabular}} & \multicolumn{1}{c}{\begin{tabular}[l]{@{}l@{}}Veto+\\ PWN\ \ \ \ \ \ \ \ \ \ \end{tabular}}  &  \multicolumn{1}{c}{\begin{tabular}[l]{@{}l@{}}Seq+\\ PWN\end{tabular}} \\ \hline

& DGNN & 5\% & 78.32$\pm$4.81 & 82.68$\pm$2.81 & 79.80$\pm$1.65 & 78.08$\pm$4.58 & 74.66$\pm$9.15 & 82.50$\pm$0.95 & 78.38$\pm$7.58 & 61.44$\pm$27.08 \\
 &  & 25\% & 74.90$\pm$5.87 & 79.38$\pm$2.39 & 65.20$\pm$11.22 & 65.38$\pm$13.62 & 61.78$\pm$9.81 & 77.36$\pm$3.91 & 53.28$\pm$12.29 & 55.50$\pm$5.54 \\
 &  & 45\% & 62.34$\pm$12.99 & 60.08$\pm$15.94 & 40.26$\pm$8.56 & 40.10$\pm$8.99 & 46.64$\pm$11.84 & 45.02$\pm$11.65 & 30.94$\pm$6.26 & 35.96$\pm$4.80 \\
 & PIGNN & 5\% & 88.9$\pm$0.4 & 89.56$\pm$0.58 & 90.72$\pm$0.31 & 92.42$\pm$0.59 & 89.74$\pm$1.26 & 88.02$\pm$0.96 & 89.96$\pm$0.58 & 90.06$\pm$0.88 \\
 &  & 25\% & 86.8$\pm$3.4 & 88.96$\pm$1.72 & 90.30$\pm$0.65 & 90.52$\pm$0.50 & 85.22$\pm$2.21 & 87.52$\pm$2.15 & 74.56$\pm$2.17 & 82.10$\pm$2.06 \\
A &  & 45\% & 82.2$\pm$4.2 & 87.22$\pm$2.37 & 79.86$\pm$7.22 & 80.64$\pm$3.36 & 60.44$\pm$8.51 & 62.38$\pm$9.04 & 42.86$\pm$6.44 & 63.52$\pm$2.81 \\
M & RNCGLN & 5\% & 83.82$\pm$5.00 & 84.06$\pm$4.05 & 85.32$\pm$2.00 & 87.04$\pm$2.01 & 83.78$\pm$2.71 & 82.08$\pm$3.42 & 88.32$\pm$2.02 & 84.48$\pm$3.53 \\
A &  & 25\% & 80.26$\pm$5.05 & 80.24$\pm$3.99 & 71.54$\pm$6.55 & 70.50$\pm$4.53 & 71.94$\pm$5.37 & 77.00$\pm$9.12 & 64.98$\pm$3.93 & 69.96$\pm$6.36 \\
Z &  & 45\% & 56.52$\pm$5.79 & 61.22$\pm$6.57 & 51.68$\pm$5.66 & 55.40$\pm$5.21 & 52.40$\pm$8.32 & 63.90$\pm$8.12 & 42.58$\pm$2.93 & 48.34$\pm$1.06 \\
O & RTGNN & 5\% & 80.8$\pm$5.3 & 81.93$\pm$2.17 & 81.46$\pm$2.44 & 84.14$\pm$2.19 & 82.24$\pm$0.86 & 83.45$\pm$1.10 & 81.95$\pm$1.42 & 82.04$\pm$1.53 \\
N &  & 25\% & 82.9$\pm$4.9 & 82.63$\pm$3.85 & 82.93$\pm$3.32 & 81.96$\pm$2.17 & 83.17$\pm$3.79 & 84.88$\pm$3.15 & 69.78$\pm$3.62 & 76.19$\pm$6.53 \\
 &  & 45\% & 86$\pm$1.3 & 84.98$\pm$1.62 & 75.71$\pm$7.21 & 74.46$\pm$5.80 & 60.86$\pm$8.45 & 58.03$\pm$12.32 & 47.17$\pm$8.88 & 60.56$\pm$3.57 \\
P & NRGNN & 5\% & 69$\pm$8 & 87.52$\pm$0.90 & 87.34$\pm$1.68 & 88.40$\pm$2.77 & 89.74$\pm$1.26 & 87.08$\pm$1.67 & 86.56$\pm$1.54 & 85.90$\pm$2.81 \\
H &  & 25\% & 55.1$\pm$5.4 & 86.92$\pm$2.84 & 85.92$\pm$0.64 & 86.08$\pm$0.90 & 85.22$\pm$2.21 & 85.52$\pm$2.32 & 72.50$\pm$4.41 & 81.24$\pm$5.63 \\
O &  & 45\% & 54.5$\pm$6.2 & 86.70$\pm$1.72 & 73.78$\pm$4.98 & 81.42$\pm$2.51 & 60.44$\pm$8.51 & 66.20$\pm$8.31 & 49.90$\pm$3.87 & 58.78$\pm$1.89 \\
T & CRGNN & 5\% & 59.04$\pm$12.73 & 44.60$\pm$12.81 & 54.28$\pm$14.68 & 54.52$\pm$14.41 & 54.80$\pm$12.30 & 52.82$\pm$12.77 & 47.36$\pm$15.24 & 37.54$\pm$46.21 \\
O &  & 25\% & 45.02$\pm$10.86 & 37.82$\pm$11.13 & 35.76$\pm$11.64 & 35.22$\pm$14.65 & 47.68$\pm$20.70 & 31.22$\pm$7.03 & 41.32$\pm$7.76 & 32.74$\pm$39.74 \\
 &  & 45\% & 29.48$\pm$15.05 & 31.20$\pm$5.21 & 23.12$\pm$5.72 & 18.82$\pm$4.97 & 33.14$\pm$13.04 & 37.74$\pm$8.31 & 27.42$\pm$4.28 & 19.78$\pm$21.92 \\
 & CGNN & 5\% & 39.08$\pm$28.12 & 28.70$\pm$12.19 & 23.32$\pm$6.22 & 33.60$\pm$21.41 & 33.82$\pm$19.18 & 25.10$\pm$7.71 & 32.08$\pm$11.53 & 56.84$\pm$25.81 \\
 &  & 25\% & 34.08$\pm$23.35 & 22.70$\pm$8.20 & 24.30$\pm$5.93 & 25.88$\pm$24.09 & 31.26$\pm$18.21 & 22.48$\pm$3.59 & 29.08$\pm$7.08 & 47.86$\pm$14.05 \\
 &  & 45\% & 16.62$\pm$9.35 & 17.40$\pm$9.27 & 20.84$\pm$7.90 & 22.96$\pm$11.08 & 22.74$\pm$12.74 & 24.44$\pm$7.80 & 25.78$\pm$7.39 & 36.52$\pm$9.30 \\
 & DeGLIF & 5\% & 88.79$\pm$2.60 & 88.77$\pm$2.79 & 87.04$\pm$6.67 & 89.41$\pm$1.99 & 88.20$\pm$2.12 & 88.73$\pm$2.32 & 88.86$\pm$2.41 & 89.09$\pm$2.13 \\
 &  & 25\% & 87.33$\pm$2.74 & 85.71$\pm$6.37 & 86.17$\pm$3.31 & 85.68$\pm$5.04 & 86.63$\pm$1.84 & 87.60$\pm$2.01 & 73.27$\pm$16.10 &79.08$\pm$4.45  \\
 &  & 45\% & 84.85$\pm$2.21 & 83.60$\pm$2.00 & 75.33$\pm$7.08 & 78.58$\pm$6.30 & 60.11$\pm$3.78 & 63.50$\pm$6.69 & 44.94$\pm$2.33 &52.92$\pm$11.73 \\ 
 \hline 
\end{tabular}}
\end{table}

\subsection{Hypothesis Testing}


All experiments to check the impact of EDN on GNN architectures and noise robust algorithms for graphs are repeated 10 times. For a particular noise level,  Let $\{x_{1}^s,\ldots x_{10}^s\}$, denote accuracy values obtained by GNN when noise is injected using SLN. Similarly let $\{x_{1}^v,\ldots, x_{1}^v\}$ denote accuracy values obtained by GNN when noise is injected using Veto Power + SLN model. We want to use the theory of hypothesis testing to check if some of the variant of EDN really lead to more degradation in performance as compared to existing noise models. To do so, let, $d_i=x_{i}^s-x_i^v$. Then $\Bar{d}=\Bar{x}^s-\Bar{x}^v$. We assume that $d_i$ are sampled from a normal distribution with an unknown mean $\mu_d$ and unknown variance $\sigma_d^2$. We define hypothesis test as follows \cite{ross2020introduction,casella2024statistical}:
\begin{itemize}
    \item[ ]

 Null Hypothesis $H_0: \mu_d\leq 0$

 \item[ ] Alternate Hypothesis $H_1: \mu_d>0$
\end{itemize}



 We would like to mention that the normal distribution has a domain of $(-\infty, \infty)$, but the possible values of $d_i$ are restricted to $[-100, 100]$. Empirically, almost all observed values of $d_i$ lie within $[-10, 10]$ and exhibit a small standard deviation. As a result, the probability of a well-fitted normal distribution assigning values beyond $[-100, 100]$ is extremely low. Therefore, it is reasonable to assume that $d_i$ are sampled from a normal distribution. The estimate for a mean of $d_i$ is given by $\Bar{d_i}$. The pooled estimator $S_d^2$ for variance of $d_i$ is given by (assuming that population variance of SLN and Veto+SLN are the same)
$$S_d^2=\frac{(n-1)S_s^2+(n-1)S_v^2}{2n-2}=\frac{S_s^2+S_v^2}{2}$$
where, $S_s^2$ and $S_v^2$ are sample variance for SLN and Veto+SLN varaints respectively. Then, the test statistic is given by 
$$T=\frac{\Bar{d}-0}{\sqrt{S_d^2\left(\frac{1}{n}+\frac{1}{n}\right)}}=\frac{\sqrt{n}\Bar{d}}{\sqrt{S_s^2+S_v^2}}$$

The significance level $\alpha$ test is to \emph{reject $H_0$ if $T\geq t_{\alpha,2n-2}$}; not reject $H_0$, otherwise.
Here $t_{\alpha,2n-2}$ denotes t-distribution with degree of freedom $2n-2$. Significance level $\alpha$ means that the probability of $H_0$ getting rejected when it is actually true is never greater than $\alpha$. In our experiment, we expect the data to support the alternative hypothesis $H_1$, but do not want to make the assertion unless the data really gives convincing support. So, we have set up the test so that the alternate hypothesis is the one that we expect data to support and we hope to prove. Alternate hypothesis in such setup is also called the research hypothesis. By choosing a small $\alpha$ as the significance level, we minimize the risk of incorrectly concluding that the data supports the research hypothesis when it is actually false \cite{casella2024statistical}. Common choices for $\alpha$ are $0.1$, $0.05$, and $0.005$; here, we choose $\alpha = 0.05$.

\subsubsection{Illustrative examples} 

\noindent
{\sf (A}) Let us look at a few examples. For Citeseer data set, GCN architecture in presence of $5\%$ noise, we have $\Bar{x}^s=73.92$, $\Bar{x}^v=72.27$,  $S_s=1.1$, $S_v=0.7$ (from Table \ref{tab:gnn}), and thus, $T=\frac{\sqrt{10}\times(73.92-72.27)}{\sqrt{1.1^2+0.7^2}}=4.00184.$
For significance level $\alpha=0.05$, the value of $t_{\alpha,18}=1.734$. As, $T>t_{0.05,18}$ so we reject $H_0$. It means for the Citeseer dataset at $5\%$ noise, GCN performs worse in the presence of Veto + SLN as compared to SLN, and the probability of incorrectly concluding this is at most 5\%. In fact, in this case, $T>t_{0.0005,18}=3.921643$, so the probability of incorrectly concluding rejection of $H_0$ is at most 0.05\%.

{\sf (B)} For Citeseer data set, GIN architecture in presence of $45\%$ noise, we have $\Bar{x}^s=42.29$, $\Bar{x}^v=37.98$,  $S_s=8.8$, $S_v=5.4$ (from Table \ref{tab:gnn}), and hence, $T= 1.32$. This means, $T<t_{0.05,18}$ and we accept $H_0$.
\subsubsection{Analysis of Hypothesis Tests:} 
We perform hypothesis testing for all datasets, for all GNN architectures and all noise robust algorithms and the results are presented in Table \ref{tab:hypo}. We divide noise levels into three subclass, Low ($5\% -15\%$ noise levels), Medium ($20\%-35\%$ noise levels), and High ($40\%-50\%$ noise levels). Overall represents aggregate across all noise levels. We revisit questions asked in Section \ref{subsec:comp_results} through the lens of hypothesis testing. For GCN architectures, we have 10 noise levels $\times$ 3 datasets $\times$ 2 sub-variants (SLN and PWN) = 60 hypothesis tests. Out of 60, we were able to reject 32 null hypotheses with $\alpha=0.05$. As in {\sf (B)}, accepting $H_0$ at a low significance level does not always mean that the null hypothesis is true, but it means that we are unable to reject it with high confidence. In example {\sf (B)} clearly $\Bar{d}>0$, but as the variances are high, we are not confident if $\Bar{d}>0$ is due to a change in noise model or due to inherent randomness in the learning process (randomness in training set sampling, stochastic gradient descent, etc). So, when we are able to reject $H_0$, we can say with high confidence that the decrease in performance is due to a change in the noise model (for example, a change from SLN to Veto+SLN). 
\begin{table}[!h]
\caption{Summary of all Hypothesis Tests: The fraction of cases in which we rejected the null hypothesis at significance level $\alpha=0.05$.  Even algorithms designed to handle graph label noise experience greater performance degradation due to EDN compared to the existing noise models. We can say this with high confidence for 34\% cases under Veto Power and  22\% cases under Sequential Flipping. The same is true for 41\% cases for GNN architectures.}
\label{tab:hypo}
\centering
\begin{tabular}{l|l|l l l}
\begin{tabular}[c]{@{}l@{}}Noise \\ Model\end{tabular} & \begin{tabular}[c]{@{}l@{}}Noise \\ Level\end{tabular} & GCN \ \ \ \ \ \ \ \ \ \ \  \ & \begin{tabular}[c]{@{}l@{}}GNN \ \ \ \ \ \ \ \ \ \ \ \ \ \ \\ Architectures\end{tabular} & \begin{tabular}[c]{@{}l@{}} Existing \\ Noise Robust \\Algorithms\end{tabular} \\ \hline
 & Low & 6/18 $\approx$ 0.33 & 24/90 $\approx$ 0.27 & 26/144 $\approx$ 0.18 \\
Veto & Medium & 10/24 $\approx$ 0.42 & 45/120 $\approx$ 0.38 & 54/192 $\approx$ 0.28 \\
Power & High & 16/18 $\approx$ 0.89 & 55/90 $\approx$ 0.61 & 85/144 $\approx$ 0.59 \\
 & Overall & 32/60 $\approx$ 0.53 & 124/300 $\approx$ 0.41 & 165/480 $\approx$ 0.34 \\ \hline
 & Low & 7/18 $\approx$ 0.38 & 23/90 $\approx$ 0.26 & 26/144 $\approx$ 0.18 \\
Sequential & Medium & 9/24 $\approx$ 0.38 & 41/120 $\approx$ 0.34 & 41/192 $\approx$ 0.21 \\
Flipping & High & 14/18 $\approx$ 0.78 & 59/90 $\approx$ 0.66 & 39/144 $\approx$ 0.27 \\
 & Overall & 30/60 $\approx$ 0.5 & 123/300 $\approx$ 0.41 & 106/480 $\approx$ 0.22 \\ \hline
\end{tabular}
\end{table}

From Table \ref{tab:hypo}, we observe that Veto Power and Sequential Flipping cause greater performance degradation than traditional noise models across different noise levels. GCN is the most affected by EDN, while other GNN architectures show slightly more robustness. However, none of the GNN architecture performs similarly to the existing noise model and EDN across noise levels. Existing noise robust algorithms also fail to completely tackle two variants of EDN. Overall, at $\alpha=0.05$ significance level, in 34\% cases, Veto power degrades the performance of noise robust algorithm more as compared to existing noise models; sequential flipping degrades more performance in $22\%$ of cases. This number increases to $41\%$ for GNN architectures and increases to $50\%$ for GCN. Also, we observe that the impact of EDN becomes prominent with an increase in noise levels.

\section{Conclusion}
In this work, we introduce a novel noise model for graph data called Edge-Dependent Noise (EDN). Unlike existing noise models used for graph data that were originally designed for i.i.d. data, EDN captures the impact of connections among nodes, on node label noise. We propose three variants of EDN - Majority Vote, Veto Power, and Sequential Flipping. In all three variants, the probability of a node's label being flipped is directly determined by its degree, making implementation feasible even for large graphs. This degree-dependency is a distinguishing feature of EDN. We theoretically compare the probabilities of label flipping as a function of the node degree for various EDN variants that we propose. Experiments followed by hypothesis testing on results reveal that EDN, especially the Veto power and Sequential flipping variants, leads to more significant performance degradation compared to existing noise models like SLN and CCN.
This highlights the critical role of node 
in understanding the impact of noise on GNN performance, making EDN a valuable tool for robust evaluations of GNNs. This underscores the need for further research into developing noise-robust algorithms specifically designed to handle the complexities of edge-dependent noise in graph data. As the differences in accuracies that we consider are in the interval $[-100, 100]$, one can pursue the hypothesis testing approach in a more principled way without resorting to normal approximations or assuming equal population variance. Our hypothesis testing framework is Berhens-Fisher problem and currently has no completely satisfactory solution (Sec. 8.4.3 of  \cite{ross2020introduction}).



\bibliographystyle{unsrt}


\newpage

\appendix


\section{Analysis of Sequential flipping + SLN}
\label{app:der}
For a given dataset having \textbf{$K$ classes}, we are interested in finding the probability with which the label of a node of \textbf{degree $n$} is changed. We assumed that the probability of transitioning between any two classes is uniform. So, when $\rho$ is the probability of calling an edge noisy, the transition probability matrix due to an edge is given by 
$$Q=\begin{bmatrix}
1- \rho\ \ & \frac{\rho}{K-1} &\ \ \frac{\rho}{K-1}  & \ldots & \ \ \ \frac{\rho}{K-1} \\ 
 \frac{\rho}{K-1}& 1-\rho \ \  & \ \ \frac{\rho}{K-1}  \ \ &  & \ \ \ \frac{\rho}{K-1}\\ 
\vdots &  & \ \ \ddots & \ \ \ddots &\ \ \  \vdots\\ 
 \frac{\rho}{K-1} &  &  & \ \ \ 1-\rho & \ \ \ \frac{\rho}{K-1} \\ 
\frac{\rho}{K-1}  &\frac{\rho}{K-1}   & \dots &\ \ \ \frac{\rho}{K-1}   &\ \ \ 1- \rho 
\end{bmatrix}$$
For ease of notation let $p=\frac{\rho}{K-1}$, the matrix get modified to 

$$
Q=\begin{bmatrix}
1- (K-1)p& p &p & \ldots &p \\ 
 p& 1-(K-1)p &p &  &p\\ 
\vdots &  & \ddots &\ddots &\vdots\\ 
 p &  &  & 1-(K-1)p & p \\ 
p  &p   & \dots &p  &1- (K-1)p 
\end{bmatrix}$$
As the label is getting sequentially updated for every incident edge, the transition probability matrix for a node with degree $n$ is $Q^n$. As $Q$ is a symmetric matrix, we will try to diagonalize the matrix in order to find a closed-form solution for $Q^n$. Observe that the sum of every row of the matrix $Q$ is the same, and they sum up to 1. This means one of the eigenvectors is $v_1=\frac{1}{\sqrt{K}}[1,1,\ldots,1]$ for eigenvalue $\lambda_1=1$. To find other eigenvalues, let $v=[x_1,\ldots,x_K]$ be an eigenvector of $Q$, then it satisfies 
$$ (1-(K-1)p)x_i+p(x_1+x_2+\ldots+x_k-x_i)=\lambda x_i,\ \ \ \ \ i=1,\ldots,k$$
 As $Q$ is a symmetric matrix, so all other eigenvectors must be orthogonal to $v_1$. Hence, $x_1+\ldots+x_K=0$, and
 $$(1-Kp)x_i=\lambda x_i,\ \ \ \ \ i=1,\ldots,k$$

 This means $(1-Kp)$ is an eigenvalue for $Q$. Eigenvector corresponding to this eigenvalue satisfies $Q-(1-Kp)I=0$, that is 
 $$Q-(1-Kp)I=\begin{bmatrix}
p\ \ & p &\ \ p  & \ldots & \ \ \ p \\ 
 p& p \ \  & \ \ p  \ \ &  & \ \ \ p\\ 
\vdots &  & \ \ \ddots & \ \ \ddots &\ \ \  \vdots\\ 
 p &  &  & \ \ \ p & \ \ \ p \\ 
p  &p   & \dots &\ \ \ p   &\ \ \ p 
\end{bmatrix}
\begin{bmatrix}
    x_1\\x_2\\ \vdots \\ x_{K-1}\\ x_K
\end{bmatrix}=0
$$
As the solution space of given system of equation have $k-1$ dimension, so we obtain $k-1$ linearly independent eigenvectors given by 
\\
$v_2=\frac{1}{\sqrt{2}}[1,-1,0,0,\ldots,0,0],\\v_3=\frac{1}{\sqrt{6}}[1,1,-2,0,\ldots,0,0],\\
v_4=\frac{1}{\sqrt{12}}[1,1,1,-3,0,\ldots,0,0], \\ \vdots \\
v_K=\frac{1}{\sqrt{(K-1)K}}[1,1,1,1,\ldots,1,-(K-1)]$

Again, as $Q$ is a symmetric matrix using eigen decomposition of a symmetric matrix, assuming eigenvectors are row vectors  

$$Q^n=\sum_{i=1}^K \lambda_i^n v_i^{\top}v_i$$
Separating the term corresponding to $\lambda=1$ gives 

\begin{equation*}
    \begin{aligned}
Q^n&= \frac{1}{K}\begin{bmatrix}
1\ \ & 1 &\ \ 1  & \ldots & \ \ \ 1 \\ 
 1& 1 \ \  & \ \ 1  \ \ &  & \ \ \ 1\\ 
\vdots &  & \ \ \ddots & \ \ \ddots &\ \ \  \vdots\\ 
 1 &  &  & \ \ \ 1 & \ \ \ 1 \\ 
1  &1   & \dots &\ \ \ 1   &\ \ \ 1 
\end{bmatrix}+
\sum_{i=2}^K \lambda_i^n v_i^{\top}v_i
\\
&=\frac{1}{K}\begin{bmatrix}
1\ \ & 1 &\ \ 1  & \ldots & \ \ \ 1 \\ 
 1& 1 \ \  & \ \ 1  \ \ &  & \ \ \ 1\\ 
\vdots &  & \ \ \ddots & \ \ \ddots &\ \ \  \vdots\\ 
 1 &  &  & \ \ \ 1 & \ \ \ 1 \\ 
1  &1   & \dots &\ \ \ 1   &\ \ \ 1 
\end{bmatrix}+(1-Kp)^n 
\begin{bmatrix}
    \vrule height 1.5ex width 0.5pt & \vrule height 1.5ex width 0.5pt& \ldots & \vrule height 1.5ex width 0.5pt \\
    v_2 & v_3 & \ldots& v_K \\
    \vrule height 1.5ex width 0.5pt & \vrule height 1.5ex width 0.5pt& \ldots & \vrule height 1.5ex width 0.5pt
\end{bmatrix}
\begin{bmatrix}
   - v_2 -\\ - v_3 -\\ \vdots \\ - v_K -
\end{bmatrix}
 \\ 
&=\frac{1}{K}\begin{bmatrix}
1\ \ & 1 &\ \ 1  & \ldots & \ \ \ 1 \\ 
 1& 1 \ \  & \ \ 1  \ \ &  & \ \ \ 1\\ 
\vdots &  & \ \ \ddots & \ \ \ddots &\ \ \  \vdots\\ 
 1 &  &  & \ \ \ 1 & \ \ \ 1 \\ 
1  &1   & \dots &\ \ \ 1   &\ \ \ 1 
\end{bmatrix}+(1-Kp)^n 
\begin{bmatrix}
\frac{K-1}{K} & -\frac{1}{K} & -\frac{1}{K}  & \ldots &  -\frac{1}{K} \\ 
 -\frac{1}{K}& \frac{K-1}{K}   &  -\frac{1}{K}  &  & -\frac{1}{K}\\ 
\vdots &  & \ddots &\ddots &  \vdots\\ 
 -\frac{1}{K} &  &  & \frac{K-1}{K} &  -\frac{1}{K} \\ 
-\frac{1}{K}  &-\frac{1}{K}   & \dots & -\frac{1}{K}   & \frac{K-1}{K} 
\end{bmatrix}
\end{aligned}
    \end{equation*}

Now, using our original notation $\rho=(K-1)p$. Using $Q^n$, starting with the true label $y_i$, the probability of the label being changed to a specific class is given by 
$$s\_sc(n)=\frac{1}{K} \left( 1-\left(1-\frac{K\rho}{K-1}\right )^n\right).$$ 

For a node of degree $n$, the probability of its label being flipped is $(K-1)\times s\_sc(n)$ and is hence given by 
$$s(n)=\frac{K-1}{K} \left( 1-\left(1-\frac{K\rho}{K-1}\right )^n\right).$$
\section{Analysis of Sequential flipping + PWN model}
\label{app:der_pair}
\begin{lemma}
\label{lemma:circ}
    The eigenvectors for a circulant matrix $$A=\begin{bmatrix}
        c_0 & c_1 &\ldots &c_{K-1}\\
        c_{K-1} & c_0 & \dots & c_{K-2}\\
        \vdots & \vdots & \vdots \vdots \vdots & \vdots\\
        c_1 & c_2& \ldots& c_0
    \end{bmatrix}$$ are the columns of the matrix
    $$F=\frac{1}{\sqrt{K}}\begin{bmatrix}
        1 & 1 & 1 & \ldots & 1\\
        1 & \omega_K & \omega_K^2 &\ldots &\omega_K^{K-1}\\
        1 & \omega_K^2 & \omega_K^4 &\ldots &\omega_K^{2(K-1)}\\
        \vdots & \vdots &\vdots & \vdots \vdots \vdots & \vdots\\
         1 & \omega_K^{(K-1)} & \omega_K^{2(K-1)} &\ldots &\omega_K^{(K-1)(K-1)}
    \end{bmatrix}$$
Where $\omega_K$ is the $K$-th root of unity. The vector of corresponding eigenvalues is given by $\lambda=[\lambda_0,\ldots,\lambda_{K-1}]'=\sqrt{K}F[c_0,c_1,\ldots,c_{K-1}]'$. F is a unitary matrix $(FF^*=F^*F=I$, where * denotes conjugate transpose). Let $\Lambda=$daig$(\lambda_0,\dots,\lambda_{K-1})$, then the eigen decomposition of $A$ is given by $A=F\Lambda F^*$.
\end{lemma}

Again for a given dataset having $K$ classes, we are interested in finding the probability with which the label of degree $n$ is changed.
For sequential flipping + Pairwise noise, transition probability matrix due to a single edge is given by 

$$Q_{pwn}=
\begin{bmatrix}
1- \rho\ \ & \rho &\ \ 0  & \ldots & \ \ \ 0 \\ 
 0& 1-\rho \ \  & \ \ \rho \ \ &  & \ \ \ 0\\ 
\vdots &  & \ \ \ddots & \ \ \ddots &\ \ \  \vdots\\ 
 0&  &  & \ \ \ 1-\rho & \ \ \ \rho\\ 
\rho &0  & \dots &\ \ \ 0  &\ \ \ 1- \rho 
\end{bmatrix}
\label{eq:pair_matrix}
$$
Similar to the Sequential flipping + SLN case, we need to calculate $Q_{pwn}^n$. As $Q_{pwn}$ is a circulant matrix, so using Lemma \ref{lemma:circ} vector of its eigen values is given by 

\begin{equation}
\label{eq:eig_pair}
\lambda=\begin{bmatrix}
        1 & 1 & 1 & \ldots & 1\\
        1 & \omega_K & \omega_K^2 &\ldots &\omega_K^{K-1}\\
        1 & \omega_K^2 & \omega_K^4 &\ldots &\omega_K^{2(K-1)}\\
        \vdots & \vdots &\vdots & \vdots \vdots \vdots & \vdots\\
         1 & \omega_K^{(K-1)} & \omega_K^{2(K-1)} &\ldots &\omega_K^{(K-1)(K-1)}
    \end{bmatrix}
    \begin{bmatrix}
        1-\rho\\ \rho \\ 0 \\ \vdots \\0
    \end{bmatrix}=\begin{bmatrix}
            1-\rho+\rho\times\omega_K^0\\
            1-\rho+\rho\times\omega_K^1\\
            1-\rho+\rho\times\omega_K^2\\
            \vdots\\
            1-\rho+\rho\times\omega_K^{(K-1)}
        \end{bmatrix}\end{equation}
 As $Q_{pwn}$ is circulant, and product of circulant matrix is circulant, finding first row $Q_{pwn}^{n}$ is sufficient. Using Lemma \ref{lemma:circ} and Equation \ref{eq:eig_pair}

 \begin{equation}\begin{aligned}
     Q_{pwn}^n&=F\begin{bmatrix} 
                 (1-\rho+\rho\times\omega_K^0)^n & 0 & 0 & \ldots & 0\\
        0 & (1-\rho+\rho\times\omega_K^1)^n & 0 &\ldots &0\\
        0 & 0 & (1-\rho+\rho\times\omega_K^2)^n &\ldots &0\\
        \vdots & \vdots &\vdots & \vdots \vdots \vdots & \vdots\\
         0 & 0 & 0 &\ldots &(1-\rho+\rho\times\omega_K^{K-1})^n
     \end{bmatrix} F^*\\&
     = F Diag(\lambda) F^*
     \end{aligned}
 \end{equation}

 First row of $Q_{pwn}^n$ (denoted by $Q_{pwn}^n[0,:]$)is given by 

 \begin{equation}
 \begin{aligned}
     Q_{pwn}^n[0,:]&=\frac{1}K{}[1,1,1,\dots,1]\times Diag(\lambda)\times 
     \begin{bmatrix}
        1 & 1 & 1 & \ldots & 1\\
        1 & \omega_K^* & \omega_K^{2*} &\ldots &\omega_K^{K-1*}\\
        1 & \omega_K^{2*} & \omega_K^{4*} &\ldots &\omega_K^{2(K-1)*}\\
        \vdots & \vdots &\vdots & \vdots \vdots \vdots & \vdots\\
         1 & \omega_K^{(K-1)^*} & \omega_K^{2(K-1)^*} &\ldots &\omega_K^{(K-1)(K-1)*}
    \end{bmatrix}\\ &= \frac{1}{K}[1,1,1,\dots,1]\times
    \begin{bmatrix}
                \lambda_0^n &  \lambda_0^n&  \lambda_0^n & \ldots &  \lambda_0^n\\
        \lambda_1^n &  \lambda_1^n\omega_K^* & \lambda_1^n \omega_K^{2*} &\ldots &\lambda_1^n\omega_K^{K-1*}\\
        \lambda_2^n & \lambda_2^n \omega_K^{2*} & \lambda_2^n \omega_K^{4*} & \ldots &\lambda_2^n \omega_K^{2(K-1)*}\\
        \vdots & \vdots &\vdots & \vdots \vdots \vdots & \vdots\\
         \lambda_{K-1}^n &   \lambda_{K-1}^n \omega_K^{(K-1)^*} &   \lambda_{K-1}^n \omega_K^{2(K-1)^*} &\ldots &  \lambda_{K-1}^n \omega_K^{(K-1)(K-1)*}
    \end{bmatrix}
    \end{aligned}
 \end{equation}

 This gives 
 \begin{equation}
 \begin{aligned}
    Q_{pwn}^n[0,j]&=\frac{1}{K}\sum_{i=0}^{K-1} \lambda_{i}^n(\omega_K^{i\times j})^*\\&=\frac{1}{K}\sum_{i=0}^{K-1} \lambda_{i}^n \omega_K^{-i\times j} \\
    &=\frac{1}{K}\sum_{i=0}^{K-1} \omega_K^{-i\times j} \sum_{m=0}^n{n\choose m} \rho^m \omega_{K}^{i\times m} (1-\rho)^{n-m}\\
    &=\sum_{m=0}^n{n\choose m} \rho^m  (1-\rho)^{n-m}\frac{1}{K}\sum_{i=0}^{K-1} \omega_K^{i(m-j)} 
    \end{aligned}
    \end{equation}

    See that, 
    $$\frac{1}{K}\sum_{i=0}^{K-1} \omega_K^{i(m-j)}=
    \left\{\begin{matrix}
1 & if   & m-j\equiv0 \mod K\\
0 &  & otherwise \\
\end{matrix}\right.$$ 

We denote $\frac{1}{K}\sum_{i=0}^{K-1} \omega_K^{i(m-j)}$ by $\delta_{m-j \mod K}.$ This gives 

$$Q_{pwn}^n[0,j]=\sum_{m=0}^n{n\choose m} \rho^m  (1-\rho)^{n-m} \delta_{m-j\mod K}$$.

Consider an example where each node belongs to one of 3 classes and then the first row of circulant matrix $Q_{pwn}^2$, can be calculated as follows. For $j=0$, and $K=3$ and $0\leq m\leq n=2$, $\delta_{m-j\mod K}=1$ only when $m=0$. So, $Q_{pwn}^2[0,0]=(1-\rho)^2$. Similarly $Q_{pwn}^2[0,1]=2\rho(1-\rho)$ and $Q_{pwn}^2[0,2]=\rho^2$.

\section{Proof of Theorem 1}
\label{app:proof_thm1}
\begin{proof}
As $n$ denotes the degree of node, and hence can take only natural numbers, so we will prove all property only for natural values of $n$.
\begin{enumerate}
    \item 

 $r(n)$ and $s(n)$ are increasing function of $n$. $q(2n)$ and $q(2n+1)$ are decreasing function of $n$.

    \textbf{For $r(n):$} Let $m>n$ be two natural numbers, as $(1-\rho)\leq1$ for $\rho \in [0,1]$, so, $(1-\rho)^m<(1-\rho)^n$. Hence,  $1-(1-\rho)^m>1-(1-\rho)^n$. Hence, $r(n)$ is an increasing function of $n$.

    \textbf{For $s(n):$} If $\rho>\frac{K-1}{K}$, then for a single flip the probability with which a node with original label $i$ remains in $i$ is $1-\rho<\frac{1}{K}$. Also, the probability with which it gets reassigned as $K$ is $\frac{\rho}{K-1}>\frac{1}{K}$. This means the probability of moving to any specific class is more than the probability of retaining the label, which completely changes the distribution of the data, and hence is not a desirable situation. From the perspective of label noise, we are interested in only $r<\frac{K-1}{K}$.
    with $\rho<\frac{K-1}{K}$, $\left(1-\frac{K\rho}{K-1}\right)<1$, and proof similar to $s(n)$ follows.

\item $r(n)\geq q(n) \forall n$.
    
    We can see 

$$r(n)=\sum_{k=1}^n {n\choose k} p^k (1-p)^{n-k}\geq \sum_{k=\lceil \frac{n}{2}\rceil}^{n} {n \choose k}p^k(1-p)^{n-k}=q(n)$$
    \\~\\

\item If $\rho<\frac{K-1}{K}$, then $s(n)<\frac{K-1}{K}$ and $s(n)=\frac{K-1}{K}$ iff $\rho=\frac{K-1}{K}$

If $\rho<\frac{K-1}{K}$, then  $\left(1-\frac{K\rho}{K-1}\right)^n<1$, and hence $1-\left(1-\frac{K\rho}{K-1}\right)^n<1$. Which finally means $s(n)<\frac{K-1}{K}.$ 
\\~\\
Now, to prove the second part of the statement, let $\rho=\frac{K-1}{K}$, 

$\Leftrightarrow 
   s(n)=\frac{K-1}{K}\left( 
  1-\left(1-\frac{K\times (K-1)}{(K-1)\times K}\right)^n\right)=\frac{K-1}{K}$
  \end{enumerate}
\end{proof}

\section{Detailed Results}
\subsection{$\rho$ values for various noise levels}
 Ready to refer $\rho$ values for Citeseer, Cora and Amazon Photo datasets for various noise levels are available in  Table \ref{tab:rho}.  We compare different noise models across eight noise-robust algorithms. Many of these algorithms are not computationally feasible for large graphs, so our analysis is limited to relatively smaller graphs. EDN-based noise models assign different flipping probabilities based on node degree, requiring the calculation of distinct probabilities for each degree up to the maximum degree $n$ of the graph. These individual computations are not expensive; therefore, injecting EDN noise remains computationally efficient even for graphs with a large number of nodes. 
\begin{table*}[!h]
\caption{$\rho$ values for different noise levels for Cora, Citeseer and Amazon Photo datasets. }\label{tab:rho}
\centering

\begin{tabular}{c|c|l|l|l|l}
\hline
Dataset & Noise Level & Majority Vote & Veto Power & \begin{tabular}[c]{@{}l@{}}Sequential\\ Flipping+SLN\end{tabular} & \begin{tabular}[c]{@{}l@{}}Sequential\\ Flipping+PW\end{tabular} \\ \hline

 & 5\% & 0.05481096 & 0.01920384 & 0.01940388 & 0.019204 \\
 & 10\% & 0.10622124 & 0.04020804 & 0.0410082 & 0.040208 \\
 & 15\% & 0.15423085 & 0.06321264 & 0.06481296 & 0.063213 \\
 & 20\% & 0.19983997 & 0.0880176 & 0.09121824 & 0.088018 \\
Citeseer & 25\% & 0.24324865 & 0.115023 & 0.12022404 & 0.115223 \\
 & 30\% & 0.28505701 & 0.14422885 & 0.15223045 & 0.144629 \\
 & 35\% & 0.32526505 & 0.17603521 & 0.18783757 & 0.176835 \\
 & 40\% & 0.36467293 & 0.21064213 & 0.22704541 & 0.211842 \\
 & 45\% & 0.4034807 & 0.24824965 & 0.27065413 & 0.25005 \\
 & 50\% & 0.44188838 & 0.28925785 & 0.31926385 & 0.292058 \\ \hline
  & 5\% & 0.07121424 & 0.01360272 & 0.01360272 & 0.013603 \\
 & 10\% & 0.12862573 & 0.02840568 & 0.02880576 & 0.028406 \\
 & 15\% & 0.17803561 & 0.04460892 & 0.04540908 & 0.044609 \\
 & 20\% & 0.22224445 & 0.0620124 & 0.06381276 & 0.062012 \\
Cora & 25\% & 0.26265253 & 0.08081616 & 0.08381676 & 0.081016 \\
 & 30\% & 0.30046009 & 0.10122024 & 0.10582116 & 0.10142 \\
 & 35\% & 0.33626725 & 0.12342468 & 0.13002601 & 0.123625 \\
 & 40\% & 0.37087417 & 0.14742949 & 0.15703141 & 0.14783 \\
 & 45\% & 0.4044809 & 0.17383477 & 0.18723745 & 0.174435 \\
 & 50\% & 0.43768754 & 0.20284057 & 0.22144429 & 0.203841 \\ \hline
 & 5\% & 0.24724945 & 0.00180036 & 0.00180036 & 0.0018 \\
 & 10\% & 0.32086417 & 0.00380076 & 0.00380076 & 0.003801 \\
 & 15\% & 0.36227245 & 0.0060012 & 0.00620124 & 0.006001 \\
 & 20\% & 0.39087818 & 0.00840168 & 0.00880176 & 0.008402 \\
Amazon Photo & 25\% & 0.41308262 & 0.01120224 & 0.01180236 & 0.011402 \\
 & 30\% & 0.43168634 & 0.01440288 & 0.01540308 & 0.014603 \\
 & 35\% & 0.44788958 & 0.01820364 & 0.01960392 & 0.018204 \\
 & 40\% & 0.46269254 & 0.02240448 & 0.02460492 & 0.022404 \\
 & 45\% & 0.47629526 & 0.02720544 & 0.03040608 & 0.027405 \\
 & 50\% & 0.48929786 & 0.0330066 & 0.03780756 & 0.033407 \\ \hline
\end{tabular}

\end{table*}

\subsection{Comparison of EDN with existing noise model}
Comparison of EDN with existing noise models for Citeseer, Cora and Amazon Photo datasets for different GNN architectures and Noise robust algorithms are in Tables \ref{tab:cs}, \ref{tab:cor}, \ref{tab:ap}, \ref{tab:nrl_citeseer}, \ref{tab:nrl_cora}, \ref{tab:nrl_amazonpho}.

\begin{table}[!h]
\caption{Comparison of noise model variants across GNN architecture for Citeseer dataset. Reported values are accuracy$\pm$std of 10 repetitions. }
\label{tab:cs}
\resizebox{\textwidth}{!}{
\begin{tabular}{c|c|llllllll}
 \multicolumn{1}{c|}{\begin{tabular}[c]{@{}c@{}}GNN \\ Architecture\end{tabular}} & \multicolumn{1}{c|}{\begin{tabular}[c]{@{}c@{}}Noise \\ Level\end{tabular}} & \multicolumn{1}{l}{SLN\ \ \ \ \ \ \ \ \ \ \ } & \multicolumn{1}{c}{\begin{tabular}[l]{@{}l@{}}MV+ \\ SLN\ \ \ \ \ \ \ \ \ \ \ \end{tabular}} & \multicolumn{1}{c}{\begin{tabular}[l]{@{}l@{}}Veto+\\ SLN \ \ \ \ \ \ \ \ \ \ \end{tabular}} & \multicolumn{1}{c}{\begin{tabular}[l]{@{}l@{}}Seq+\\ SLN\ \ \ \ \ \ \ \ \ \ \end{tabular}} & \multicolumn{1}{l}{PWN\ \ \ \ \ \ \ \ \ \ } & \multicolumn{1}{c}{\begin{tabular}[l]{@{}l@{}}MV+\\ PWN\ \ \ \ \ \ \ \ \ \ \end{tabular}} & \multicolumn{1}{c}{\begin{tabular}[l]{@{}l@{}}Veto+\\ PWN\ \ \ \ \ \ \ \ \ \ \end{tabular}}  &  \multicolumn{1}{c}{\begin{tabular}[l]{@{}l@{}}Seq+\\ PWN\end{tabular}} \\ \hline
 & 5\% & 73.92$\pm$1.1 & 72.64$\pm$0.5 & 72.27$\pm$0.7 & 72.34$\pm$0.6 & 73.99$\pm$1 & 72.91$\pm$0.6 & 72.24$\pm$0.6 & 72.23$\pm$0.58 \\
 & 10\% & 72.23$\pm$1.4 & 71.47$\pm$1 & 70.39$\pm$0.8 & 70.19$\pm$0.9 & 72.18$\pm$1.2 & 71.35$\pm$1.00 & 69.99$\pm$0.90 & 69.91$\pm$ 0.73 \\
 & 15\% & 70.17$\pm$1.47 & 71.50$\pm$0.99 & 70.42$\pm$0.81 & 70.23$\pm$0.90 & 70.04$\pm$1.51 & 71.36$\pm$1.00 & 69.98$\pm$0.87 & 69.91$\pm$0.72 \\
 & 20\% & 68.09$\pm$1.59 & 67.59$\pm$1.39 & 65.07$\pm$1.12 & 65.02$\pm$1.32 & 67.15$\pm$1.74 & 66.84$\pm$1.45 & 63.04$\pm$1.60 & 62.82$\pm$1.51 \\
GCN & 25\% & 65.55$\pm$1.74 & 65.34$\pm$1.20 & 61.64$\pm$1.43 & 61.79$\pm$1.42 & 64.29$\pm$1.91 & 63.93$\pm$1.22 & 58.42$\pm$1.76 & 58.17$\pm$2.43 \\
 & 30\% & 62.49$\pm$2.06 & 65.34$\pm$1.17 & 61.63$\pm$1.47 & 61.81$\pm$1.41 & 60.43$\pm$2.21 & 63.90$\pm$1.23 & 58.41$\pm$1.76 & 58.18$\pm$ 2.47 \\
 & 35\% & 59.34$\pm$2.76 & 62.38$\pm$1.46 & 57.69$\pm$1.72 & 58.03$\pm$1.50 & 56.44$\pm$2.60 & 60.74$\pm$1.73 & 53.31$\pm$2.30 & 53.45$\pm$2.29 \\
 & 40\% & 56.17$\pm$2.28 & 55.76$\pm$1.70 & 48.50$\pm$2.58 & 49.65$\pm$2.46 & 51.85$\pm$2.21 & 52.48$\pm$1.66 & 42.91$\pm$2.22 & 43.19$\pm$2.41 \\
 & 45\% & 52.47$\pm$2.39 & 52.33$\pm$1.55 & 43.86$\pm$2.21 & 45.57$\pm$2.21 & 47.30$\pm$2.34 & 47.36$\pm$1.25 & 37.93$\pm$2.02 & 38.05$\pm$2.56 \\
 & 50\% & 48.81$\pm$2.58 & 48.33$\pm$1.63 & 38.71$\pm$1.81 & 41.39$\pm$2.54 & 41.89$\pm$2.28 & 41.93$\pm$1.80 & 33.23$\pm$2.08 & 33.39$\pm$2.12 \\ \hline
 & 5\% & 71.92$\pm$4.50 & 69.52$\pm$1.93 & 68.59$\pm$2.76 & 67.76$\pm$3.35 & 67.99$\pm$3.27 & 68.56$\pm$3.01 & 68.78$\pm$2.22 & 68.58$\pm$1.95 \\
 & 10\% & 66.94$\pm$2.70 & 68.34$\pm$2.73 & 65.13$\pm$3.81 & 63.48$\pm$8.73 & 66.69$\pm$4.17 & 66.29$\pm$4.12 & 61.95$\pm$9.74 & 65.52$\pm$2.98 \\
 & 15\% & 66.22$\pm$3.43 & 68.34$\pm$2.73 & 65.13$\pm$3.81 & 63.48$\pm$8.73 & 63.60$\pm$3.12 & 66.86$\pm$3.55 & 62.92$\pm$9.88 & 65.52$\pm$2.98 \\
 & 20\% & 64.82$\pm$3.76 & 64.42$\pm$2.45 & 56.81$\pm$5.62 & 56.84$\pm$4.40 & 59.31$\pm$3.31 & 62.91$\pm$3.28 & 56.77$\pm$5.64 & 53.34$\pm$8.04 \\
GIN & 25\% & 60.12$\pm$5.71 & 63.39$\pm$3.58 & 54.51$\pm$4.01 & 55.42$\pm$6.26 & 54.22$\pm$9.29 & 61.47$\pm$3.51 & 49.34$\pm$7.72 & 49.98$\pm$4.16 \\
 & 30\% & 55.98$\pm$3.94 & 63.39$\pm$3.58 & 53.78$\pm$5.67 & 55.69$\pm$6.13 & 53.14$\pm$6.02 & 61.67$\pm$3.13 & 49.34$\pm$7.72 & 49.98$\pm$4.16 \\
 & 35\% & 55.28$\pm$5.92 & 61.74$\pm$2.73 & 46.11$\pm$5.54 & 49.90$\pm$6.51 & 48.10$\pm$3.36 & 56.05$\pm$5.91 & 45.41$\pm$7.25 & 45.22$\pm$4.81 \\
 & 40\% & 51.04$\pm$6.90 & 54.35$\pm$3.15 & 40.40$\pm$7.99 & 41.89$\pm$6.24 & 44.71$\pm$7.95 & 46.04$\pm$4.67 & 37.49$\pm$5.99 & 36.70$\pm$10.03 \\
 & 45\% & 42.29$\pm$8.80 & 47.87$\pm$7.42 & 37.98$\pm$5.40 & 39.28$\pm$5.53 & 39.19$\pm$3.55 & 46.07$\pm$6.53 & 34.25$\pm$5.86 & 33.50$\pm$1.65 \\
 & 50\% & 42.18$\pm$9.75 & 42.64$\pm$11.5 & 31.57$\pm$3.23 & 33.92$\pm$6.21 & 36.21$\pm$6.22 & 38.58$\pm$5.08 & 28.98$\pm$5.36 & 29.08$\pm$1.983 \\ \hline
 & 5\% & 75.95$\pm$0.96 & 74.96$\pm$0.51 & 74.69$\pm$0.49 & 74.61$\pm$0.62 & 75.97$\pm$0.88 & 75.07$\pm$0.56 & 74.00$\pm$0.53 & 75.69$\pm$1.30 \\
 & 10\% & 74.75$\pm$0.99 & 74.13$\pm$0.82 & 73.53$\pm$0.74 & 73.61$\pm$0.79 & 74.82$\pm$0.99 & 74.20$\pm$0.68 & 72.76$\pm$0.76 & 74.66$\pm$1.65 \\
 & 15\% & 73.87$\pm$1.10 & 74.18$\pm$0.83 & 73.55$\pm$0.76 & 73.57$\pm$0.75 & 73.35$\pm$1.36 & 74.15$\pm$0.72 & 72.71$\pm$0.78 & 74.64$\pm$1.66 \\
 & 20\% & 72.78$\pm$1.21 & 72.20$\pm$1.10 & 70.95$\pm$1.11 & 70.95$\pm$1.03 & 71.45$\pm$1.44 & 70.93$\pm$1.20 & 67.41$\pm$1.19 & 69.74$\pm$1.97 \\
GraphSAGE & 25\% & 70.96$\pm$1.49 & 70.70$\pm$1.07 & 68.42$\pm$1.36 & 68.99$\pm$0.91 & 68.93$\pm$1.92 & 68.33$\pm$1.07 & 63.27$\pm$1.43 & 65.70$\pm$2.16 \\
 & 30\% & 68.82$\pm$1.64 & 70.75$\pm$1.11 & 68.34$\pm$1.29 & 68.97$\pm$0.85 & 65.25$\pm$2.39 & 68.34$\pm$1.14 & 63.33$\pm$1.38 & 65.76$\pm$2.13 \\
 & 35\% & 66.26$\pm$2.10 & 68.77$\pm$1.30 & 66.03$\pm$1.49 & 66.43$\pm$0.83 & 60.78$\pm$3.35 & 64.72$\pm$1.42 & 58.59$\pm$1.77 & 60.93$\pm$2.53 \\
 & 40\% & 63.58$\pm$1.99 & 63.62$\pm$1.31 & 57.76$\pm$1.89 & 58.64$\pm$1.72 & 55.92$\pm$3.13 & 55.35$\pm$1.78 & 47.32$\pm$2.22 & 49.19$\pm$3.59 \\
 & 45\% & 59.61$\pm$1.99 & 59.97$\pm$1.79 & 52.89$\pm$2.11 & 54.17$\pm$2.08 & 49.71$\pm$2.85 & 49.19$\pm$1.56 & 41.19$\pm$2.29 & 44.32$\pm$4.10 \\
 & 50\% & 55.60$\pm$2.01 & 55.31$\pm$2.26 & 47.49$\pm$2.23 & 49.52$\pm$2.86 & 43.99$\pm$2.57 & 42.40$\pm$2.33 & 35.28$\pm$2.33 & 38.75$\pm$3.75 \\ \hline
 & 5\% & 76.46$\pm$1.53 & 74.75$\pm$0.67 & 74.33$\pm$0.89 & 74.36$\pm$0.67 & 76.62$\pm$1.49 & 74.93$\pm$0.85 & 74.39$\pm$0.66 & 76.61$\pm$1.39 \\
 & 10\% & 75.65$\pm$1.56 & 74.29$\pm$0.82 & 73.88$\pm$1.12 & 73.93$\pm$0.88 & 76.22$\pm$1.46 & 74.39$\pm$0.83 & 73.75$\pm$0.74 & 76.19$\pm$1.35 \\
 & 15\% & 75.38$\pm$1.66 & 74.18$\pm$0.77 & 73.84$\pm$1.09 & 73.88$\pm$0.92 & 75.77$\pm$1.45 & 74.36$\pm$0.79 & 73.76$\pm$0.69 & 76.14$\pm$1.46 \\
 & 20\% & 75.10$\pm$1.46 & 73.59$\pm$1.13 & 72.88$\pm$1.38 & 72.77$\pm$1.16 & 74.72$\pm$1.39 & 72.96$\pm$1.10 & 71.84$\pm$0.99 & 74.10$\pm$2.25 \\
GAT & 25\% & 74.59$\pm$1.61 & 73.06$\pm$1.04 & 71.81$\pm$1.06 & 72.16$\pm$1.39 & 73.31$\pm$1.84 & 71.32$\pm$1.50 & 68.13$\pm$1.45 & 71.42$\pm$2.64 \\
 & 30\% & 73.71$\pm$1.60 & 73.15$\pm$1.13 & 71.87$\pm$1.16 & 72.16$\pm$1.40 & 71.04$\pm$2.33 & 71.40$\pm$1.53 & 68.18$\pm$1.41 & 71.44$\pm$2.72 \\
 & 35\% & 73.15$\pm$1.51 & 72.88$\pm$1.50 & 71.47$\pm$1.04 & 71.30$\pm$1.37 & 67.48$\pm$2.75 & 68.86$\pm$2.02 & 64.05$\pm$2.02 & 67.73$\pm$3.77 \\
 & 40\% & 72.47$\pm$1.46 & 71.82$\pm$2.04 & 68.75$\pm$2.25 & 69.10$\pm$1.42 & 62.03$\pm$3.60 & 60.98$\pm$3.01 & 50.85$\pm$2.81 & 54.47$\pm$4.91 \\
 & 45\% & 70.88$\pm$1.64 & 70.45$\pm$1.81 & 66.60$\pm$2.33 & 67.51$\pm$1.87 & 55.00$\pm$3.75 & 53.93$\pm$2.64 & 43.41$\pm$3.18 & 45.97$\pm$6.50 \\
 & 50\% & 70.25$\pm$2.67 & 69.21$\pm$1.72 & 63.27$\pm$3.02 & 65.30$\pm$3.78 & 45.96$\pm$3.47 & 45.14$\pm$3.37 & 35.28$\pm$3.14 & 38.19$\pm$6.87 \\ \hline
 & 5\% & 76.28$\pm$0.86 & 75.64$\pm$0.52 & 76.28$\pm$0.50 & 76.54$\pm$0.54 & 76.10$\pm$1.40 & 75.50$\pm$0.14 & 76.16$\pm$0.35 & 75.50$\pm$0.95 \\
 & 10\% & 74.68$\pm$0.74 & 74.34$\pm$0.63 & 74.92$\pm$0.72 & 74.54$\pm$1.13 & 75.18$\pm$0.34 & 74.56$\pm$0.40 & 74.76$\pm$0.65 & 74.42$\pm$1.24 \\
 & 15\% & 73.78$\pm$1.29 & 74.34$\pm$0.78 & 74.98$\pm$0.64 & 74.56$\pm$0.96 & 73.60$\pm$0.86 & 74.72$\pm$0.54 & 74.64$\pm$0.55 & 74.49$\pm$1.17 \\
 & 20\% & 72.48$\pm$1.80 & 71.52$\pm$0.75 & 71.02$\pm$1.33 & 71.58$\pm$1.28 & 70.92$\pm$0.90 & 70.70$\pm$0.97 & 69.24$\pm$1.16 & 69.16$\pm$1.72 \\
Graph Transformer & 25\% & 70.24$\pm$1.63 & 70.36$\pm$0.60 & 68.84$\pm$0.68 & 68.68$\pm$0.86 & 68.86$\pm$1.03 & 67.48$\pm$1.12 & 64.94$\pm$0.68 & 65.45$\pm$2.00 \\
 & 30\% & 67.68$\pm$1.06 & 70.40$\pm$0.51 & 69.14$\pm$1.06 & 68.56$\pm$0.72 & 64.62$\pm$2.29 & 67.56$\pm$0.99 & 64.92$\pm$0.91 & 65.42$\pm$1.98 \\
 & 35\% & 65.40$\pm$1.00 & 68.06$\pm$1.22 & 66.96$\pm$1.55 & 67.48$\pm$1.55 & 58.90$\pm$1.81 & 63.78$\pm$1.55 & 60.06$\pm$1.62 & 61.23$\pm$2.33 \\
 & 40\% & 61.62$\pm$0.64 & 63.44$\pm$0.81 & 57.02$\pm$0.80 & 58.02$\pm$1.03 & 54.64$\pm$2.01 & 54.70$\pm$2.89 & 47.72$\pm$0.43 & 50.34$\pm$3.17 \\
 & 45\% & 58.10$\pm$1.24 & 59.22$\pm$1.42 & 52.38$\pm$0.60 & 53.58$\pm$0.94 & 49.04$\pm$2.25 & 48.94$\pm$2.43 & 43.00$\pm$1.01 & 44.09$\pm$2.77 \\
 & 50\% & 52.88$\pm$0.35 & 53.68$\pm$2.05 & 46.20$\pm$0.72 & 47.94$\pm$1.25 & 43.32$\pm$0.77 & 42.36$\pm$2.06 & 35.64$\pm$1.04 & 38.58$\pm$2.97 \\ \hline
\end{tabular}}
\end{table}

\begin{table}[!h]
\caption{Comparison of noise model variants across GNN architecture for Cora dataset. Reported values are accuracy$\pm$std of 10 repetitions. }\label{tab:cor}
\resizebox{\textwidth}{!}{
\begin{tabular}{c|c|llllllll}
 \multicolumn{1}{c|}{\begin{tabular}[c]{@{}c@{}}GNN \\ Architecture\end{tabular}} & \multicolumn{1}{c|}{\begin{tabular}[c]{@{}c@{}}Noise \\ Level\end{tabular}} & \multicolumn{1}{l}{SLN\ \ \ \ \ \ \ \ \ \ \ } & \multicolumn{1}{c}{\begin{tabular}[l]{@{}l@{}}MV+ \\ SLN\ \ \ \ \ \ \ \ \ \ \ \end{tabular}} & \multicolumn{1}{c}{\begin{tabular}[l]{@{}l@{}}Veto+\\ SLN \ \ \ \ \ \ \ \ \ \ \end{tabular}} & \multicolumn{1}{c}{\begin{tabular}[l]{@{}l@{}}Seq+\\ SLN\ \ \ \ \ \ \ \ \ \ \end{tabular}} & \multicolumn{1}{l}{PWN\ \ \ \ \ \ \ \ \ \ } & \multicolumn{1}{c}{\begin{tabular}[l]{@{}l@{}}MV+\\ PWN\ \ \ \ \ \ \ \ \ \ \end{tabular}} & \multicolumn{1}{c}{\begin{tabular}[l]{@{}l@{}}Veto+\\ PWN\ \ \ \ \ \ \ \ \ \ \end{tabular}}  &  \multicolumn{1}{c}{\begin{tabular}[l]{@{}l@{}}Seq+\\ PWN\end{tabular}} \\ \hline
 & 5\% & 84.73$\pm$0.94 & 85.00$\pm$0.44 & 85.19$\pm$0.74 & 85.09$\pm$0.73 & 84.27$\pm$0.98 & 85.36$\pm$0.48 & 85.36$\pm$0.50 & 84.58$\pm$0.54 \\
 & 10\% & 83.03$\pm$1.14 & 83.90$\pm$0.69 & 83.46$\pm$0.69 & 83.44$\pm$0.80 & 82.08$\pm$1.53 & 83.60$\pm$1.00 & 82.41$\pm$1.09 & 83.69$\pm$0.76 \\
 & 15\% & 81.08$\pm$1.48 & 83.92$\pm$0.71 & 83.51$\pm$0.63 & 83.47$\pm$0.76 & 79.21$\pm$1.82 & 83.62$\pm$1.00 & 82.40$\pm$1.07 & 83.68$\pm$0.77 \\
 & 20\% & 79.06$\pm$1.46 & 80.38$\pm$1.32 & 78.37$\pm$1.58 & 78.49$\pm$1.47 & 75.76$\pm$1.75 & 78.38$\pm$2.04 & 74.45$\pm$1.68 & 79.89$\pm$1.45 \\
GCN & 25\% & 76.46$\pm$1.48 & 77.46$\pm$1.99 & 74.49$\pm$2.09 & 74.65$\pm$2.03 & 71.97$\pm$1.77 & 74.84$\pm$2.58 & 68.85$\pm$2.30 & 76.35$\pm$1.44 \\
 & 30\% & 73.68$\pm$1.50 & 77.51$\pm$1.99 & 74.50$\pm$2.01 & 74.64$\pm$2.04 & 67.67$\pm$1.98 & 74.90$\pm$2.57 & 68.81$\pm$2.29 & 76.34$\pm$1.48 \\
 & 35\% & 70.41$\pm$1.76 & 74.79$\pm$2.24 & 71.08$\pm$1.97 & 71.36$\pm$1.95 & 62.22$\pm$2.47 & 70.45$\pm$2.73 & 63.18$\pm$2.46 & 71.77$\pm$2.11 \\
 & 40\% & 66.90$\pm$2.24 & 67.84$\pm$2.88 & 62.05$\pm$2.52 & 62.76$\pm$2.53 & 56.91$\pm$2.77 & 60.07$\pm$3.21 & 50.62$\pm$2.73 & 58.63$\pm$3.65 \\
 & 45\% & 62.29$\pm$2.29 & 63.16$\pm$2.94 & 57.12$\pm$2.54 & 58.06$\pm$2.54 & 50.52$\pm$2.76 & 52.86$\pm$3.87 & 44.28$\pm$2.62 & 50.18$\pm$3.92 \\
 & 50\% & 57.76$\pm$2.36 & 57.87$\pm$2.98 & 51.78$\pm$2.76 & 52.64$\pm$2.61 & 44.28$\pm$2.77 & 46.56$\pm$3.77 & 38.45$\pm$2.82 & 39.99$\pm$5.43 \\ \hline
 & 5\% & 80.95$\pm$2.01 & 81.11$\pm$0.88 & 80.45$\pm$2.63 & 82.11$\pm$2.60 & 80.35$\pm$2.38 & 83.34$\pm$1.55 & 80.77$\pm$2.21 & 79.36$\pm$1.99 \\
 & 10\% & 81.81$\pm$2.46 & 81.74$\pm$2.96 & 78.94$\pm$1.74 & 79.86$\pm$2.21 & 80.49$\pm$2.32 & 80.64$\pm$2.21 & 76.93$\pm$2.55 & 77.98$\pm$1.38 \\
 & 15\% & 80.48$\pm$3.17 & 82.24$\pm$2.21 & 78.94$\pm$1.74 & 78.97$\pm$2.68 & 78.35$\pm$3.00 & 80.64$\pm$2.21 & 76.95$\pm$2.56 & 77.98$\pm$1.38 \\
 & 20\% & 76.14$\pm$3.79 & 81.77$\pm$2.53 & 75.71$\pm$4.33 & 76.91$\pm$4.93 & 75.87$\pm$3.46 & 75.86$\pm$4.51 & 71.21$\pm$3.29 & 71.80$\pm$1.37 \\
GIN & 25\% & 79.02$\pm$2.77 & 77.87$\pm$5.65 & 74.90$\pm$2.73 & 76.57$\pm$2.59 & 73.66$\pm$3.74 & 73.32$\pm$4.28 & 64.84$\pm$3.21 & 67.62$\pm$2.51 \\
 & 30\% & 76.63$\pm$4.38 & 77.86$\pm$5.67 & 74.90$\pm$2.73 & 76.57$\pm$2.59 & 68.97$\pm$6.19 & 73.32$\pm$4.28 & 64.46$\pm$3.97 & 67.62$\pm$2.51 \\
 & 35\% & 75.32$\pm$3.96 & 76.85$\pm$7.28 & 74.70$\pm$4.44 & 75.47$\pm$4.78 & 64.09$\pm$4.61 & 68.90$\pm$3.99 & 58.04$\pm$8.27 & 63.14$\pm$3.09 \\
 & 40\% & 73.82$\pm$2.89 & 77.07$\pm$4.45 & 71.50$\pm$5.96 & 69.71$\pm$9.40 & 64.09$\pm$4.41 & 59.54$\pm$5.84 & 51.36$\pm$3.94 & 51.84$\pm$4.15 \\
 & 45\% & 71.02$\pm$4.38 & 75.17$\pm$2.36 & 66.79$\pm$9.06 & 65.95$\pm$11.8 & 48.40$\pm$5.01 & 53.90$\pm$4.37 & 42.55$\pm$10.1 & 48.36$\pm$3.33 \\
 & 50\% & 63.94$\pm$10.7 & 74.26$\pm$3.11 & 59.52$\pm$10.9 & 64.11$\pm$9.59 & 45.16$\pm$5.17 & 45.77$\pm$9.01 & 34.73$\pm$10.6 & 40.46$\pm$3.88 \\ \hline
 & 5\% & 83.10$\pm$1.19 & 83.38$\pm$0.68 & 83.37$\pm$1.18 & 83.41$\pm$1.09 & 82.84$\pm$1.31 & 83.56$\pm$0.46 & 83.22$\pm$1.13 & 84.43$\pm$0.81 \\
 & 10\% & 80.39$\pm$1.51 & 81.55$\pm$0.97 & 79.97$\pm$1.08 & 79.92$\pm$1.01 & 79.54$\pm$1.72 & 81.12$\pm$1.15 & 79.70$\pm$1.24 & 82.3$\pm$0.85 \\
 & 15\% & 77.81$\pm$1.74 & 81.54$\pm$0.97 & 79.95$\pm$1.08 & 79.89$\pm$1.00 & 76.22$\pm$2.20 & 81.09$\pm$1.17 & 79.70$\pm$1.25 & 82.31$\pm$0.96 \\
 & 20\% & 74.34$\pm$1.74 & 75.61$\pm$1.71 & 73.41$\pm$1.87 & 73.50$\pm$1.71 & 72.25$\pm$2.13 & 74.45$\pm$1.84 & 71.64$\pm$1.91 & 75.25$\pm$1.85 \\
GraphSAGE & 25\% & 70.72$\pm$1.85 & 72.39$\pm$2.37 & 68.83$\pm$1.91 & 69.02$\pm$2.12 & 67.69$\pm$2.50 & 69.64$\pm$2.76 & 66.53$\pm$1.93 & 71.6$\pm$2.48 \\
 & 30\% & 67.71$\pm$2.36 & 72.41$\pm$2.39 & 68.89$\pm$1.90 & 69.06$\pm$2.17 & 63.08$\pm$2.71 & 69.67$\pm$2.79 & 66.54$\pm$1.93 & 71.67$\pm$2.53 \\
 & 35\% & 63.46$\pm$1.99 & 67.89$\pm$2.18 & 64.37$\pm$1.93 & 64.76$\pm$1.57 & 57.97$\pm$2.99 & 64.52$\pm$2.94 & 61.74$\pm$2.46 & 66.47$\pm$3.91 \\
 & 40\% & 59.70$\pm$2.91 & 58.83$\pm$2.77 & 55.69$\pm$2.70 & 55.86$\pm$2.80 & 52.71$\pm$3.32 & 54.79$\pm$2.91 & 51.43$\pm$2.56 & 54.14$\pm$5.24 \\
 & 45\% & 55.27$\pm$2.50 & 53.93$\pm$3.23 & 51.14$\pm$2.74 & 51.35$\pm$2.62 & 47.20$\pm$3.15 & 49.40$\pm$3.33 & 46.21$\pm$2.12 & 47.77$\pm$5.17 \\
 & 50\% & 49.93$\pm$2.09 & 49.75$\pm$3.12 & 46.37$\pm$2.51 & 46.75$\pm$2.68 & 41.97$\pm$3.32 & 44.09$\pm$3.29 & 41.15$\pm$2.13 & 41.34$\pm$5.36 \\ \hline
 & 5\% & 79.50$\pm$1.80 & 80.05$\pm$1.02 & 80.39$\pm$1.14 & 80.45$\pm$1.20 & 79.35$\pm$1.67 & 79.43$\pm$1.51 & 79.45$\pm$1.37 & 79.06$\pm$1.5 \\
 & 10\% & 77.63$\pm$2.14 & 77.89$\pm$1.39 & 78.23$\pm$1.32 & 77.89$\pm$1.47 & 76.38$\pm$1.95 & 77.39$\pm$1.21 & 76.11$\pm$2.22 & 76.49$\pm$1.91 \\
 & 15\% & 75.26$\pm$1.76 & 77.71$\pm$1.37 & 77.87$\pm$1.55 & 77.81$\pm$1.66 & 73.47$\pm$2.42 & 77.37$\pm$1.19 & 76.57$\pm$2.09 & 76.35$\pm$2.08 \\
 & 20\% & 72.70$\pm$2.76 & 73.07$\pm$2.48 & 73.87$\pm$1.80 & 73.58$\pm$2.62 & 68.99$\pm$3.07 & 70.27$\pm$2.81 & 69.94$\pm$2.52 & 70.12$\pm$2.43 \\
GAT & 25\% & 70.33$\pm$2.16 & 69.58$\pm$2.69 & 70.77$\pm$2.57 & 70.48$\pm$1.44 & 65.87$\pm$3.48 & 65.64$\pm$2.34 & 64.80$\pm$2.47 & 66.09$\pm$2.37 \\
 & 30\% & 67.73$\pm$2.38 & 69.91$\pm$2.39 & 70.44$\pm$2.82 & 70.01$\pm$1.73 & 61.14$\pm$2.79 & 65.79$\pm$2.57 & 64.93$\pm$2.63 & 66.21$\pm$2.0 \\
 & 35\% & 64.98$\pm$2.73 & 67.51$\pm$2.90 & 68.04$\pm$2.14 & 67.76$\pm$2.31 & 55.89$\pm$2.96 & 62.11$\pm$3.75 & 60.69$\pm$2.28 & 62.76$\pm$3.2 \\
 & 40\% & 61.17$\pm$2.76 & 61.25$\pm$3.38 & 60.39$\pm$3.33 & 60.51$\pm$3.23 & 50.89$\pm$3.38 & 52.90$\pm$3.04 & 50.40$\pm$2.70 & 51.55$\pm$4.5 \\
 & 45\% & 57.46$\pm$3.68 & 57.14$\pm$3.53 & 55.49$\pm$2.94 & 55.56$\pm$3.14 & 44.76$\pm$3.42 & 46.71$\pm$3.98 & 43.82$\pm$3.07 & 45.25$\pm$5.46 \\
 & 50\% & 51.73$\pm$2.34 & 51.24$\pm$3.76 & 51.31$\pm$3.41 & 51.88$\pm$3.22 & 40.33$\pm$3.41 & 40.99$\pm$4.48 & 39.22$\pm$2.83 & 39.62$\pm$4.73 \\ \hline
 & 5\% & 84.42$\pm$0.84 & 84.78$\pm$0.29 & 84.70$\pm$0.41 & 84.64$\pm$0.67 & 84.06$\pm$0.71 & 84.70$\pm$0.14 & 84.96$\pm$0.54 & 84.12$\pm$0.99 \\
 & 10\% & 82.86$\pm$1.11 & 83.84$\pm$0.44 & 83.68$\pm$0.79 & 83.32$\pm$0.70 & 82.52$\pm$1.04 & 83.94$\pm$0.26 & 83.42$\pm$0.99 & 82.25$\pm$1.32 \\
 & 15\% & 81.30$\pm$1.20 & 83.60$\pm$0.32 & 83.28$\pm$0.86 & 83.54$\pm$0.77 & 80.10$\pm$0.97 & 84.14$\pm$0.34 & 83.50$\pm$1.00 & 82.17$\pm$1.28 \\
 & 20\% & 77.56$\pm$0.69 & 79.66$\pm$1.04 & 78.90$\pm$0.81 & 78.94$\pm$0.82 & 76.18$\pm$0.79 & 77.36$\pm$0.69 & 76.70$\pm$2.15 & 75.28$\pm$1.78 \\
Graph Transformer & 25\% & 75.30$\pm$0.91 & 76.56$\pm$0.70 & 77.18$\pm$0.67 & 77.50$\pm$0.87 & 72.12$\pm$1.05 & 72.18$\pm$1.08 & 72.88$\pm$0.95 & 70.33$\pm$2.01 \\
 & 30\% & 73.08$\pm$1.92 & 76.66$\pm$0.86 & 77.56$\pm$0.56 & 77.58$\pm$1.02 & 67.56$\pm$1.87 & 71.96$\pm$1.14 & 73.24$\pm$1.28 & 70.56$\pm$2.0 \\
 & 35\% & 69.66$\pm$0.48 & 74.26$\pm$1.18 & 74.00$\pm$0.81 & 74.74$\pm$0.51 & 61.80$\pm$1.84 & 68.90$\pm$1.51 & 69.20$\pm$1.50 & 65.81$\pm$2.28 \\
 & 40\% & 66.36$\pm$1.57 & 66.36$\pm$1.93 & 65.90$\pm$0.78 & 65.74$\pm$0.42 & 56.54$\pm$1.84 & 57.40$\pm$2.12 & 57.76$\pm$1.69 & 54.38$\pm$2.36 \\
 & 45\% & 61.16$\pm$1.75 & 62.16$\pm$1.86 & 61.32$\pm$1.56 & 60.62$\pm$2.15 & 50.08$\pm$2.15 & 51.32$\pm$0.92 & 51.28$\pm$1.44 & 48.11$\pm$2.4 \\
 & 50\% & 55.70$\pm$2.32 & 55.88$\pm$1.59 & 55.60$\pm$1.59 & 55.78$\pm$2.14 & 43.44$\pm$2.76 & 44.28$\pm$1.50 & 44.86$\pm$1.65 & 41.66$\pm$2.85 \\ \hline
\end{tabular}}
\end{table}

\begin{table}[!h]
\caption{Comparison of noise model variants across GNN architecture for Amazon Photo dataset. Reported values are accuracy$\pm$std of 10 repetitions. } \label{tab:ap}
\resizebox{\textwidth}{!}{
\begin{tabular}{c|c|llllllll}
 \multicolumn{1}{c|}{\begin{tabular}[c]{@{}c@{}}GNN \\ Architecture\end{tabular}} & \multicolumn{1}{c|}{\begin{tabular}[c]{@{}c@{}}Noise \\ Level\end{tabular}} & \multicolumn{1}{l}{SLN\ \ \ \ \ \ \ \ \ \ \ } & \multicolumn{1}{c}{\begin{tabular}[l]{@{}l@{}}MV+ \\ SLN\ \ \ \ \ \ \ \ \ \ \ \end{tabular}} & \multicolumn{1}{c}{\begin{tabular}[l]{@{}l@{}}Veto+\\ SLN \ \ \ \ \ \ \ \ \ \ \end{tabular}} & \multicolumn{1}{c}{\begin{tabular}[l]{@{}l@{}}Seq+\\ SLN\ \ \ \ \ \ \ \ \ \ \end{tabular}} & \multicolumn{1}{l}{PWN\ \ \ \ \ \ \ \ \ \ } & \multicolumn{1}{c}{\begin{tabular}[l]{@{}l@{}}MV+\\ PWN\ \ \ \ \ \ \ \ \ \ \end{tabular}} & \multicolumn{1}{c}{\begin{tabular}[l]{@{}l@{}}Veto+\\ PWN\ \ \ \ \ \ \ \ \ \ \end{tabular}}  &  \multicolumn{1}{c}{\begin{tabular}[l]{@{}l@{}}Seq+\\ PWN\end{tabular}} \\ \hline
 & 5\% & 86.78$\pm$1.43 & 83.65$\pm$6.28 & 85.05$\pm$4.56 & 83.55$\pm$6.18 & 86.66$\pm$1.46 & 84.65$\pm$5.59 & 85.35$\pm$4.21 & 84.85$\pm$2.89 \\
 & 10\% & 86.45$\pm$1.16 & 82.29$\pm$12.38 & 84.14$\pm$3.82 & 83.54$\pm$4.67 & 85.58$\pm$1.32 & 84.38$\pm$5.67 & 83.60$\pm$3.27 & 80.91$\pm$5.22 \\
 & 15\% & 85.49$\pm$1.21 & 82.30$\pm$12.38 & 84.16$\pm$3.85 & 83.55$\pm$4.67 & 83.53$\pm$1.19 & 84.39$\pm$5.69 & 83.58$\pm$3.26 & 80.92$\pm$5.18 \\
 & 20\% & 82.83$\pm$3.66 & 82.86$\pm$6.06 & 83.05$\pm$3.65 & 80.82$\pm$5.00 & 79.72$\pm$4.84 & 77.60$\pm$10.34 & 77.68$\pm$6.76 & 72.0$\pm$6.66 \\
GCN & 25\% & 83.96$\pm$2.99 & 78.28$\pm$9.38 & 81.87$\pm$4.08 & 80.58$\pm$7.92 & 77.00$\pm$5.47 & 74.15$\pm$9.42 & 70.81$\pm$7.69 & 66.82$\pm$8.03 \\
 & 30\% & 81.15$\pm$9.72 & 78.26$\pm$9.39 & 81.88$\pm$4.03 & 80.66$\pm$7.91 & 70.94$\pm$12.13 & 74.16$\pm$9.41 & 70.82$\pm$7.69 & 66.83$\pm$8.05 \\
 & 35\% & 83.01$\pm$3.82 & 76.80$\pm$11.90 & 77.65$\pm$7.65 & 78.01$\pm$9.09 & 62.77$\pm$7.32 & 69.21$\pm$9.21 & 63.43$\pm$4.89 & 61.85$\pm$12.71 \\
 & 40\% & 78.75$\pm$4.02 & 76.00$\pm$10.41 & 71.06$\pm$12.34 & 70.89$\pm$9.41 & 54.20$\pm$7.31 & 56.45$\pm$8.28 & 49.66$\pm$3.14 & 45.15$\pm$6.21 \\
 & 45\% & 73.45$\pm$5.67 & 75.36$\pm$8.82 & 66.92$\pm$11.98 & 67.85$\pm$11.36 & 44.15$\pm$4.85 & 49.57$\pm$7.92 & 40.74$\pm$2.65 & 40.03$\pm$5.61 \\
 & 50\% & 72.71$\pm$4.76 & 71.59$\pm$6.88 & 64.10$\pm$8.61 & 66.44$\pm$7.91 & 36.19$\pm$4.19 & 42.51$\pm$7.34 & 33.15$\pm$3.93 & 33.98$\pm$6.6 \\ \hline
 & 5\% & 80.24$\pm$4.32 & 81.32$\pm$2.95 & 67.86$\pm$17.91 & 64.96$\pm$15.31 & 66.24$\pm$17.36 & 78.02$\pm$3.71 & 77.10$\pm$5.15 & 34.24$\pm$6.73 \\
 & 10\% & 67.88$\pm$12.96 & 76.66$\pm$7.74 & 56.66$\pm$13.40 & 50.70$\pm$20.65 & 69.98$\pm$16.78 & 73.40$\pm$9.67 & 63.00$\pm$15.79 & 35.82$\pm$6.21 \\
 & 15\% & 56.42$\pm$19.08 & 76.76$\pm$7.82 & 56.58$\pm$12.77 & 52.92$\pm$23.29 & 58.96$\pm$13.60 & 73.42$\pm$9.68 & 63.08$\pm$15.79 & 35.82$\pm$6.21 \\
 & 20\% & 50.32$\pm$10.53 & 68.50$\pm$9.65 & 43.04$\pm$10.86 & 31.88$\pm$8.94 & 58.22$\pm$14.88 & 73.46$\pm$6.61 & 48.64$\pm$12.52 & 32.14$\pm$7.05 \\
GIN & 25\% & 50.58$\pm$19.81 & 67.24$\pm$16.88 & 35.04$\pm$2.90 & 33.62$\pm$8.99 & 50.16$\pm$10.96 & 69.38$\pm$14.52 & 41.78$\pm$10.62 & 33.94$\pm$7.74 \\
 & 30\% & 59.58$\pm$15.97 & 68.20$\pm$17.34 & 35.30$\pm$3.04 & 34.12$\pm$9.61 & 57.52$\pm$6.68 & 69.86$\pm$14.56 & 42.00$\pm$10.78 & 33.94$\pm$7.74 \\
 & 35\% & 43.02$\pm$9.76 & 53.28$\pm$13.74 & 32.54$\pm$3.71 & 30.80$\pm$5.63 & 50.94$\pm$19.75 & 54.24$\pm$17.36 & 34.76$\pm$10.25 & 24.00$\pm$5.93 \\
 & 40\% & 38.20$\pm$7.31 & 47.32$\pm$10.51 & 23.80$\pm$6.83 & 25.54$\pm$5.70 & 41.90$\pm$10.70 & 49.56$\pm$7.88 & 30.10$\pm$9.24 & 25.28$\pm$5.34 \\
 & 45\% & 34.38$\pm$7.71 & 40.28$\pm$4.80 & 24.74$\pm$4.46 & 27.04$\pm$7.18 & 33.52$\pm$17.39 & 48.84$\pm$17.31 & 25.14$\pm$3.94 & 22.58$\pm$7.66 \\
 & 50\% & 31.00$\pm$8.10 & 36.98$\pm$8.71 & 21.14$\pm$2.03 & 28.56$\pm$1.88 & 33.00$\pm$12.67 & 36.12$\pm$17.27 & 21.78$\pm$3.86 & 23.58$\pm$2.76 \\ \hline
 & 5\% & 90.56$\pm$0.86 & 90.41$\pm$0.61 & 90.29$\pm$0.82 & 90.20$\pm$0.64 & 90.39$\pm$1.01 & 90.64$\pm$0.50 & 90.29$\pm$0.79 & 89.93$\pm$1.16 \\
 & 10\% & 88.49$\pm$0.84 & 89.26$\pm$0.44 & 89.32$\pm$0.78 & 89.35$\pm$1.14 & 87.76$\pm$1.17 & 89.26$\pm$0.49 & 88.71$\pm$0.88 & 87.6$\pm$1.46 \\
 & 15\% & 85.86$\pm$1.99 & 89.28$\pm$0.43 & 89.32$\pm$0.77 & 89.35$\pm$1.15 & 84.68$\pm$2.27 & 89.25$\pm$0.48 & 88.70$\pm$0.87 & 87.6$\pm$1.48 \\
 & 20\% & 84.10$\pm$1.48 & 85.14$\pm$1.06 & 85.38$\pm$1.30 & 85.42$\pm$1.22 & 81.63$\pm$2.54 & 83.38$\pm$2.46 & 82.79$\pm$2.36 & 79.14$\pm$3.13 \\
GraphSAGE & 25\% & 81.73$\pm$2.29 & 84.06$\pm$1.34 & 82.15$\pm$1.89 & 82.15$\pm$2.09 & 77.96$\pm$2.87 & 80.25$\pm$2.78 & 77.73$\pm$3.13 & 74.08$\pm$4.06 \\
 & 30\% & 77.97$\pm$2.17 & 84.06$\pm$1.33 & 82.15$\pm$1.88 & 82.16$\pm$2.08 & 72.25$\pm$2.84 & 80.04$\pm$2.82 & 77.73$\pm$3.13 & 74.08$\pm$4.06 \\
 & 35\% & 75.72$\pm$1.59 & 80.37$\pm$2.55 & 78.16$\pm$1.69 & 78.29$\pm$2.20 & 65.55$\pm$3.68 & 75.51$\pm$3.75 & 70.94$\pm$3.25 & 66.42$\pm$4.38 \\
 & 40\% & 70.79$\pm$2.14 & 73.37$\pm$1.66 & 66.77$\pm$3.97 & 68.63$\pm$3.74 & 58.99$\pm$2.38 & 63.46$\pm$4.19 & 54.94$\pm$4.07 & 51.58$\pm$4.72 \\
 & 45\% & 65.88$\pm$3.47 & 68.72$\pm$2.34 & 61.36$\pm$4.87 & 62.66$\pm$4.42 & 50.78$\pm$4.01 & 55.60$\pm$4.85 & 47.78$\pm$4.23 & 43.61$\pm$6.1 \\
 & 50\% & 60.79$\pm$3.71 & 64.32$\pm$2.99 & 53.65$\pm$5.30 & 56.09$\pm$5.07 & 44.42$\pm$4.67 & 47.71$\pm$3.86 & 39.13$\pm$4.36 & 36.25$\pm$5.77 \\ \hline
 & 5\% & 78.20$\pm$1.79 & 79.07$\pm$1.89 & 77.75$\pm$1.95 & 77.29$\pm$1.60 & 77.36$\pm$2.45 & 78.79$\pm$1.52 & 76.30$\pm$1.64 & 76.81$\pm$1.79 \\
 & 10\% & 73.70$\pm$2.73 & 75.99$\pm$1.53 & 74.40$\pm$2.02 & 74.36$\pm$1.77 & 73.07$\pm$2.49 & 76.18$\pm$2.21 & 73.34$\pm$2.27 & 72.51$\pm$1.99 \\
 & 15\% & 69.35$\pm$2.71 & 75.97$\pm$1.72 & 74.26$\pm$1.98 & 74.30$\pm$1.79 & 69.57$\pm$2.85 & 76.06$\pm$2.10 & 73.59$\pm$2.80 & 72.67$\pm$2.21 \\
 & 20\% & 66.41$\pm$2.89 & 69.95$\pm$1.86 & 66.82$\pm$1.90 & 66.32$\pm$1.99 & 65.73$\pm$3.21 & 68.28$\pm$2.92 & 65.38$\pm$2.00 & 63.31$\pm$3.42 \\
GAT & 25\% & 63.20$\pm$3.03 & 66.08$\pm$2.52 & 63.13$\pm$2.32 & 62.52$\pm$2.51 & 61.62$\pm$3.64 & 63.45$\pm$3.42 & 61.21$\pm$1.97 & 58.14$\pm$3.82 \\
 & 30\% & 59.29$\pm$3.91 & 65.89$\pm$2.65 & 63.02$\pm$2.24 & 62.88$\pm$2.70 & 56.64$\pm$4.11 & 63.72$\pm$3.37 & 61.14$\pm$2.08 & 58.25$\pm$4.07 \\
 & 35\% & 55.32$\pm$3.37 & 61.61$\pm$3.37 & 59.03$\pm$2.54 & 58.43$\pm$3.53 & 52.45$\pm$3.48 & 58.66$\pm$3.67 & 56.92$\pm$2.47 & 54.62$\pm$3.04 \\
 & 40\% & 51.17$\pm$4.30 & 53.19$\pm$2.79 & 50.46$\pm$2.77 & 50.45$\pm$3.31 & 48.67$\pm$4.03 & 50.72$\pm$3.03 & 47.93$\pm$2.76 & 44.76$\pm$2.35 \\
 & 45\% & 47.79$\pm$3.67 & 48.55$\pm$3.52 & 46.59$\pm$3.19 & 46.70$\pm$3.37 & 45.01$\pm$2.82 & 47.40$\pm$4.53 & 42.69$\pm$3.05 & 40.29$\pm$4.36 \\
 & 50\% & 44.12$\pm$4.35 & 45.68$\pm$2.85 & 41.16$\pm$3.31 & 40.56$\pm$3.59 & 39.93$\pm$2.99 & 43.33$\pm$3.73 & 38.03$\pm$3.27 & 36.2$\pm$3.08 \\ \hline
 & 5\% & 85.57$\pm$0.83 & 80.87$\pm$8.55 & 84.13$\pm$1.88 & 84.25$\pm$1.36 & 85.48$\pm$0.95 & 85.80$\pm$0.72 & 84.32$\pm$1.22 & 84.94$\pm$0.61 \\
 & 10\% & 83.17$\pm$1.89 & 83.60$\pm$1.08 & 81.84$\pm$1.88 & 82.17$\pm$2.25 & 82.45$\pm$1.06 & 83.58$\pm$1.47 & 81.83$\pm$2.44 & 81.27$\pm$1.17 \\
 & 15\% & 79.73$\pm$1.42 & 83.58$\pm$1.03 & 82.15$\pm$1.43 & 81.91$\pm$2.07 & 78.93$\pm$1.54 & 83.65$\pm$1.40 & 81.94$\pm$2.45 & 81.45$\pm$1.05 \\
 & 20\% & 77.52$\pm$2.20 & 78.81$\pm$1.16 & 73.40$\pm$5.65 & 76.66$\pm$2.37 & 74.06$\pm$3.01 & 76.04$\pm$2.76 & 75.76$\pm$2.32 & 74.48$\pm$3.05 \\
Graph Transformer & 25\% & 74.34$\pm$3.35 & 76.04$\pm$2.34 & 71.73$\pm$2.25 & 72.54$\pm$3.45 & 71.27$\pm$4.41 & 73.05$\pm$2.94 & 72.28$\pm$1.72 & 69.8$\pm$1.86 \\
 & 30\% & 69.06$\pm$2.85 & 76.65$\pm$2.27 & 72.27$\pm$1.56 & 74.51$\pm$2.07 & 67.01$\pm$4.39 & 73.39$\pm$2.96 & 72.35$\pm$1.29 & 69.63$\pm$2.03 \\
 & 35\% & 64.24$\pm$2.99 & 72.60$\pm$2.20 & 69.80$\pm$2.56 & 68.63$\pm$2.74 & 62.16$\pm$2.99 & 69.08$\pm$4.08 & 68.73$\pm$2.45 & 65.13$\pm$2.87 \\
 & 40\% & 62.37$\pm$0.45 & 62.00$\pm$3.18 & 63.27$\pm$2.03 & 62.32$\pm$2.40 & 58.16$\pm$3.18 & 55.05$\pm$6.65 & 60.63$\pm$2.46 & 55.33$\pm$4.81 \\
 & 45\% & 58.77$\pm$2.13 & 58.33$\pm$4.13 & 57.54$\pm$2.01 & 57.76$\pm$1.11 & 54.38$\pm$2.85 & 49.03$\pm$6.71 & 54.25$\pm$3.26 & 49.8$\pm$4.64 \\
 & 50\% & 52.70$\pm$1.47 & 51.94$\pm$4.08 & 52.50$\pm$2.53 & 50.93$\pm$6.66 & 45.96$\pm$2.73 & 44.23$\pm$5.22 & 48.08$\pm$2.30 & 44.65$\pm$5.19 \\ \hline
\end{tabular}}
\end{table}

\begin{table}[!h]
\caption{Comparison of noise model variants across graph label noise robust algorithms for Citeseer dataset. Reported values are accuracy$\pm$std of 10 repetitions. }
\label{tab:nrl_citeseer}
\resizebox{\textwidth}{!}{
\begin{tabular}{c|c|llllllll}
 \begin{tabular}[c]{@{}c@{}}Noise\\ Robust \\ Methods\end{tabular} & \multicolumn{1}{c|}{\begin{tabular}[c]{@{}c@{}}Noise \\ Level\end{tabular}} & \multicolumn{1}{l}{SLN\ \ \ \ \ \ \ \ \ \ \ } & \multicolumn{1}{c}{\begin{tabular}[l]{@{}l@{}}MV+ \\ SLN\ \ \ \ \ \ \ \ \ \ \ \end{tabular}} & \multicolumn{1}{c}{\begin{tabular}[l]{@{}l@{}}Veto+\\ SLN \ \ \ \ \ \ \ \ \ \ \end{tabular}} & \multicolumn{1}{c}{\begin{tabular}[l]{@{}l@{}}Seq+\\ SLN\ \ \ \ \ \ \ \ \ \ \end{tabular}} & \multicolumn{1}{l}{PWN\ \ \ \ \ \ \ \ \ \ } & \multicolumn{1}{c}{\begin{tabular}[l]{@{}l@{}}MV+\\ PWN\ \ \ \ \ \ \ \ \ \ \end{tabular}} & \multicolumn{1}{c}{\begin{tabular}[l]{@{}l@{}}Veto+\\ PWN\ \ \ \ \ \ \ \ \ \ \end{tabular}}  &  \multicolumn{1}{c}{\begin{tabular}[l]{@{}l@{}}Seq+\\ PWN\end{tabular}} \\ \hline
 & 5\% & 73.92$\pm$1.07 & 72.64$\pm$0.47 & 72.27$\pm$0.70 & 72.34$\pm$0.58 & 73.99$\pm$0.99 & 72.91$\pm$0.60 & 72.24$\pm$0.63 & 72.23$\pm$0.58 \\
 & 10\% & 72.23$\pm$1.35 & 71.47$\pm$0.97 & 70.39$\pm$0.84 & 70.19$\pm$0.92 & 72.18$\pm$1.18 & 71.35$\pm$1.00 & 69.99$\pm$0.90 & 69.91$\pm$ 0.73 \\
 & 15\% & 70.17$\pm$1.47 & 71.50$\pm$0.99 & 70.42$\pm$0.81 & 70.23$\pm$0.90 & 70.04$\pm$1.51 & 71.36$\pm$1.00 & 69.98$\pm$0.87 & 69.91$\pm$0.72 \\
GCN & 20\% & 68.09$\pm$1.59 & 67.59$\pm$1.39 & 65.07$\pm$1.12 & 65.02$\pm$1.32 & 67.15$\pm$1.74 & 66.84$\pm$1.45 & 63.04$\pm$1.60 & 62.82$\pm$1.51 \\
 & 25\% & 65.55$\pm$1.74 & 65.34$\pm$1.20 & 61.64$\pm$1.43 & 61.79$\pm$1.42 & 64.29$\pm$1.91 & 63.93$\pm$1.22 & 58.42$\pm$1.76 & 58.17$\pm$2.43 \\
 & 30\% & 62.49$\pm$2.06 & 65.34$\pm$1.17 & 61.63$\pm$1.47 & 61.81$\pm$1.41 & 60.43$\pm$2.21 & 63.90$\pm$1.23 & 58.41$\pm$1.76 & 58.18$\pm$ 2.47 \\
 & 35\% & 59.34$\pm$2.76 & 62.38$\pm$1.46 & 57.69$\pm$1.72 & 58.03$\pm$1.50 & 56.44$\pm$2.60 & 60.74$\pm$1.73 & 53.31$\pm$2.30 & 53.45$\pm$2.29 \\
 & 40\% & 56.17$\pm$2.28 & 55.76$\pm$1.70 & 48.50$\pm$2.58 & 49.65$\pm$2.46 & 51.85$\pm$2.21 & 52.48$\pm$1.66 & 42.91$\pm$2.22 & 43.19$\pm$2.41 \\
 & 45\% & 52.47$\pm$2.39 & 52.33$\pm$1.55 & 43.86$\pm$2.21 & 45.57$\pm$2.21 & 47.30$\pm$2.34 & 47.36$\pm$1.25 & 37.93$\pm$2.02 & 38.05$\pm$2.56 \\
 & 50\% & 48.81$\pm$2.58 & 48.33$\pm$1.63 & 38.71$\pm$1.81 & 41.39$\pm$2.54 & 41.89$\pm$2.28 & 41.93$\pm$1.80 & 33.23$\pm$2.08 & 33.39$\pm$2.12 \\ \hline
 & 5\% & 66.46$\pm$2.84 & 65.15$\pm$2.37 & 62.17$\pm$2.41 & 60.28$\pm$3.58 & 64.04$\pm$2.22 & 66.06$\pm$1.90 & 61.92$\pm$8.79 & 58.12$\pm$13.30 \\
 & 10\% & 62.08$\pm$2.91 & 61.30$\pm$4.17 & 55.61$\pm$5.64 & 58.73$\pm$4.48 & 60.92$\pm$4.72 & 63.10$\pm$1.94 & 55.62$\pm$8.30 & 54.98$\pm$6.40 \\
 & 15\% & 59.88$\pm$2.93 & 61.19$\pm$2.74 & 57.88$\pm$5.52 & 57.58$\pm$4.74 & 56.64$\pm$7.73 & 62.80$\pm$2.37 & 58.64$\pm$5.10 & 55.54$\pm$12.57 \\
 & 20\% & 56.10$\pm$3.77 & 57.35$\pm$4.33 & 43.57$\pm$8.98 & 49.19$\pm$5.13 & 52.83$\pm$4.43 & 56.86$\pm$4.57 & 51.18$\pm$4.23 & 54.04$\pm$7.09 \\
DGNN & 25\% & 53.07$\pm$4.92 & 51.18$\pm$10.06 & 45.02$\pm$6.99 & 44.70$\pm$6.52 & 49.78$\pm$5.92 & 53.62$\pm$6.91 & 46.24$\pm$6.07 & 42.28$\pm$12.04 \\
 & 30\% & 46.85$\pm$8.87 & 56.83$\pm$2.33 & 42.69$\pm$7.29 & 43.90$\pm$8.80 & 49.27$\pm$8.32 & 56.55$\pm$3.89 & 41.69$\pm$8.40 & 43.96$\pm$9.00 \\
 & 35\% & 45.47$\pm$6.33 & 49.32$\pm$7.73 & 39.64$\pm$8.64 & 40.65$\pm$5.40 & 40.58$\pm$8.96 & 53.42$\pm$4.70 & 41.64$\pm$7.89 & 44.70$\pm$3.35 \\
 & 40\% & 38.74$\pm$8.89 & 44.01$\pm$7.75 & 31.46$\pm$6.04 & 32.28$\pm$8.11 & 36.76$\pm$7.34 & 44.09$\pm$8.14 & 37.44$\pm$7.27 & 33.96$\pm$7.38 \\
 & 45\% & 41.89$\pm$6.73 & 43.71$\pm$4.72 & 29.48$\pm$6.39 & 27.27$\pm$9.29 & 32.62$\pm$6.84 & 44.71$\pm$5.86 & 32.83$\pm$4.41 & 34.20$\pm$7.13 \\
 & 50\% & 33.25$\pm$8.00 & 39.23$\pm$6.73 & 25.96$\pm$5.42 & 27.17$\pm$5.13 & 29.62$\pm$7.02 & 38.66$\pm$6.20 & 31.26$\pm$4.34 & 27.48$\pm$5.88 \\ \hline
 & 5\% & 76.58$\pm$2.04 & 72.83$\pm$3.60 & 71.34$\pm$5.36 & 71.61$\pm$5.36 & 74.02$\pm$2.20 & 73.60$\pm$1.74 & 71.63$\pm$5.38 & 73.14$\pm$2.17 \\
 & 10\% & 73.19$\pm$5.30 & 72.03$\pm$2.89 & 71.88$\pm$5.81 & 70.48$\pm$6.23 & 72.95$\pm$2.72 & 73.01$\pm$2.05 & 71.47$\pm$3.29 & 71.68$\pm$3.91 \\
 & 15\% & 72.26$\pm$5.23 & 72.03$\pm$2.89 & 71.88$\pm$5.81 & 70.48$\pm$6.23 & 71.10$\pm$4.76 & 72.98$\pm$2.10 & 71.47$\pm$3.29 & 71.68$\pm$3.91 \\
 & 20\% & 70.80$\pm$4.83 & 68.48$\pm$5.68 & 68.15$\pm$4.52 & 66.71$\pm$7.75 & 68.83$\pm$4.37 & 69.45$\pm$3.38 & 66.47$\pm$5.45 & 68.42$\pm$3.21 \\
PIGNN & 25\% & 71.61$\pm$3.67 & 69.62$\pm$3.46 & 66.79$\pm$6.75 & 66.11$\pm$8.15 & 66.19$\pm$5.57 & 68.11$\pm$4.49 & 62.79$\pm$4.89 & 63.62$\pm$7.72 \\
 & 30\% & 71.02$\pm$4.71 & 69.62$\pm$3.46 & 66.79$\pm$6.75 & 66.11$\pm$8.15 & 62.67$\pm$5.16 & 68.10$\pm$4.51 & 62.79$\pm$4.89 & 63.62$\pm$7.72 \\
 & 35\% & 66.62$\pm$5.84 & 64.72$\pm$7.16 & 64.76$\pm$8.13 & 67.66$\pm$5.28 & 57.74$\pm$7.03 & 64.68$\pm$4.45 & 55.45$\pm$7.81 & 59.72$\pm$2.61 \\
 & 40\% & 67.42$\pm$4.07 & 61.92$\pm$8.26 & 57.63$\pm$8.84 & 58.49$\pm$10.92 & 51.33$\pm$6.57 & 56.83$\pm$5.93 & 44.39$\pm$6.45 & 45.52$\pm$10.72 \\
 & 45\% & 60.79$\pm$11.05 & 56.18$\pm$9.03 & 56.32$\pm$3.70 & 58.49$\pm$7.80 & 44.47$\pm$7.21 & 49.97$\pm$7.33 & 38.14$\pm$4.95 & 44.48$\pm$9.20 \\
 & 50\% & 60.54$\pm$7.37 & 57.45$\pm$7.71 & 50.13$\pm$7.12 & 52.20$\pm$9.57 & 40.65$\pm$5.59 & 39.72$\pm$5.44 & 33.82$\pm$5.40 & 40.40$\pm$7.91 \\ \hline
 & 5\% & 72.15$\pm$3.13 & 69.00$\pm$3.88 & 68.33$\pm$1.98 & 67.54$\pm$1.51 & 69.86$\pm$3.23 & 68.05$\pm$2.46 & 68.87$\pm$2.75 & 72.08$\pm$3.06 \\
 & 10\% & 69.37$\pm$2.64 & 67.06$\pm$3.44 & 66.86$\pm$3.28 & 67.05$\pm$2.75 & 68.25$\pm$3.19 & 66.03$\pm$1.75 & 67.41$\pm$3.08 & 68.42$\pm$2.53 \\
 & 15\% & 67.73$\pm$3.53 & 67.06$\pm$3.44 & 66.86$\pm$3.28 & 67.05$\pm$2.75 & 66.06$\pm$2.11 & 66.03$\pm$1.75 & 67.41$\pm$3.08 & 68.42$\pm$2.53 \\
 & 20\% & 66.08$\pm$3.83 & 65.52$\pm$4.33 & 64.77$\pm$4.65 & 64.59$\pm$4.42 & 62.36$\pm$3.81 & 63.95$\pm$3.71 & 62.77$\pm$3.67 & 66.16$\pm$3.22 \\
RNCGLN & 25\% & 65.09$\pm$4.12 & 62.51$\pm$2.37 & 64.76$\pm$4.86 & 62.41$\pm$4.41 & 58.22$\pm$2.80 & 61.01$\pm$2.82 & 56.50$\pm$2.01 & 58.88$\pm$3.38 \\
 & 30\% & 63.75$\pm$4.67 & 62.51$\pm$2.37 & 64.76$\pm$4.86 & 62.41$\pm$4.41 & 55.84$\pm$3.24 & 61.01$\pm$2.82 & 56.50$\pm$2.01 & 58.88$\pm$3.38 \\
 & 35\% & 58.27$\pm$4.05 & 61.52$\pm$3.32 & 59.26$\pm$5.01 & 60.25$\pm$4.37 & 53.61$\pm$2.80 & 58.91$\pm$2.37 & 51.61$\pm$2.90 & 52.02$\pm$1.05 \\
 & 40\% & 57.78$\pm$4.86 & 54.80$\pm$2.29 & 50.66$\pm$5.56 & 50.33$\pm$3.87 & 47.61$\pm$3.65 & 51.35$\pm$2.41 & 41.07$\pm$2.31 & 44.72$\pm$4.01 \\
 & 45\% & 51.68$\pm$5.01 & 51.01$\pm$3.35 & 44.24$\pm$4.52 & 45.94$\pm$4.38 & 41.87$\pm$2.96 & 46.81$\pm$3.73 & 37.95$\pm$3.20 & 40.72$\pm$1.63 \\
 & 50\% & 47.35$\pm$2.63 & 48.06$\pm$3.17 & 42.58$\pm$4.59 & 42.43$\pm$4.39 & 37.87$\pm$2.92 & 40.29$\pm$3.17 & 32.19$\pm$3.05 & 38.62$\pm$4.42 \\ \hline
 & 5\% & 73.98$\pm$4.38 & 74.31$\pm$1.04 & 74.26$\pm$1.53 & 73.95$\pm$1.35 & 74.18$\pm$0.80 & 74.07$\pm$1.15 & 74.12$\pm$0.97 & 75.08$\pm$1.32 \\
 & 10\% & 73.39$\pm$1.50 & 73.57$\pm$1.55 & 74.67$\pm$0.55 & 74.22$\pm$0.90 & 72.50$\pm$2.82 & 73.67$\pm$1.25 & 73.39$\pm$1.27 & 72.37$\pm$1.90 \\
 & 15\% & 72.92$\pm$1.32 & 73.88$\pm$1.69 & 74.75$\pm$0.97 & 74.25$\pm$0.86 & 71.71$\pm$2.08 & 73.77$\pm$1.20 & 73.86$\pm$1.13 & 72.40$\pm$1.98 \\
 & 20\% & 71.28$\pm$2.51 & 73.74$\pm$1.61 & 73.81$\pm$1.17 & 73.66$\pm$1.69 & 69.66$\pm$2.11 & 72.21$\pm$2.05 & 69.41$\pm$2.21 & 71.47$\pm$3.00 \\
RTGNN & 25\% & 72.47$\pm$1.78 & 72.81$\pm$1.55 & 72.82$\pm$1.97 & 71.95$\pm$3.23 & 65.87$\pm$2.81 & 71.07$\pm$2.65 & 66.09$\pm$2.46 & 70.22$\pm$3.38 \\
 & 30\% & 71.77$\pm$1.99 & 73.09$\pm$1.69 & 72.73$\pm$2.21 & 72.00$\pm$3.37 & 62.01$\pm$2.56 & 70.67$\pm$2.37 & 67.34$\pm$2.81 & 69.03$\pm$2.47 \\
 & 35\% & 71.14$\pm$2.49 & 72.77$\pm$1.38 & 72.53$\pm$1.59 & 72.07$\pm$1.54 & 58.38$\pm$3.08 & 68.53$\pm$3.11 & 63.19$\pm$1.92 & 64.50$\pm$1.33 \\
 & 40\% & 72.50$\pm$1.45 & 73.10$\pm$1.29 & 69.53$\pm$3.85 & 71.21$\pm$2.24 & 53.08$\pm$2.38 & 59.80$\pm$4.92 & 51.65$\pm$2.83 & 54.23$\pm$4.27 \\
 & 45\% & 70.25$\pm$1.97 & 71.09$\pm$2.30 & 64.59$\pm$5.95 & 69.16$\pm$2.81 & 42.61$\pm$4.55 & 53.90$\pm$3.99 & 45.36$\pm$3.73 & 54.05$\pm$3.09 \\
 & 50\% & 67.93$\pm$3.38 & 68.45$\pm$3.17 & 61.55$\pm$5.86 & 65.08$\pm$4.54 & 36.68$\pm$4.00 & 44.48$\pm$3.88 & 37.79$\pm$5.15 & 44.22$\pm$8.49 \\ \hline
 & 5\% & 75.76$\pm$0.99 & 74.19$\pm$1.47 & 74.22$\pm$1.19 & 74.14$\pm$1.15 & 71.69$\pm$3.75 & 73.56$\pm$2.35 & 74.53$\pm$1.63 & 76.58$\pm$3.30 \\
 & 10\% & 73.12$\pm$2.92 & 73.30$\pm$1.69 & 73.71$\pm$2.23 & 72.94$\pm$3.14 & 70.34$\pm$4.61 & 73.72$\pm$1.86 & 73.62$\pm$2.95 & 73.44$\pm$2.69 \\
 & 15\% & 72.88$\pm$2.25 & 73.30$\pm$1.69 & 73.71$\pm$2.23 & 72.94$\pm$3.14 & 69.84$\pm$3.25 & 73.72$\pm$1.86 & 73.62$\pm$2.95 & 73.44$\pm$2.69 \\
 & 20\% & 71.97$\pm$3.18 & 73.73$\pm$1.77 & 70.95$\pm$3.54 & 71.44$\pm$2.62 & 69.74$\pm$3.94 & 72.42$\pm$2.80 & 69.37$\pm$5.16 & 66.66$\pm$5.61 \\
NRGNN & 25\% & 73.78$\pm$1.02 & 72.68$\pm$2.15 & 70.86$\pm$4.00 & 70.40$\pm$4.18 & 67.06$\pm$2.90 & 72.08$\pm$2.89 & 66.07$\pm$6.16 & 69.92$\pm$1.67 \\
 & 30\% & 72.02$\pm$2.94 & 72.68$\pm$2.15 & 70.86$\pm$4.00 & 70.40$\pm$4.18 & 64.49$\pm$5.70 & 72.08$\pm$2.89 & 66.07$\pm$6.16 & 69.92$\pm$1.67 \\
 & 35\% & 70.75$\pm$3.18 & 72.57$\pm$2.98 & 71.54$\pm$2.60 & 71.35$\pm$3.99 & 58.90$\pm$5.68 & 71.68$\pm$2.69 & 61.20$\pm$3.92 & 66.36$\pm$5.41 \\
 & 40\% & 68.24$\pm$10.15 & 70.00$\pm$5.80 & 67.28$\pm$5.79 & 68.48$\pm$5.03 & 57.15$\pm$5.37 & 64.95$\pm$6.39 & 48.13$\pm$4.68 & 63.52$\pm$5.09 \\
 & 45\% & 70.80$\pm$2.43 & 70.73$\pm$5.18 & 65.17$\pm$6.89 & 64.09$\pm$8.24 & 46.98$\pm$3.99 & 55.79$\pm$5.34 & 40.30$\pm$5.00 & 49.60$\pm$10.83 \\
 & 50\% & 68.73$\pm$6.74 & 70.84$\pm$3.20 & 65.90$\pm$6.15 & 67.39$\pm$4.83 & 40.58$\pm$6.03 & 43.34$\pm$6.42 & 32.92$\pm$5.02 & 40.16$\pm$7.06 \\ \hline

\end{tabular}}
\end{table}

\begin{table}[!h]
\addtocounter{table}{-1}
\caption{ \textbf{Continued:}  comparison of noise model variants across graph label noise robust algorithms for Citeseer dataset. Reported values are accuracy$\pm$std of 10 repetitions. }
\resizebox{\textwidth}{!}{
\begin{tabular}{c|c|llllllll}
 \begin{tabular}[c]{@{}c@{}}Noise\\ Robust \\ Methods\end{tabular} & \multicolumn{1}{c|}{\begin{tabular}[c]{@{}c@{}}Noise \\ Level\end{tabular}} & \multicolumn{1}{l}{SLN\ \ \ \ \ \ \ \ \ \ \ } & \multicolumn{1}{c}{\begin{tabular}[l]{@{}l@{}}MV+ \\ SLN\ \ \ \ \ \ \ \ \ \ \ \end{tabular}} & \multicolumn{1}{c}{\begin{tabular}[l]{@{}l@{}}Veto+\\ SLN \ \ \ \ \ \ \ \ \ \ \end{tabular}} & \multicolumn{1}{c}{\begin{tabular}[l]{@{}l@{}}Seq+\\ SLN\ \ \ \ \ \ \ \ \ \ \end{tabular}} & \multicolumn{1}{l}{PWN\ \ \ \ \ \ \ \ \ \ } & \multicolumn{1}{c}{\begin{tabular}[l]{@{}l@{}}MV+\\ PWN\ \ \ \ \ \ \ \ \ \ \end{tabular}} & \multicolumn{1}{c}{\begin{tabular}[l]{@{}l@{}}Veto+\\ PWN\ \ \ \ \ \ \ \ \ \ \end{tabular}}  &  \multicolumn{1}{c}{\begin{tabular}[l]{@{}l@{}}Seq+\\ PWN\end{tabular}} \\ \hline
& 5\% & 76.34$\pm$2.41 & 74.74$\pm$1.86 & 74.64$\pm$1.79 & 74.24$\pm$2.41 & 75.09$\pm$1.07 & 74.78$\pm$2.31 & 74.30$\pm$1.48 & 74.88$\pm$3.26 \\
 & 10\% & 74.75$\pm$1.85 & 73.73$\pm$2.19 & 73.90$\pm$2.22 & 74.16$\pm$1.71 & 72.93$\pm$1.98 & 73.47$\pm$2.12 & 72.53$\pm$2.97 & 72.34$\pm$5.26 \\
 & 15\% & 72.65$\pm$3.76 & 74.18$\pm$0.99 & 73.87$\pm$2.05 & 74.04$\pm$1.84 & 68.46$\pm$4.22 & 73.28$\pm$2.26 & 72.27$\pm$3.01 & 72.64$\pm$4.89 \\
 & 20\% & 73.16$\pm$3.11 & 72.61$\pm$3.61 & 71.57$\pm$2.74 & 71.29$\pm$3.06 & 67.57$\pm$4.26 & 72.31$\pm$2.21 & 65.18$\pm$3.82 & 66.80$\pm$2.54 \\
CRGNN & 25\% & 73.22$\pm$1.63 & 70.35$\pm$6.12 & 71.21$\pm$3.16 & 72.50$\pm$1.97 & 65.98$\pm$4.25 & 70.15$\pm$1.75 & 58.29$\pm$9.09 & 65.34$\pm$3.45 \\
 & 30\% & 66.57$\pm$8.42 & 72.85$\pm$1.59 & 71.32$\pm$3.09 & 72.48$\pm$1.96 & 58.30$\pm$3.59 & 69.26$\pm$3.68 & 58.08$\pm$9.21 & 65.18$\pm$3.21 \\
 & 35\% & 70.30$\pm$2.01 & 71.13$\pm$4.37 & 69.48$\pm$3.92 & 69.83$\pm$3.81 & 55.05$\pm$4.78 & 67.03$\pm$2.64 & 52.26$\pm$8.92 & 57.52$\pm$3.56 \\
 & 40\% & 67.04$\pm$8.01 & 70.09$\pm$1.42 & 59.55$\pm$9.25 & 62.90$\pm$11.23 & 51.98$\pm$2.71 & 54.97$\pm$8.59 & 44.16$\pm$7.27 & 50.08$\pm$7.07 \\
 & 45\% & 64.84$\pm$4.24 & 64.16$\pm$8.81 & 57.43$\pm$7.86 & 63.37$\pm$5.93 & 45.19$\pm$3.63 & 52.46$\pm$3.34 & 40.44$\pm$3.00 & 42.56$\pm$6.28 \\
 & 50\% & 58.61$\pm$10.66 & 60.17$\pm$9.69 & 48.89$\pm$9.34 & 56.31$\pm$9.96 & 39.24$\pm$2.95 & 44.35$\pm$7.17 & 31.72$\pm$4.61 & 39.62$\pm$4.48 \\ \hline
& 5\% & 77.80$\pm$0.83 & 74.20$\pm$1.74 & 73.69$\pm$1.85 & 73.11$\pm$3.40 & 72.67$\pm$4.49 & 73.51$\pm$2.24 & 74.21$\pm$1.59 & 76.60$\pm$4.10 \\
 & 10\% & 68.26$\pm$7.61 & 72.25$\pm$4.08 & 73.21$\pm$2.53 & 71.24$\pm$4.68 & 70.31$\pm$5.93 & 73.15$\pm$2.19 & 72.30$\pm$2.85 & 71.56$\pm$5.94 \\
 & 15\% & 69.00$\pm$6.46 & 72.24$\pm$4.10 & 73.21$\pm$2.53 & 71.24$\pm$4.68 & 70.54$\pm$4.92 & 73.15$\pm$2.19 & 72.67$\pm$2.14 & 71.58$\pm$5.86 \\
 & 20\% & 67.20$\pm$9.02 & 72.26$\pm$1.97 & 70.18$\pm$3.51 & 69.19$\pm$2.38 & 65.47$\pm$8.13 & 71.48$\pm$1.92 & 64.10$\pm$4.91 & 67.00$\pm$8.75 \\
CGNN & 25\% & 69.86$\pm$4.46 & 70.54$\pm$3.82 & 69.04$\pm$5.15 & 70.09$\pm$4.03 & 61.63$\pm$9.43 & 67.52$\pm$6.32 & 60.02$\pm$7.17 & 63.08$\pm$9.39 \\
 & 30\% & 68.08$\pm$5.78 & 70.52$\pm$3.80 & 69.04$\pm$5.15 & 70.11$\pm$4.02 & 60.42$\pm$8.46 & 67.50$\pm$6.31 & 60.02$\pm$7.17 & 63.08$\pm$9.39 \\
 & 35\% & 64.80$\pm$8.75 & 67.83$\pm$5.76 & 63.80$\pm$7.97 & 62.36$\pm$8.36 & 51.63$\pm$7.13 & 66.05$\pm$5.01 & 54.83$\pm$3.89 & 57.58$\pm$10.76 \\
 & 40\% & 60.94$\pm$10.82 & 60.79$\pm$7.18 & 53.98$\pm$10.53 & 54.99$\pm$8.00 & 50.20$\pm$8.70 & 56.50$\pm$5.54 & 46.61$\pm$3.00 & 47.64$\pm$7.87 \\
 & 45\% & 58.62$\pm$8.03 & 55.96$\pm$11.66 & 50.57$\pm$11.89 & 50.62$\pm$10.95 & 44.70$\pm$6.39 & 50.49$\pm$6.26 & 41.45$\pm$3.50 & 46.02$\pm$4.30 \\
 & 50\% & 50.79$\pm$7.70 & 53.76$\pm$10.94 & 44.16$\pm$8.14 & 46.03$\pm$7.97 & 41.16$\pm$3.85 & 43.77$\pm$4.04 & 35.67$\pm$4.35 & 40.00$\pm$5.58 \\ \hline
 & 5\% & 77.58$\pm$1.10 & 77.36$\pm$1.24 & 77.64$\pm$1.71 & 77.50$\pm$1.80 & 77.92$\pm$1.24 & 77.34$\pm$1.41 & 77.64$\pm$1.42 & 77.74$\pm$1.2 \\
 & 10\% & 76.92$\pm$1.46 & 76.88$\pm$1.31 & 78.04$\pm$1.57 & 77.54$\pm$1.80 & 77.34$\pm$1.10 & 77.24$\pm$1.88 & 77.06$\pm$1.83 & 77.44$\pm$1.21  \\
 & 15\% & 77.80$\pm$1.72 & 76.86$\pm$1.32 & 77.80$\pm$1.74 & 77.58$\pm$1.78 & 76.54$\pm$1.72 & 77.18$\pm$1.90 & 77.02$\pm$1.82 &  77.32$\pm$1.14\\
 & 20\% & 77.02$\pm$1.64 & 76.30$\pm$1.06 & 76.52$\pm$1.59 & 76.78$\pm$2.21 & 75.78$\pm$1.58 & 76.70$\pm$2.21 & 75.30$\pm$2.70 & 75.42$\pm$2.27 \\
DeGLIF & 25\% & 76.10$\pm$1.21 & 76.24$\pm$1.48 & 75.86$\pm$2.28 & 76.04$\pm$1.96 & 74.86$\pm$2.20 & 75.94$\pm$2.44 & 73.40$\pm$1.72 & 74.4$\pm$2.95 \\
 & 30\% & 75.04$\pm$1.61 & 76.22$\pm$1.49 & 75.80$\pm$2.17 & 76.18$\pm$1.92 & 73.54$\pm$2.07 & 75.98$\pm$2.45 & 73.38$\pm$1.75 & 74.42$\pm$3.12  \\
 & 35\% & 74.06$\pm$1.45 & 75.66$\pm$0.82 & 75.12$\pm$1.77 & 75.40$\pm$2.05 & 70.54$\pm$2.22 & 73.82$\pm$2.06 & 69.62$\pm$3.61 & 71.98$\pm$3.28 \\
 & 40\% & 73.18$\pm$1.69 & 74.52$\pm$1.61 & 72.80$\pm$2.67 & 73.06$\pm$2.76 & 67.12$\pm$2.05 & 68.54$\pm$4.28 & 56.60$\pm$3.14 & 59.06$\pm$2.17 \\
 & 45\% & 70.58$\pm$1.44 & 72.38$\pm$2.06 & 71.18$\pm$2.99 & 71.98$\pm$2.90 & 59.48$\pm$3.76 & 62.02$\pm$3.25 & 47$\pm$5.34 & 52.48$\pm$8.28 \\
 & 50\% & 67.38$\pm$0.88 & 70.44$\pm$2.13 & 68.34$\pm$1.96 & 69.36$\pm$1.85 & 51.76$\pm$2.95 & 52.2$\pm$4.95 & 37.02$\pm$2.47 & 45.36$\pm$7.48 \\ \hline
\end{tabular}}
\end{table}

\begin{table}[!h]
\caption{Comparison of noise model variants across graph label noise robust algorithms for the Cora dataset. Reported values are accuracy$\pm$std of 10 repetitions. }\label{tab:nrl_cora}
\resizebox{\textwidth}{!}{
\begin{tabular}{c|c|llllllll}
 \begin{tabular}[c]{@{}c@{}}Noise\\ Robust \\ Methods\end{tabular} & \multicolumn{1}{c|}{\begin{tabular}[c]{@{}c@{}}Noise \\ Level\end{tabular}} & \multicolumn{1}{l}{SLN\ \ \ \ \ \ \ \ \ \ \ } & \multicolumn{1}{c}{\begin{tabular}[l]{@{}l@{}}MV+ \\ SLN\ \ \ \ \ \ \ \ \ \ \ \end{tabular}} & \multicolumn{1}{c}{\begin{tabular}[l]{@{}l@{}}Veto+\\ SLN \ \ \ \ \ \ \ \ \ \ \end{tabular}} & \multicolumn{1}{c}{\begin{tabular}[l]{@{}l@{}}Seq+\\ SLN\ \ \ \ \ \ \ \ \ \ \end{tabular}} & \multicolumn{1}{l}{PWN\ \ \ \ \ \ \ \ \ \ } & \multicolumn{1}{c}{\begin{tabular}[l]{@{}l@{}}MV+\\ PWN\ \ \ \ \ \ \ \ \ \ \end{tabular}} & \multicolumn{1}{c}{\begin{tabular}[l]{@{}l@{}}Veto+\\ PWN\ \ \ \ \ \ \ \ \ \ \end{tabular}}  &  \multicolumn{1}{c}{\begin{tabular}[l]{@{}l@{}}Seq+\\ PWN\end{tabular}} \\ \hline
 & 5\% & 84.73$\pm$0.94 & 85.00$\pm$0.44 & 85.19$\pm$0.74 & 85.09$\pm$0.73 & 84.27$\pm$0.98 & 85.36$\pm$0.48 & 85.36$\pm$0.50 & 84.58$\pm$0.54 \\
 & 10\% & 83.03$\pm$1.14 & 83.90$\pm$0.69 & 83.46$\pm$0.69 & 83.44$\pm$0.80 & 82.08$\pm$1.53 & 83.60$\pm$1.00 & 82.41$\pm$1.09 & 83.69$\pm$0.76 \\
 & 15\% & 81.08$\pm$1.48 & 83.92$\pm$0.71 & 83.51$\pm$0.63 & 83.47$\pm$0.76 & 79.21$\pm$1.82 & 83.62$\pm$1.00 & 82.40$\pm$1.07 & 83.68$\pm$0.77 \\
 & 20\% & 79.06$\pm$1.46 & 80.38$\pm$1.32 & 78.37$\pm$1.58 & 78.49$\pm$1.47 & 75.76$\pm$1.75 & 78.38$\pm$2.04 & 74.45$\pm$1.68 & 79.89$\pm$1.45 \\
GCN & 25\% & 76.46$\pm$1.48 & 77.46$\pm$1.99 & 74.49$\pm$2.09 & 74.65$\pm$2.03 & 71.97$\pm$1.77 & 74.84$\pm$2.58 & 68.85$\pm$2.30 & 76.35$\pm$1.44 \\
 & 30\% & 73.68$\pm$1.50 & 77.51$\pm$1.99 & 74.50$\pm$2.01 & 74.64$\pm$2.04 & 67.67$\pm$1.98 & 74.90$\pm$2.57 & 68.81$\pm$2.29 & 76.34$\pm$1.48 \\
 & 35\% & 70.41$\pm$1.76 & 74.79$\pm$2.24 & 71.08$\pm$1.97 & 71.36$\pm$1.95 & 62.22$\pm$2.47 & 70.45$\pm$2.73 & 63.18$\pm$2.46 & 71.77$\pm$2.11 \\
 & 40\% & 66.90$\pm$2.24 & 67.84$\pm$2.88 & 62.05$\pm$2.52 & 62.76$\pm$2.53 & 56.91$\pm$2.77 & 60.07$\pm$3.21 & 50.62$\pm$2.73 & 58.63$\pm$3.65 \\
 & 45\% & 62.29$\pm$2.29 & 63.16$\pm$2.94 & 57.12$\pm$2.54 & 58.06$\pm$2.54 & 50.52$\pm$2.76 & 52.86$\pm$3.87 & 44.28$\pm$2.62 & 50.18$\pm$3.92 \\
 & 50\% & 57.76$\pm$2.36 & 57.87$\pm$2.98 & 51.78$\pm$2.76 & 52.64$\pm$2.61 & 44.28$\pm$2.77 & 46.56$\pm$3.77 & 38.45$\pm$2.82 & 39.99$\pm$5.43 \\ \hline
 & 5\% & 78.32$\pm$4.81 & 82.68$\pm$2.81 & 79.80$\pm$1.65 & 78.08$\pm$4.58 & 74.66$\pm$9.15 & 82.50$\pm$0.95 & 78.38$\pm$7.58 & 79.18$\pm$1.75 \\
 & 10\% & 78.80$\pm$5.07 & 81.06$\pm$2.48 & 68.90$\pm$11.77 & 70.38$\pm$9.41 & 76.74$\pm$2.93 & 82.14$\pm$0.84 & 70.94$\pm$8.00 & 77.96$\pm$2.94 \\
 & 15\% & 72.60$\pm$12.99 & 81.04$\pm$2.57 & 67.34$\pm$13.42 & 70.28$\pm$9.06 & 72.40$\pm$4.04 & 82.22$\pm$0.91 & 74.00$\pm$7.08 & 75.76$\pm$2.94 \\
 & 20\% & 77.08$\pm$4.10 & 80.66$\pm$2.09 & 70.38$\pm$9.93 & 67.88$\pm$8.07 & 67.66$\pm$8.22 & 78.82$\pm$2.40 & 63.24$\pm$10.91 & 64.12$\pm$11.15 \\
DGNN & 25\% & 74.90$\pm$5.87 & 79.38$\pm$2.39 & 65.20$\pm$11.22 & 65.38$\pm$13.62 & 61.78$\pm$9.81 & 77.36$\pm$3.91 & 53.28$\pm$12.29 & 66.54$\pm$7.27 \\
 & 30\% & 60.32$\pm$21.55 & 79.54$\pm$2.03 & 64.58$\pm$11.17 & 64.74$\pm$14.17 & 61.82$\pm$11.15 & 77.14$\pm$3.63 & 48.86$\pm$11.79 & 63.24$\pm$7.41 \\
 & 35\% & 59.96$\pm$16.53 & 74.50$\pm$8.68 & 55.26$\pm$18.68 & 61.84$\pm$8.26 & 53.28$\pm$10.14 & 77.08$\pm$4.68 & 49.30$\pm$17.08 & 63.06$\pm$3.70 \\
 & 40\% & 65.90$\pm$15.86 & 70.68$\pm$7.12 & 54.36$\pm$9.28 & 52.16$\pm$14.51 & 45.02$\pm$10.60 & 58.94$\pm$9.51 & 38.84$\pm$8.29 & 48.66$\pm$8.05 \\
 & 45\% & 62.34$\pm$12.99 & 60.08$\pm$15.94 & 40.26$\pm$8.56 & 40.10$\pm$8.99 & 46.64$\pm$11.84 & 45.02$\pm$11.65 & 30.94$\pm$6.26 & 52.36$\pm$8.54 \\
 & 50\% & 64.62$\pm$8.05 & 56.08$\pm$15.76 & 38.42$\pm$9.73 & 40.52$\pm$9.75 & 33.98$\pm$7.06 & 33.96$\pm$5.16 & 21.48$\pm$5.34 & 39.64$\pm$6.61 \\ \hline
 & 5\% & \multicolumn{1}{l}{80.19$\pm$3.16} & \multicolumn{1}{l}{81.48$\pm$2.16} & \multicolumn{1}{l}{81.81$\pm$2.22} & \multicolumn{1}{l}{81.71$\pm$2.15} & \multicolumn{1}{l}{80.05$\pm$3.50} & \multicolumn{1}{l}{81.58$\pm$1.97} & \multicolumn{1}{l}{81.79$\pm$1.88} & 81.48$\pm$2.84 \\
 & 10\% & \multicolumn{1}{l}{80.93$\pm$2.77} & \multicolumn{1}{l}{81.98$\pm$1.80} & \multicolumn{1}{l}{81.22$\pm$1.94} & \multicolumn{1}{l}{80.81$\pm$2.27} & \multicolumn{1}{l}{78.07$\pm$3.30} & \multicolumn{1}{l}{81.33$\pm$1.47} & \multicolumn{1}{l}{80.81$\pm$3.26} & 80.98$\pm$2.52 \\
 & 15\% & \multicolumn{1}{l}{81.82$\pm$1.80} & \multicolumn{1}{l}{81.98$\pm$1.80} & \multicolumn{1}{l}{81.22$\pm$1.94} & \multicolumn{1}{l}{80.81$\pm$2.27} & \multicolumn{1}{l}{80.77$\pm$2.25} & \multicolumn{1}{l}{81.34$\pm$1.49} & \multicolumn{1}{l}{80.81$\pm$3.26} & 81.02$\pm$2.55 \\
 & 20\% & \multicolumn{1}{l}{80.02$\pm$2.33} & \multicolumn{1}{l}{80.88$\pm$1.66} & \multicolumn{1}{l}{80.56$\pm$2.63} & \multicolumn{1}{l}{80.27$\pm$2.77} & \multicolumn{1}{l}{79.40$\pm$3.52} & \multicolumn{1}{l}{77.21$\pm$1.62} & \multicolumn{1}{l}{75.22$\pm$5.47} & 75.94$\pm$9.03 \\
PIGNN & 25\% & \multicolumn{1}{l}{79.84$\pm$2.30} & \multicolumn{1}{l}{80.11$\pm$2.09} & \multicolumn{1}{l}{80.66$\pm$2.37} & \multicolumn{1}{l}{78.41$\pm$5.17} & \multicolumn{1}{l}{76.54$\pm$3.43} & \multicolumn{1}{l}{75.71$\pm$3.57} & \multicolumn{1}{l}{72.85$\pm$3.56} & 72.87$\pm$8.57 \\
 & 30\% & \multicolumn{1}{l}{79.65$\pm$2.88} & \multicolumn{1}{l}{80.10$\pm$2.08} & \multicolumn{1}{l}{80.66$\pm$2.37} & \multicolumn{1}{l}{78.41$\pm$5.17} & \multicolumn{1}{l}{70.53$\pm$6.29} & \multicolumn{1}{l}{75.71$\pm$3.57} & \multicolumn{1}{l}{72.86$\pm$3.58} & 72.87$\pm$8.57 \\
 & 35\% & \multicolumn{1}{l}{77.91$\pm$2.99} & \multicolumn{1}{l}{80.16$\pm$1.44} & \multicolumn{1}{l}{76.12$\pm$4.32} & \multicolumn{1}{l}{77.76$\pm$3.50} & \multicolumn{1}{l}{68.97$\pm$6.49} & \multicolumn{1}{l}{70.50$\pm$3.60} & \multicolumn{1}{l}{67.43$\pm$7.27} & 69.02$\pm$10.41 \\
 & 40\% & \multicolumn{1}{l}{77.50$\pm$3.33} & \multicolumn{1}{l}{77.57$\pm$2.22} & \multicolumn{1}{l}{76.80$\pm$3.13} & \multicolumn{1}{l}{76.87$\pm$3.57} & \multicolumn{1}{l}{62.53$\pm$4.75} & \multicolumn{1}{l}{58.87$\pm$7.18} & \multicolumn{1}{l}{53.01$\pm$6.52} & 59.01$\pm$10.26 \\
 & 45\% & \multicolumn{1}{l}{75.68$\pm$2.69} & \multicolumn{1}{l}{75.74$\pm$2.92} & \multicolumn{1}{l}{74.95$\pm$6.39} & \multicolumn{1}{l}{74.61$\pm$5.72} & \multicolumn{1}{l}{49.64$\pm$10.92} & \multicolumn{1}{l}{46.89$\pm$9.64} & \multicolumn{1}{l}{47.60$\pm$7.82} & 57.05$\pm$10.22 \\
 & 50\% & \multicolumn{1}{l}{72.63$\pm$7.50} & \multicolumn{1}{l}{72.11$\pm$5.20} & \multicolumn{1}{l}{69.66$\pm$8.65} & \multicolumn{1}{l}{73.30$\pm$4.55} & \multicolumn{1}{l}{45.71$\pm$8.38} & \multicolumn{1}{l}{38.17$\pm$4.90} & \multicolumn{1}{l}{37.64$\pm$7.69} & 45.66$\pm$11.34 \\ \hline
 & 5\% & 83.82$\pm$5.00 & 84.06$\pm$4.05 & 85.32$\pm$2.00 & 87.04$\pm$2.01 & 83.78$\pm$2.71 & 82.08$\pm$3.42 & 88.32$\pm$2.02 & 81.38$\pm$0.61 \\
 & 10\% & 85.52$\pm$1.86 & 83.14$\pm$4.53 & 80.24$\pm$5.90 & 83.90$\pm$1.57 & 81.86$\pm$4.17 & 84.46$\pm$4.02 & 83.10$\pm$2.46 & 77.46$\pm$1.97 \\
 & 15\% & 84.50$\pm$1.95 & 83.14$\pm$4.53 & 80.24$\pm$5.90 & 83.90$\pm$1.57 & 82.04$\pm$1.87 & 84.46$\pm$4.02 & 83.10$\pm$2.46 & 77.46$\pm$1.97 \\
 & 20\% & 81.24$\pm$3.62 & 85.02$\pm$0.80 & 77.46$\pm$7.44 & 76.52$\pm$4.67 & 75.88$\pm$5.39 & 83.08$\pm$5.55 & 75.50$\pm$7.64 & 74.16$\pm$1.98 \\
RNCGLN & 25\% & 80.26$\pm$5.05 & 80.24$\pm$3.99 & 71.54$\pm$6.55 & 70.50$\pm$4.53 & 71.94$\pm$5.37 & 77.00$\pm$9.12 & 64.98$\pm$3.93 & 70.60$\pm$1.39 \\
 & 30\% & 79.46$\pm$4.74 & 80.24$\pm$3.99 & 71.54$\pm$6.55 & 70.50$\pm$4.53 & 70.36$\pm$10.52 & 77.00$\pm$9.12 & 64.98$\pm$3.93 & 70.60$\pm$1.39 \\
 & 35\% & 64.16$\pm$6.37 & 70.26$\pm$4.10 & 68.54$\pm$2.74 & 70.34$\pm$8.76 & 66.32$\pm$8.12 & 70.04$\pm$11.88 & 59.96$\pm$4.31 & 66.80$\pm$4.23 \\
 & 40\% & 67.42$\pm$9.70 & 67.14$\pm$7.35 & 55.20$\pm$5.68 & 62.42$\pm$9.45 & 58.42$\pm$7.35 & 60.54$\pm$7.37 & 49.44$\pm$4.91 & 56.90$\pm$3.12 \\
 & 45\% & 56.52$\pm$5.79 & 61.22$\pm$6.57 & 51.68$\pm$5.66 & 55.40$\pm$5.21 & 52.40$\pm$8.32 & 63.90$\pm$8.12 & 42.58$\pm$2.93 & 49.72$\pm$0.79 \\
 & 50\% & 60.66$\pm$14.26 & 53.28$\pm$4.08 & 53.68$\pm$12.53 & 45.14$\pm$1.67 & 44.48$\pm$5.88 & 45.48$\pm$4.18 & 38.28$\pm$5.98 & 47.76$\pm$4.98 \\ \hline
 & 5\% & \multicolumn{1}{l}{73.54$\pm$2.02} & \multicolumn{1}{l}{74.42$\pm$2.12} & \multicolumn{1}{l}{74.64$\pm$1.78} & \multicolumn{1}{l}{75.04$\pm$1.70} & \multicolumn{1}{l}{75.29$\pm$3.09} & \multicolumn{1}{l}{73.91$\pm$2.63} & \multicolumn{1}{l}{74.28$\pm$1.90} & 73.95$\pm$1.58 \\
 & 10\% & \multicolumn{1}{l}{75.24$\pm$2.98} & \multicolumn{1}{l}{72.75$\pm$2.22} & \multicolumn{1}{l}{74.30$\pm$2.38} & \multicolumn{1}{l}{75.25$\pm$1.82} & \multicolumn{1}{l}{77.11$\pm$1.63} & \multicolumn{1}{l}{71.20$\pm$2.03} & \multicolumn{1}{l}{73.72$\pm$2.28} & 73.95$\pm$1.80 \\
 & 15\% & \multicolumn{1}{l}{75.59$\pm$1.95} & \multicolumn{1}{l}{72.70$\pm$2.28} & \multicolumn{1}{l}{74.23$\pm$2.38} & \multicolumn{1}{l}{75.35$\pm$1.99} & \multicolumn{1}{l}{75.15$\pm$0.96} & \multicolumn{1}{l}{71.23$\pm$2.04} & \multicolumn{1}{l}{73.78$\pm$2.33} & 73.87$\pm$1.93 \\
 & 20\% & \multicolumn{1}{l}{75.29$\pm$2.52} & \multicolumn{1}{l}{73.84$\pm$2.19} & \multicolumn{1}{l}{73.11$\pm$5.47} & \multicolumn{1}{l}{73.68$\pm$4.54} & \multicolumn{1}{l}{75.28$\pm$3.65} & \multicolumn{1}{l}{62.60$\pm$7.62} & \multicolumn{1}{l}{67.79$\pm$3.90} & 72.12$\pm$2.46 \\
RTGNN & 25\% & \multicolumn{1}{l}{75.67$\pm$3.00} & \multicolumn{1}{l}{72.20$\pm$3.70} & \multicolumn{1}{l}{74.12$\pm$3.52} & \multicolumn{1}{l}{73.16$\pm$5.07} & \multicolumn{1}{l}{70.44$\pm$3.29} & \multicolumn{1}{l}{62.89$\pm$6.02} & \multicolumn{1}{l}{55.39$\pm$9.03} & 68.75$\pm$4.56 \\
 & 30\% & \multicolumn{1}{l}{73.24$\pm$5.53} & \multicolumn{1}{l}{72.52$\pm$3.72} & \multicolumn{1}{l}{74.15$\pm$3.52} & \multicolumn{1}{l}{73.32$\pm$5.18} & \multicolumn{1}{l}{68.93$\pm$5.97} & \multicolumn{1}{l}{63.38$\pm$6.40} & \multicolumn{1}{l}{55.44$\pm$8.99} & 63.31$\pm$8.59 \\
 & 35\% & \multicolumn{1}{l}{68.48$\pm$6.92} & \multicolumn{1}{l}{72.84$\pm$3.32} & \multicolumn{1}{l}{72.57$\pm$4.43} & \multicolumn{1}{l}{72.57$\pm$4.80} & \multicolumn{1}{l}{64.33$\pm$6.31} & \multicolumn{1}{l}{62.73$\pm$5.18} & \multicolumn{1}{l}{52.42$\pm$8.65} & 56.78$\pm$7.72 \\
 & 40\% & \multicolumn{1}{l}{72.08$\pm$2.58} & \multicolumn{1}{l}{70.66$\pm$4.66} & \multicolumn{1}{l}{69.83$\pm$4.28} & \multicolumn{1}{l}{69.11$\pm$8.94} & \multicolumn{1}{l}{58.00$\pm$10.09} & \multicolumn{1}{l}{51.37$\pm$7.00} & \multicolumn{1}{l}{45.84$\pm$9.49} & 49.20$\pm$6.39 \\
 & 45\% & \multicolumn{1}{l}{66.44$\pm$7.61} & \multicolumn{1}{l}{64.95$\pm$8.76} & \multicolumn{1}{l}{68.27$\pm$6.98} & \multicolumn{1}{l}{71.45$\pm$5.01} & \multicolumn{1}{l}{57.38$\pm$6.77} & \multicolumn{1}{l}{51.81$\pm$3.87} & \multicolumn{1}{l}{40.51$\pm$8.34} & 42.77$\pm$7.32 \\
 & 50\% & \multicolumn{1}{l}{66.67$\pm$3.96} & \multicolumn{1}{l}{64.58$\pm$4.74} & \multicolumn{1}{l}{68.03$\pm$8.16} & \multicolumn{1}{l}{63.99$\pm$8.89} & \multicolumn{1}{l}{52.10$\pm$12.26} & \multicolumn{1}{l}{46.72$\pm$5.49} & \multicolumn{1}{l}{33.64$\pm$7.12} & 42.53$\pm$10.47 \\ \hline
 & 5\% & \multicolumn{1}{l}{75.13$\pm$3.77} & \multicolumn{1}{l}{75.47$\pm$1.73} & \multicolumn{1}{l}{75.65$\pm$2.62} & \multicolumn{1}{l}{73.19$\pm$2.98} & \multicolumn{1}{l}{75.40$\pm$2.21} & \multicolumn{1}{l}{74.81$\pm$1.82} & \multicolumn{1}{l}{76.12$\pm$2.36} & 74.99$\pm$2.20 \\
 & 10\% & \multicolumn{1}{l}{76.62$\pm$2.74} & \multicolumn{1}{l}{73.07$\pm$3.09} & \multicolumn{1}{l}{76.20$\pm$2.56} & \multicolumn{1}{l}{76.10$\pm$3.08} & \multicolumn{1}{l}{74.44$\pm$3.36} & \multicolumn{1}{l}{72.67$\pm$3.21} & \multicolumn{1}{l}{75.76$\pm$4.24} & 75.28$\pm$2.29 \\
 & 15\% & \multicolumn{1}{l}{76.62$\pm$2.74} & \multicolumn{1}{l}{73.07$\pm$3.09} & \multicolumn{1}{l}{76.20$\pm$2.56} & \multicolumn{1}{l}{74.67$\pm$2.73} & \multicolumn{1}{l}{74.41$\pm$3.37} & \multicolumn{1}{l}{72.67$\pm$3.21} & \multicolumn{1}{l}{74.72$\pm$1.98} & 75.28$\pm$2.29 \\
 & 20\% & \multicolumn{1}{l}{74.88$\pm$2.38} & \multicolumn{1}{l}{73.68$\pm$2.36} & \multicolumn{1}{l}{76.09$\pm$3.05} & \multicolumn{1}{l}{74.04$\pm$3.26} & \multicolumn{1}{l}{72.99$\pm$1.83} & \multicolumn{1}{l}{70.43$\pm$2.00} & \multicolumn{1}{l}{72.00$\pm$5.14} & 71.79$\pm$4.01 \\
NRGNN & 25\% & \multicolumn{1}{l}{75.03$\pm$4.14} & \multicolumn{1}{l}{73.52$\pm$1.88} & \multicolumn{1}{l}{75.46$\pm$2.96} & \multicolumn{1}{l}{75.05$\pm$1.90} & \multicolumn{1}{l}{69.89$\pm$7.82} & \multicolumn{1}{l}{64.88$\pm$7.89} & \multicolumn{1}{l}{64.97$\pm$6.98} & 65.47$\pm$7.14 \\
 & 30\% & \multicolumn{1}{l}{75.03$\pm$4.14} & \multicolumn{1}{l}{73.52$\pm$1.88} & \multicolumn{1}{l}{75.46$\pm$2.96} & \multicolumn{1}{l}{73.32$\pm$4.07} & \multicolumn{1}{l}{69.89$\pm$7.82} & \multicolumn{1}{l}{64.88$\pm$7.89} & \multicolumn{1}{l}{65.63$\pm$4.66} & 65.47$\pm$7.14 \\
 & 35\% & \multicolumn{1}{l}{74.44$\pm$2.57} & \multicolumn{1}{l}{71.19$\pm$4.15} & \multicolumn{1}{l}{73.37$\pm$3.82} & \multicolumn{1}{l}{70.88$\pm$6.71} & \multicolumn{1}{l}{67.62$\pm$4.43} & \multicolumn{1}{l}{61.54$\pm$8.91} & \multicolumn{1}{l}{59.04$\pm$6.91} & 62.00$\pm$8.64 \\
 & 40\% & \multicolumn{1}{l}{73.59$\pm$4.67} & \multicolumn{1}{l}{65.95$\pm$9.41} & \multicolumn{1}{l}{73.99$\pm$5.97} & \multicolumn{1}{l}{69.08$\pm$7.87} & \multicolumn{1}{l}{56.46$\pm$11.03} & \multicolumn{1}{l}{53.70$\pm$8.77} & \multicolumn{1}{l}{49.72$\pm$7.41} & 52.93$\pm$5.00 \\
 & 45\% & \multicolumn{1}{l}{71.63$\pm$7.77} & \multicolumn{1}{l}{65.53$\pm$9.80} & \multicolumn{1}{l}{69.96$\pm$7.19} & \multicolumn{1}{l}{70.07$\pm$5.05} & \multicolumn{1}{l}{50.58$\pm$13.34} & \multicolumn{1}{l}{40.34$\pm$9.91} & \multicolumn{1}{l}{40.91$\pm$10.40} & 49.96$\pm$6.07 \\
 & 50\% & \multicolumn{1}{l}{70.50$\pm$9.53} & \multicolumn{1}{l}{65.64$\pm$7.06} & \multicolumn{1}{l}{68.95$\pm$11.43} & \multicolumn{1}{l}{69.70$\pm$8.39} & \multicolumn{1}{l}{47.45$\pm$13.40} & \multicolumn{1}{l}{33.76$\pm$5.54} & \multicolumn{1}{l}{39.15$\pm$12.91} & 45.79$\pm$4.53 \\ \hline
\end{tabular}}
\end{table}

\begin{table}[!h]
\addtocounter{table}{-1}
\caption{ \textbf{Continued:}  Comparison of noise model variants across graph label noise robust algorithms for the Cora dataset. Reported values are accuracy$\pm$std of 10 repetitions.}
\resizebox{\textwidth}{!}{
\begin{tabular}{c|c|llllllll}
 \begin{tabular}[c]{@{}c@{}}Noise\\ Robust \\ Methods\end{tabular} & \multicolumn{1}{c|}{\begin{tabular}[c]{@{}c@{}}Noise \\ Level\end{tabular}} & \multicolumn{1}{l}{SLN\ \ \ \ \ \ \ \ \ \ \ } & \multicolumn{1}{c}{\begin{tabular}[l]{@{}l@{}}MV+ \\ SLN\ \ \ \ \ \ \ \ \ \ \ \end{tabular}} & \multicolumn{1}{c}{\begin{tabular}[l]{@{}l@{}}Veto+\\ SLN \ \ \ \ \ \ \ \ \ \ \end{tabular}} & \multicolumn{1}{c}{\begin{tabular}[l]{@{}l@{}}Seq+\\ SLN\ \ \ \ \ \ \ \ \ \ \end{tabular}} & \multicolumn{1}{l}{PWN\ \ \ \ \ \ \ \ \ \ } & \multicolumn{1}{c}{\begin{tabular}[l]{@{}l@{}}MV+\\ PWN\ \ \ \ \ \ \ \ \ \ \end{tabular}} & \multicolumn{1}{c}{\begin{tabular}[l]{@{}l@{}}Veto+\\ PWN\ \ \ \ \ \ \ \ \ \ \end{tabular}}  &  \multicolumn{1}{c}{\begin{tabular}[l]{@{}l@{}}Seq+\\ PWN\end{tabular}} \\ \hline

  & 5\% & 84.10$\pm$1.86 & 83.99$\pm$1.48 & 84.18$\pm$1.72 & 84.36$\pm$1.53 & 84.26$\pm$1.64 & 84.28$\pm$0.92 & 83.70$\pm$1.58 & 82.16$\pm$4.84 \\
 & 10\% & 83.42$\pm$1.26 & 83.32$\pm$1.46 & 82.02$\pm$3.32 & 82.63$\pm$3.58 & 81.42$\pm$1.75 & 82.75$\pm$1.61 & 80.48$\pm$3.14 & 82.10$\pm$2.27 \\
 & 15\% & 80.21$\pm$7.68 & 83.31$\pm$1.82 & 82.41$\pm$3.57 & 83.16$\pm$3.59 & 80.44$\pm$1.45 & 82.77$\pm$1.33 & 80.61$\pm$3.49 & 82.13$\pm$2.56 \\
 & 20\% & 80.64$\pm$2.24 & 80.63$\pm$1.60 & 81.69$\pm$0.66 & 80.24$\pm$2.57 & 78.28$\pm$2.95 & 77.31$\pm$2.83 & 74.05$\pm$3.75 & 74.63$\pm$6.15 \\
CRGNN & 25\% & 78.25$\pm$1.88 & 76.40$\pm$2.57 & 76.43$\pm$5.53 & 77.58$\pm$2.64 & 75.71$\pm$2.56 & 74.05$\pm$2.41 & 67.48$\pm$4.64 & 69.94$\pm$6.77 \\
 & 30\% & 76.17$\pm$4.27 & 78.09$\pm$3.08 & 76.40$\pm$5.59 & 77.24$\pm$2.92 & 65.10$\pm$6.40 & 72.67$\pm$3.60 & 67.67$\pm$4.50 & 69.53$\pm$6.89 \\
 & 35\% & 72.92$\pm$4.01 & 74.80$\pm$3.62 & 73.15$\pm$4.90 & 74.44$\pm$4.62 & 62.59$\pm$5.91 & 65.96$\pm$3.99 & 61.36$\pm$5.26 & 62.62$\pm$4.83 \\
 & 40\% & 68.39$\pm$6.59 & 61.27$\pm$9.52 & 70.56$\pm$3.67 & 69.63$\pm$6.39 & 55.96$\pm$5.79 & 56.46$\pm$7.09 & 53.60$\pm$5.54 & 51.72$\pm$6.62 \\
 & 45\% & 65.02$\pm$10.42 & 60.58$\pm$12.98 & 62.98$\pm$6.41 & 60.63$\pm$7.09 & 48.32$\pm$9.40 & 49.27$\pm$6.13 & 46.61$\pm$7.83 & 44.22$\pm$7.37 \\
 & 50\% & 59.74$\pm$5.32 & 50.87$\pm$16.45 & 54.58$\pm$9.87 & 51.34$\pm$5.75 & 45.03$\pm$7.34 & 42.24$\pm$6.08 & 40.71$\pm$6.80 & 42.36$\pm$4.73 \\ \hline
 & 5\% & 83.70$\pm$3.67 & 82.43$\pm$4.47 & 83.15$\pm$3.11 & 82.81$\pm$3.26 & 83.57$\pm$2.19 & 83.15$\pm$2.76 & 78.94$\pm$11.48 & 83.27$\pm$2.92 \\
 & 10\% & 81.39$\pm$3.23 & 81.15$\pm$4.09 & 83.31$\pm$1.25 & 83.11$\pm$1.58 & 82.22$\pm$1.56 & 80.24$\pm$3.88 & 76.80$\pm$11.27 & 82.07$\pm$3.84 \\
 & 15\% & 82.06$\pm$2.29 & 81.02$\pm$4.29 & 83.35$\pm$1.32 & 83.13$\pm$1.52 & 78.33$\pm$4.17 & 80.26$\pm$3.83 & 76.75$\pm$11.17 & 82.08$\pm$3.79 \\
 & 20\% & 80.06$\pm$3.79 & 79.84$\pm$2.47 & 79.75$\pm$4.56 & 79.96$\pm$4.54 & 78.17$\pm$2.31 & 75.84$\pm$4.21 & 73.87$\pm$5.14 & 74.95$\pm$10.66 \\
CGNN & 25\% & 79.46$\pm$3.45 & 74.93$\pm$5.98 & 77.06$\pm$4.52 & 76.29$\pm$4.45 & 75.82$\pm$3.25 & 70.34$\pm$9.60 & 70.54$\pm$3.38 & 71.37$\pm$10.88 \\
 & 30\% & 79.22$\pm$3.23 & 71.57$\pm$14.45 & 77.10$\pm$4.56 & 76.24$\pm$4.49 & 69.54$\pm$5.62 & 70.36$\pm$9.61 & 70.58$\pm$3.39 & 71.37$\pm$10.90 \\
 & 35\% & 74.46$\pm$6.30 & 75.87$\pm$3.55 & 73.35$\pm$11.50 & 74.33$\pm$11.82 & 63.24$\pm$4.79 & 67.85$\pm$4.68 & 61.89$\pm$9.57 & 67.36$\pm$10.40 \\
 & 40\% & 73.56$\pm$2.44 & 64.61$\pm$16.26 & 70.72$\pm$13.22 & 70.37$\pm$14.74 & 55.89$\pm$3.07 & 52.08$\pm$11.59 & 50.77$\pm$7.99 & 54.60$\pm$9.32 \\
 & 45\% & 67.76$\pm$6.19 & 63.57$\pm$9.94 & 65.63$\pm$13.02 & 65.15$\pm$13.00 & 51.20$\pm$7.78 & 47.45$\pm$6.24 & 46.39$\pm$8.02 & 47.56$\pm$10.42 \\
 & 50\% & 65.27$\pm$10.08 & 60.78$\pm$10.93 & 61.10$\pm$18.00 & 63.46$\pm$12.93 & 43.75$\pm$7.42 & 41.89$\pm$6.64 & 38.24$\pm$4.82 & 43.29$\pm$11.56 \\ \hline
 & 5\% & 88.79$\pm$2.60 & 88.77$\pm$2.79 & 87.04$\pm$6.67 & 89.41$\pm$1.99 & 88.20$\pm$2.12 & 88.73$\pm$2.32 & 88.86$\pm$2.41 & 84.46$\pm$0.8\\
 & 10\% & 88.18$\pm$2.43 & 84.99$\pm$10.31 & 85.32$\pm$5.52 & 87.67$\pm$3.37 & 87.91$\pm$2.27 & 89.13$\pm$2.21 & 83.20$\pm$10.25 & 84.66$\pm$1.34 \\
 & 15\% & 88.34$\pm$2.08 & 85.05$\pm$10.33 & 85.56$\pm$5.64 & 88.20$\pm$3.00 & 85.68$\pm$5.49 & 89.07$\pm$2.19 & 83.40$\pm$10.35 & 83.68$\pm$1.23 \\
 & 20\% & 88.29$\pm$1.64 & 86.57$\pm$5.97 & 75.16$\pm$24.20 & 83.70$\pm$6.00 & 84.14$\pm$8.42 & 87.63$\pm$1.88 & 81.98$\pm$6.83 & 82.41$\pm$1.33 \\
DeGLIF & 25\% & 87.33$\pm$2.74 & 85.71$\pm$6.37 & 86.17$\pm$3.31 & 85.68$\pm$5.04 & 86.63$\pm$1.84 & 87.60$\pm$2.01 & 73.27$\pm$16.10 & 80.74$\pm$1.15 \\
 & 30\% & 88.11$\pm$1.47 & 85.88$\pm$6.49 & 85.68$\pm$4.32 & 85.47$\pm$4.64 & 82.74$\pm$3.94 & 87.70$\pm$2.09 & 71.72$\pm$15.85 & 80.58$\pm$0.9 \\
 & 35\% & 87.73$\pm$1.93 & 88.66$\pm$0.57 & 82.34$\pm$5.53 & 82.18$\pm$5.70 & 75.29$\pm$6.52 & 83.76$\pm$3.54 & 66.87$\pm$8.56 &  79.1$\pm$1.6\\
 & 40\% & 86.75$\pm$2.57 & 86.07$\pm$1.27 & 73.32$\pm$9.49 & 81.26$\pm$3.58 & 72.46$\pm$6.93 & 75.86$\pm$7.05 & 52.50$\pm$7.85 &  69.52$\pm$3.58\\
 & 45\% & 84.85$\pm$2.21 & 83.60$\pm$2.00 & 75.33$\pm$7.08 & 78.58$\pm$6.30 & 60.11$\pm$3.78 & 63.50$\pm$6.69 & 44.94$\pm$2.33 &  61.2$\pm$6.07\\
 & 50\% & 80.60$\pm$4.75 & 83.35$\pm$1.93 & 67.70$\pm$13.65 & 73.88$\pm$7.90 & 48.40$\pm$7.31 & 46.81$\pm$9.18 & 36.33$\pm$3.81 &  44.16$\pm$5.2\\ \hline

\end{tabular}}
\end{table}

\begin{table}[!h]
\caption{Comparison of noise model variants across graph label noise robust algorithms for the Amazon Photo dataset. Reported values are accuracy$\pm$std of 10 repetitions. }\label{tab:nrl_amazonpho}
\resizebox{\textwidth}{!}{
\begin{tabular}{c|c|llllllll}
 \begin{tabular}[c]{@{}c@{}}Noise\\ Robust \\ Methods\end{tabular} & \multicolumn{1}{c|}{\begin{tabular}[c]{@{}c@{}}Noise \\ Level\end{tabular}} & \multicolumn{1}{l}{SLN\ \ \ \ \ \ \ \ \ \ \ } & \multicolumn{1}{c}{\begin{tabular}[l]{@{}l@{}}MV+ \\ SLN\ \ \ \ \ \ \ \ \ \ \ \end{tabular}} & \multicolumn{1}{c}{\begin{tabular}[l]{@{}l@{}}Veto+\\ SLN \ \ \ \ \ \ \ \ \ \ \end{tabular}} & \multicolumn{1}{c}{\begin{tabular}[l]{@{}l@{}}Seq+\\ SLN\ \ \ \ \ \ \ \ \ \ \end{tabular}} & \multicolumn{1}{l}{PWN\ \ \ \ \ \ \ \ \ \ } & \multicolumn{1}{c}{\begin{tabular}[l]{@{}l@{}}MV+\\ PWN\ \ \ \ \ \ \ \ \ \ \end{tabular}} & \multicolumn{1}{c}{\begin{tabular}[l]{@{}l@{}}Veto+\\ PWN\ \ \ \ \ \ \ \ \ \ \end{tabular}}  &  \multicolumn{1}{c}{\begin{tabular}[l]{@{}l@{}}Seq+\\ PWN\end{tabular}} \\ \hline
 & 5\% & 86.78$\pm$1.43 & 83.65$\pm$6.28 & 85.05$\pm$4.56 & 83.55$\pm$6.18 & 86.66$\pm$1.46 & 84.65$\pm$5.59 & 85.35$\pm$4.21 & 84.85$\pm$2.89 \\
 & 10\% & 86.45$\pm$1.16 & 82.29$\pm$12.38 & 84.14$\pm$3.82 & 83.54$\pm$4.67 & 85.58$\pm$1.32 & 84.38$\pm$5.67 & 83.60$\pm$3.27 & 80.91$\pm$5.22 \\
 & 15\% & 85.49$\pm$1.21 & 82.30$\pm$12.38 & 84.16$\pm$3.85 & 83.55$\pm$4.67 & 83.53$\pm$1.19 & 84.39$\pm$5.69 & 83.58$\pm$3.26 & 80.92$\pm$5.18 \\
GCN & 20\% & 82.83$\pm$3.66 & 82.86$\pm$6.06 & 83.05$\pm$3.65 & 80.82$\pm$5.00 & 79.72$\pm$4.84 & 77.60$\pm$10.34 & 77.68$\pm$6.76 & 72.0$\pm$6.66 \\
 & 25\% & 83.96$\pm$2.99 & 78.28$\pm$9.38 & 81.87$\pm$4.08 & 80.58$\pm$7.92 & 77.00$\pm$5.47 & 74.15$\pm$9.42 & 70.81$\pm$7.69 & 66.82$\pm$8.03 \\
 & 30\% & 81.15$\pm$9.72 & 78.26$\pm$9.39 & 81.88$\pm$4.03 & 80.66$\pm$7.91 & 70.94$\pm$12.13 & 74.16$\pm$9.41 & 70.82$\pm$7.69 & 66.83$\pm$8.05 \\
 & 35\% & 83.01$\pm$3.82 & 76.80$\pm$11.90 & 77.65$\pm$7.65 & 78.01$\pm$9.09 & 62.77$\pm$7.32 & 69.21$\pm$9.21 & 63.43$\pm$4.89 & 61.85$\pm$12.71 \\
 & 40\% & 78.75$\pm$4.02 & 76.00$\pm$10.41 & 71.06$\pm$12.34 & 70.89$\pm$9.41 & 54.20$\pm$7.31 & 56.45$\pm$8.28 & 49.66$\pm$3.14 & 45.15$\pm$6.21 \\
 & 45\% & 73.45$\pm$5.67 & 75.36$\pm$8.82 & 66.92$\pm$11.98 & 67.85$\pm$11.36 & 44.15$\pm$4.85 & 49.57$\pm$7.92 & 40.74$\pm$2.65 & 40.03$\pm$5.61 \\
 & 50\% & 72.71$\pm$4.76 & 71.59$\pm$6.88 & 64.10$\pm$8.61 & 66.44$\pm$7.91 & 36.19$\pm$4.19 & 42.51$\pm$7.34 & 33.15$\pm$3.93 & 33.98$\pm$6.6 \\ \hline
 & 5\% & 78.32$\pm$4.81 & 82.68$\pm$2.81 & 79.80$\pm$1.65 & 78.08$\pm$4.58 & 74.66$\pm$9.15 & 82.50$\pm$0.95 & 78.38$\pm$7.58 & 61.44$\pm$27.08 \\
 & 10\% & 78.80$\pm$5.07 & 81.06$\pm$2.48 & 68.90$\pm$11.77 & 70.38$\pm$9.41 & 76.74$\pm$2.93 & 82.14$\pm$0.84 & 70.94$\pm$8.00 & 58.92$\pm$23.53 \\
 & 15\% & 72.60$\pm$12.99 & 81.04$\pm$2.57 & 67.34$\pm$13.42 & 70.28$\pm$9.06 & 72.40$\pm$4.04 & 82.22$\pm$0.91 & 74.00$\pm$7.08 & 59.48$\pm$21.26 \\
 & 20\% & 77.08$\pm$4.10 & 80.66$\pm$2.09 & 70.38$\pm$9.93 & 67.88$\pm$8.07 & 67.66$\pm$8.22 & 78.82$\pm$2.40 & 63.24$\pm$10.91 & 49.32$\pm$20.89 \\
DGNN & 25\% & 74.90$\pm$5.87 & 79.38$\pm$2.39 & 65.20$\pm$11.22 & 65.38$\pm$13.62 & 61.78$\pm$9.81 & 77.36$\pm$3.91 & 53.28$\pm$12.29 & 55.50$\pm$5.54 \\
 & 30\% & 60.32$\pm$21.55 & 79.54$\pm$2.03 & 64.58$\pm$11.17 & 64.74$\pm$14.17 & 61.82$\pm$11.15 & 77.14$\pm$3.63 & 48.86$\pm$11.79 & 47.58$\pm$17.63 \\
 & 35\% & 59.96$\pm$16.53 & 74.50$\pm$8.68 & 55.26$\pm$18.68 & 61.84$\pm$8.26 & 53.28$\pm$10.14 & 77.08$\pm$4.68 & 49.30$\pm$17.08 & 43.64$\pm$12.16 \\
 & 40\% & 65.90$\pm$15.86 & 70.68$\pm$7.12 & 54.36$\pm$9.28 & 52.16$\pm$14.51 & 45.02$\pm$10.60 & 58.94$\pm$9.51 & 38.84$\pm$8.29 & 36.16$\pm$10.11 \\
 & 45\% & 62.34$\pm$12.99 & 60.08$\pm$15.94 & 40.26$\pm$8.56 & 40.10$\pm$8.99 & 46.64$\pm$11.84 & 45.02$\pm$11.65 & 30.94$\pm$6.26 & 35.96$\pm$4.80 \\
 & 50\% & 64.62$\pm$8.05 & 56.08$\pm$15.76 & 38.42$\pm$9.73 & 40.52$\pm$9.75 & 33.98$\pm$7.06 & 33.96$\pm$5.16 & 21.48$\pm$5.34 & 30.84$\pm$4.52 \\ \hline
 & 5\% & 88.9$\pm$0.4 & 89.56$\pm$0.58 & 90.72$\pm$0.31 & 92.42$\pm$0.59 & 89.74$\pm$1.26 & 88.02$\pm$0.96 & 89.96$\pm$0.58 & 90.06$\pm$0.88 \\
 & 10\% & 90.1$\pm$1.7 & 89.76$\pm$0.57 & 90.80$\pm$0.53 & 90.56$\pm$0.36 & 90.28$\pm$0.33 & 88.60$\pm$1.21 & 88.16$\pm$2.47 & 89.10$\pm$1.80 \\
 & 15\% & 91$\pm$1.3 & 89.66$\pm$0.57 & 90.70$\pm$0.55 & 90.70$\pm$0.60 & 89.94$\pm$0.76 & 88.72$\pm$1.25 & 88.24$\pm$2.08 & 90.00$\pm$1.31 \\
 & 20\% & 88.1$\pm$1.8 & 89.28$\pm$0.77 & 89.88$\pm$0.47 & 89.92$\pm$0.77 & 86.96$\pm$2.16 & 87.62$\pm$2.28 & 81.84$\pm$2.14 & 86.22$\pm$2.34 \\
PIGNN & 25\% & 86.8$\pm$3.4 & 88.96$\pm$1.72 & 90.30$\pm$0.65 & 90.52$\pm$0.50 & 85.22$\pm$2.21 & 87.52$\pm$2.15 & 74.56$\pm$2.17 & 82.10$\pm$2.06 \\
 & 30\% & 87.6$\pm$0.7 & 89.28$\pm$1.50 & 90.00$\pm$0.87 & 89.80$\pm$1.33 & 84.74$\pm$2.88 & 87.92$\pm$1.78 & 74.46$\pm$2.84 & 82.08$\pm$2.14 \\
 & 35\% & 86.3$\pm$1.8 & 88.50$\pm$1.94 & 89.44$\pm$1.07 & 88.26$\pm$1.61 & 80.28$\pm$2.50 & 85.74$\pm$3.42 & 68.84$\pm$2.21 & 80.32$\pm$3.22 \\
 & 40\% & 82.5$\pm$5.2 & 88.24$\pm$1.05 & 81.78$\pm$4.91 & 85.44$\pm$2.54 & 72.60$\pm$4.82 & 76.26$\pm$7.91 & 53.78$\pm$4.28 & 71.98$\pm$4.49 \\
 & 45\% & 82.2$\pm$4.2 & 87.22$\pm$2.37 & 79.86$\pm$7.22 & 80.64$\pm$3.36 & 60.44$\pm$8.51 & 62.38$\pm$9.04 & 42.86$\pm$6.44 & 63.52$\pm$2.81 \\
 & 50\% & 76.9$\pm$4.3 & 87.22$\pm$2.11 & 75.68$\pm$3.31 & 78.70$\pm$4.57 & 49.82$\pm$8.58 & 50.70$\pm$9.25 & 34.38$\pm$4.72 & 56.26$\pm$1.92 \\ \hline
 & 5\% & 83.82$\pm$5.00 & 84.06$\pm$4.05 & 85.32$\pm$2.00 & 87.04$\pm$2.01 & 83.78$\pm$2.71 & 82.08$\pm$3.42 & 88.32$\pm$2.02 & 84.48$\pm$3.53 \\
 & 10\% & 85.52$\pm$1.86 & 83.14$\pm$4.53 & 80.24$\pm$5.90 & 83.90$\pm$1.57 & 81.86$\pm$4.17 & 84.46$\pm$4.02 & 83.10$\pm$2.46 & 83.54$\pm$4.51 \\
 & 15\% & 84.50$\pm$1.95 & 83.14$\pm$4.53 & 80.24$\pm$5.90 & 83.90$\pm$1.57 & 82.04$\pm$1.87 & 84.46$\pm$4.02 & 83.10$\pm$2.46 & 83.54$\pm$4.51 \\
 & 20\% & 81.24$\pm$3.62 & 85.02$\pm$0.80 & 77.46$\pm$7.44 & 76.52$\pm$4.67 & 75.88$\pm$5.39 & 83.08$\pm$5.55 & 75.50$\pm$7.64 & 75.76$\pm$7.18 \\
RNCGLN & 25\% & 80.26$\pm$5.05 & 80.24$\pm$3.99 & 71.54$\pm$6.55 & 70.50$\pm$4.53 & 71.94$\pm$5.37 & 77.00$\pm$9.12 & 64.98$\pm$3.93 & 69.96$\pm$6.36 \\
 & 30\% & 79.46$\pm$4.74 & 80.24$\pm$3.99 & 71.54$\pm$6.55 & 70.50$\pm$4.53 & 70.36$\pm$10.52 & 77.00$\pm$9.12 & 64.98$\pm$3.93 & 69.96$\pm$6.36 \\
 & 35\% & 64.16$\pm$6.37 & 70.26$\pm$4.10 & 68.54$\pm$2.74 & 70.34$\pm$8.76 & 66.32$\pm$8.12 & 70.04$\pm$11.88 & 59.96$\pm$4.31 & 67.94$\pm$9.42 \\
 & 40\% & 67.42$\pm$9.70 & 67.14$\pm$7.35 & 55.20$\pm$5.68 & 62.42$\pm$9.45 & 58.42$\pm$7.35 & 60.54$\pm$7.37 & 49.44$\pm$4.91 & 57.28$\pm$8.30 \\
 & 45\% & 56.52$\pm$5.79 & 61.22$\pm$6.57 & 51.68$\pm$5.66 & 55.40$\pm$5.21 & 52.40$\pm$8.32 & 63.90$\pm$8.12 & 42.58$\pm$2.93 & 48.34$\pm$1.06 \\
 & 50\% & 60.66$\pm$14.26 & 53.28$\pm$4.08 & 53.68$\pm$12.53 & 45.14$\pm$1.67 & 44.48$\pm$5.88 & 45.48$\pm$4.18 & 38.28$\pm$5.98 & 45.32$\pm$2.65 \\ \hline
 & 5\% & 80.8$\pm$5.3 & 81.93$\pm$2.17 & 81.46$\pm$2.44 & 84.14$\pm$2.19 & 82.24$\pm$0.86 & 83.45$\pm$1.10 & 81.95$\pm$1.42 & 82.04$\pm$1.53 \\
 & 10\% & 82.2$\pm$5.3 & 80.80$\pm$1.06 & 83.20$\pm$1.55 & 83.12$\pm$2.01 & 83.09$\pm$0.93 & 82.03$\pm$2.15 & 82.12$\pm$1.49 & 82.85$\pm$2.32 \\
 & 15\% & 83.6$\pm$2.7 & 81.29$\pm$1.87 & 82.16$\pm$2.43 & 82.89$\pm$1.75 & 82.18$\pm$1.02 & 81.81$\pm$2.07 & 81.75$\pm$1.43 & 81.96$\pm$1.42 \\
 & 20\% & 84$\pm$2.2 & 83.11$\pm$3.04 & 82.52$\pm$2.48 & 82.67$\pm$2.16 & 83.85$\pm$1.35 & 84.50$\pm$3.96 & 77.35$\pm$6.71 & 78.33$\pm$8.36 \\
RTGNN & 25\% & 82.9$\pm$4.9 & 82.63$\pm$3.85 & 82.93$\pm$3.32 & 81.96$\pm$2.17 & 83.17$\pm$3.79 & 84.88$\pm$3.15 & 69.78$\pm$3.62 & 76.19$\pm$6.53 \\
 & 30\% & 78.8$\pm$6.4 & 82.69$\pm$4.25 & 82.88$\pm$3.83 & 82.25$\pm$1.89 & 81.62$\pm$3.80 & 85.79$\pm$2.88 & 70.53$\pm$3.80 & 76.03$\pm$6.71 \\
 & 35\% & 79.3$\pm$5.4 & 83.56$\pm$4.32 & 81.93$\pm$2.37 & 82.83$\pm$3.04 & 78.67$\pm$5.08 & 81.04$\pm$9.22 & 64.12$\pm$6.47 & 73.21$\pm$4.48 \\
 & 40\% & 83.5$\pm$4.7 & 83.23$\pm$3.07 & 74.40$\pm$8.00 & 80.99$\pm$3.04 & 67.18$\pm$8.84 & 70.42$\pm$11.24 & 51.86$\pm$6.85 & 70.48$\pm$4.20 \\
 & 45\% & 86$\pm$1.3 & 84.98$\pm$1.62 & 75.71$\pm$7.21 & 74.46$\pm$5.80 & 60.86$\pm$8.45 & 58.03$\pm$12.32 & 47.17$\pm$8.88 & 60.56$\pm$3.57 \\
 & 50\% & 79.9$\pm$5 & 85.05$\pm$1.16 & 68.25$\pm$10.02 & 68.62$\pm$5.65 & 51.60$\pm$10.05 & 47.89$\pm$14.48 & 42.44$\pm$5.21 & 55.36$\pm$3.04 \\ \hline
 & 5\% & 69$\pm$8 & 87.52$\pm$0.90 & 87.34$\pm$1.68 & 88.40$\pm$2.77 & 89.74$\pm$1.26 & 87.08$\pm$1.67 & 86.56$\pm$1.54 & 85.90$\pm$2.81 \\
 & 10\% & 68.1$\pm$6.8 & 86.58$\pm$1.88 & 88.02$\pm$1.40 & 87.88$\pm$1.22 & 90.28$\pm$0.33 & 87.26$\pm$1.83 & 86.04$\pm$2.95 & 85.00$\pm$2.58 \\
 & 15\% & 72.4$\pm$5.5 & 86.34$\pm$2.17 & 87.46$\pm$1.84 & 87.14$\pm$0.90 & 89.94$\pm$0.76 & 87.26$\pm$1.94 & 85.88$\pm$2.69 & 84.62$\pm$3.32 \\
 & 20\% & 66.5$\pm$5.1 & 86.78$\pm$2.29 & 87.44$\pm$0.86 & 86.20$\pm$1.92 & 86.96$\pm$2.16 & 87.76$\pm$1.25 & 78.96$\pm$4.57 & 82.34$\pm$5.03 \\
NRGNN & 25\% & 55.1$\pm$5.4 & 86.92$\pm$2.84 & 85.92$\pm$0.64 & 86.08$\pm$0.90 & 85.22$\pm$2.21 & 85.52$\pm$2.32 & 72.50$\pm$4.41 & 81.24$\pm$5.63 \\
 & 30\% & 60$\pm$7.1 & 86.90$\pm$2.73 & 86.00$\pm$0.76 & 85.94$\pm$0.40 & 84.74$\pm$2.88 & 85.22$\pm$2.14 & 72.32$\pm$4.41 & 81.00$\pm$5.43 \\
 & 35\% & 58.3$\pm$6 & 85.40$\pm$4.60 & 85.44$\pm$1.46 & 84.72$\pm$2.12 & 80.28$\pm$2.50 & 84.34$\pm$3.26 & 67.28$\pm$3.30 & 78.48$\pm$7.20 \\
 & 40\% & 60$\pm$5.1 & 85.44$\pm$4.15 & 79.38$\pm$3.36 & 81.32$\pm$2.96 & 72.60$\pm$4.82 & 75.40$\pm$6.37 & 55.44$\pm$4.89 & 69.90$\pm$6.54 \\
 & 45\% & 54.5$\pm$6.2 & 86.70$\pm$1.72 & 73.78$\pm$4.98 & 81.42$\pm$2.51 & 60.44$\pm$8.51 & 66.20$\pm$8.31 & 49.90$\pm$3.87 & 58.78$\pm$1.89 \\
 & 50\% & 51.5$\pm$5.9 & 85.92$\pm$0.87 & 69.54$\pm$6.52 & 71.34$\pm$7.86 & 49.82$\pm$8.58 & 51.24$\pm$7.32 & 41.64$\pm$4.98 & 56.56$\pm$5.51 \\ \hline
 
\end{tabular}}
\end{table}

\begin{table}[!h]
\addtocounter{table}{-1}
\caption{ \textbf{Continued:} Comparison of noise model variants across graph label noise robust algorithms for the Amazon Photo dataset. Reported values are accuracy$\pm$std of 10 repetitions.}
\resizebox{\textwidth}{!}{
\begin{tabular}{c|c|llllllll}
 \begin{tabular}[c]{@{}c@{}}Noise\\ Robust \\ Methods\end{tabular} & \multicolumn{1}{c|}{\begin{tabular}[c]{@{}c@{}}Noise \\ Level\end{tabular}} & \multicolumn{1}{l}{SLN\ \ \ \  \ \ \ \ \ \ \ \ } & \multicolumn{1}{c}{\begin{tabular}[l]{@{}l@{}}MV+ \\ SLN\ \ \ \ \ \ \ \ \ \ \ \ \end{tabular}} & \multicolumn{1}{c}{\begin{tabular}[l]{@{}l@{}}Veto+\\ SLN \ \ \ \ \ \ \ \ \ \ \ \end{tabular}} & \multicolumn{1}{c}{\begin{tabular}[l]{@{}l@{}}Seq+\\ SLN\ \ \ \ \ \ \ \ \ \ \ \ \end{tabular}} & \multicolumn{1}{l}{PWN\ \ \ \ \ \ \ \ \ \ \ } & \multicolumn{1}{c}{\begin{tabular}[l]{@{}l@{}}MV+\\ PWN\ \ \ \ \ \ \ \ \ \ \end{tabular}} & \multicolumn{1}{c}{\begin{tabular}[l]{@{}l@{}}Veto+\\ PWN\ \ \ \ \ \ \ \ \ \ \end{tabular}}  &  \multicolumn{1}{c}{\begin{tabular}[l]{@{}l@{}}Seq+\\ PWN\end{tabular}} \\ \hline
& 5\% & 59.04$\pm$12.73 & 44.60$\pm$12.81 & 54.28$\pm$14.68 & 54.52$\pm$14.41 & 54.80$\pm$12.30 & 52.82$\pm$12.77 & 47.36$\pm$15.24 & 37.54$\pm$46.21 \\
 & 10\% & 47.82$\pm$7.18 & 49.66$\pm$8.69 & 52.98$\pm$20.93 & 56.80$\pm$15.44 & 49.50$\pm$22.16 & 48.52$\pm$16.00 & 42.60$\pm$10.71 & 37.70$\pm$46.43 \\
 & 15\% & 45.18$\pm$11.56 & 50.88$\pm$10.12 & 52.66$\pm$19.01 & 55.70$\pm$17.17 & 46.32$\pm$16.69 & 50.60$\pm$14.56 & 43.18$\pm$8.97 & 36.84$\pm$45.31 \\
 & 20\% & 28.26$\pm$16.21 & 36.92$\pm$11.71 & 41.66$\pm$13.60 & 30.96$\pm$15.38 & 51.22$\pm$18.45 & 49.02$\pm$12.78 & 46.08$\pm$13.86 & 33.22$\pm$40.32 \\
CRGNN & 25\% & 45.02$\pm$10.86 & 37.82$\pm$11.13 & 35.76$\pm$11.64 & 35.22$\pm$14.65 & 47.68$\pm$20.70 & 31.22$\pm$7.03 & 41.32$\pm$7.76 & 32.74$\pm$39.74 \\
 & 30\% & 23.20$\pm$10.72 & 34.80$\pm$13.36 & 34.20$\pm$8.65 & 34.04$\pm$14.97 & 35.78$\pm$10.66 & 37.42$\pm$9.29 & 39.16$\pm$4.63 & 33.24$\pm$40.32 \\
 & 35\% & 25.78$\pm$10.25 & 34.08$\pm$12.27 & 29.32$\pm$5.16 & 27.10$\pm$5.88 & 30.62$\pm$13.45 & 34.92$\pm$5.10 & 36.94$\pm$7.39 & 26.36$\pm$31.04 \\
 & 40\% & 30.56$\pm$4.71 & 35.72$\pm$10.82 & 29.00$\pm$8.54 & 22.84$\pm$6.12 & 34.36$\pm$13.35 & 36.58$\pm$13.16 & 30.90$\pm$9.06 & 24.52$\pm$28.75 \\
 & 45\% & 29.48$\pm$15.05 & 31.20$\pm$5.21 & 23.12$\pm$5.72 & 18.82$\pm$4.97 & 33.14$\pm$13.04 & 37.74$\pm$8.31 & 27.42$\pm$4.28 & 19.78$\pm$21.92 \\
 & 50\% & 22.32$\pm$7.22 & 23.46$\pm$11.85 & 18.96$\pm$9.50 & 22.50$\pm$6.74 & 22.00$\pm$4.49 & 30.28$\pm$6.45 & 27.88$\pm$3.31 & 20.48$\pm$22.98 \\ \hline
 & 5\% & 39.08$\pm$28.12 & 28.70$\pm$12.19 & 23.32$\pm$6.22 & 33.60$\pm$21.41 & 33.82$\pm$19.18 & 25.10$\pm$7.71 & 32.08$\pm$11.53 & 56.84$\pm$25.81 \\
 & 10\% & 34.84$\pm$21.85 & 21.94$\pm$9.72 & 26.16$\pm$9.72 & 27.18$\pm$22.91 & 36.48$\pm$14.53 & 22.44$\pm$7.35 & 26.90$\pm$9.63 & 59.74$\pm$18.83 \\
 & 15\% & 31.26$\pm$24.51 & 19.46$\pm$10.57 & 26.30$\pm$10.02 & 27.52$\pm$22.67 & 41.60$\pm$23.20 & 22.44$\pm$7.34 & 26.66$\pm$9.52 & 60.02$\pm$19.02 \\
 & 20\% & 35.18$\pm$19.95 & 27.46$\pm$8.12 & 27.58$\pm$14.10 & 29.90$\pm$21.35 & 29.64$\pm$14.72 & 19.42$\pm$6.97 & 31.82$\pm$10.28 & 59.46$\pm$25.09 \\
CGNN & 25\% & 34.08$\pm$23.35 & 22.70$\pm$8.20 & 24.30$\pm$5.93 & 25.88$\pm$24.09 & 31.26$\pm$18.21 & 22.48$\pm$3.59 & 29.08$\pm$7.08 & 47.86$\pm$14.05 \\
 & 30\% & 31.98$\pm$14.61 & 22.32$\pm$8.40 & 24.38$\pm$5.96 & 25.72$\pm$23.70 & 29.58$\pm$16.67 & 22.38$\pm$3.70 & 29.10$\pm$7.20 & 47.50$\pm$13.35 \\
 & 35\% & 22.42$\pm$10.69 & 16.68$\pm$7.75 & 28.20$\pm$16.37 & 28.30$\pm$23.33 & 40.74$\pm$20.02 & 21.70$\pm$8.10 & 25.56$\pm$5.51 & 50.58$\pm$12.89 \\
 & 40\% & 28.72$\pm$11.36 & 18.12$\pm$10.43 & 24.04$\pm$9.91 & 20.62$\pm$22.20 & 22.00$\pm$16.94 & 24.02$\pm$7.07 & 27.00$\pm$4.35 & 43.52$\pm$19.07 \\
 & 45\% & 16.62$\pm$9.35 & 17.40$\pm$9.27 & 20.84$\pm$7.90 & 22.96$\pm$11.08 & 22.74$\pm$12.74 & 24.44$\pm$7.80 & 25.78$\pm$7.39 & 36.52$\pm$9.30 \\
 & 50\% & 25.10$\pm$9.55 & 20.32$\pm$8.24 & 23.74$\pm$5.31 & 24.00$\pm$8.78 & 25.22$\pm$12.10 & 19.60$\pm$8.02 & 24.78$\pm$8.12 & 37.12$\pm$6.93 \\ \hline
 & 5\% & 88.79$\pm$2.60 & 88.77$\pm$2.79 & 87.04$\pm$6.67 & 89.41$\pm$1.99 & 88.20$\pm$2.12 & 88.73$\pm$2.32 & 88.86$\pm$2.41 &89.09 $\pm$2.13 \\
 & 10\% & 88.18$\pm$2.43 & 84.99$\pm$10.31 & 85.32$\pm$5.52 & 87.67$\pm$3.37 & 87.91$\pm$2.27 & 89.13$\pm$2.21 & 83.20$\pm$10.25 & 87.11$\pm$2.74 \\
 & 15\% & 88.34$\pm$2.08 & 85.05$\pm$10.33 & 85.56$\pm$5.64 & 88.20$\pm$3.00 & 85.68$\pm$5.49 & 89.07$\pm$2.19 & 83.40$\pm$10.35 & 87.40$\pm$2.41  \\
 & 20\% & 88.29$\pm$1.64 & 86.57$\pm$5.97 & 75.16$\pm$24.20 & 83.70$\pm$6.00 & 84.14$\pm$8.42 & 87.63$\pm$1.88 & 81.98$\pm$6.83 & 82.98$\pm$2.91 \\
DeGLIF & 25\% & 87.33$\pm$2.74 & 85.71$\pm$6.37 & 86.17$\pm$3.31 & 85.68$\pm$5.04 & 86.63$\pm$1.84 & 87.60$\pm$2.01 & 73.27$\pm$16.10 & 79.08$\pm$4.45 \\
 & 30\% & 88.11$\pm$1.47 & 85.88$\pm$6.49 & 85.68$\pm$4.32 & 85.47$\pm$4.64 & 82.74$\pm$3.94 & 87.70$\pm$2.09 & 71.72$\pm$15.85 & 79.25$\pm$4.25  \\
 & 35\% & 87.73$\pm$1.93 & 88.66$\pm$0.57 & 82.34$\pm$5.53 & 82.18$\pm$5.70 & 75.29$\pm$6.52 & 83.76$\pm$3.54 & 66.87$\pm$8.56 & 77.09$\pm$4.9 \\
 & 40\% & 86.75$\pm$2.57 & 86.07$\pm$1.27 & 73.32$\pm$9.49 & 81.26$\pm$3.58 & 72.46$\pm$6.93 & 75.86$\pm$7.05 & 52.50$\pm$7.85 & 61.99$\pm$10.1 \\
 & 45\% & 84.85$\pm$2.21 & 83.60$\pm$2.00 & 75.33$\pm$7.08 & 78.58$\pm$6.30 & 60.11$\pm$3.78 & 63.50$\pm$6.69 & 44.94$\pm$2.33 & 52.92$\pm$11.73 \\
 & 50\% & 80.60$\pm$4.75 & 83.35$\pm$1.93 & 67.70$\pm$13.65 & 73.88$\pm$7.90 & 48.40$\pm$7.31 & 46.81$\pm$9.18 & 36.33$\pm$3.81 & 40.52$\pm$9.96 \\ \hline

\end{tabular}}
\end{table}


\begin{thebibliography}{88}

\bibitem{Xiao2021GraphNN}
Shunxin Xiao, Shiping Wang, Yuanfei Dai, and Wenzhong Guo.
\newblock Graph neural networks in node classification: survey and evaluation.
\newblock {\em Machine Vision and Applications}, 33, 2021.

\bibitem{Zhou2018GraphNN}
Jie Zhou, Ganqu Cui, Z.~Zhang, Cheng Yang, Zhiyuan Liu, and Maosong Sun.
\newblock Graph neural networks: A review of methods and applications.
\newblock {\em AI open}, 2020.

\bibitem{Ju2024ASO}
Wei Ju, Siyu Yi, Yifan Wang, Zhiping Xiao, Zhengyan Mao, Hourun Li, Yiyang Gu, Yifang Qin, Nan Yin, Senzhang Wang, Xinwang Liu, Xiao Luo, Philip~S. Yu, and Ming Zhang.
\newblock A survey of graph neural networks in real world: Imbalance, noise, privacy and ood challenges.
\newblock {\em ArXiv}, abs/2403.04468, 2024.

\bibitem{tut_nh}
Sandhya Tripathi and N.~Hemachandra.
\newblock Label noise: Problems and solutions.
\newblock {\em Tutorial at IEEE DSAA}, 2020.

\bibitem{Yuan2023LearningOGcgnn}
Jingyang Yuan, Xiao Luo, Yifang Qin, Yusheng Zhao, Wei Ju, and Ming Zhang.
\newblock Learning on graphs under label noise.
\newblock {\em ICASSP}, 2023.

\bibitem{NT2019LearningGN}
Hoang NT, Choong~Jun Jin, and Tsuyoshi Murata.
\newblock Learning graph neural networks with noisy labels.
\newblock {\em arXiv preprint arXiv:1905.01591}, 2019.

\bibitem{Du2021NoiserobustGL}
Xuefeng Du, Tian Bian, Yu~Rong, Bo~Han, Tongliang Liu, Tingyang Xu, Wenbing Huang, Yixuan Li, and Junzhou Huang.
\newblock Noise-robust graph learning by estimating and leveraging pairwise interactions.
\newblock {\em Trans. Mach. Learn. Res.}, 2021.

\bibitem{Zhu2024RobustNC}
Yonghua Zhu, Lei Feng, Zhenyun Deng, Yang Chen, Robert Amor, and Michael Witbrock.
\newblock Robust node classification on graph data with graph and label noise.
\newblock In {\em AAAI Conference on Artificial Intelligence}, 2024.

\bibitem{dai2021nrgnn}
Enyan Dai, Charu Aggarwal, and Suhang Wang.
\newblock Nrgnn: Learning a label noise resistant graph neural network on sparsely and noisily labeled graphs.
\newblock In {\em Proceedings of the 27th ACM SIGKDD}, pages 227--236, 2021.

\bibitem{qian2023robust}
Siyi Qian, Haochao Ying, Renjun Hu, Jingbo Zhou, Jintai Chen, Danny~Ziyi Chen, and Jian Wu.
\newblock Robust training of graph neural networks via noise governance.
\newblock {\em ACM International Conference on Web Search and Data Mining}, 2023.

\bibitem{Li2024ContrastiveLOcrgnn}
Xianxian Li, Qiyu Li, De~Li, Haodong Qian, and Jinyan Wang.
\newblock Contrastive learning of graphs under label noise.
\newblock {\em Neural networks}, 2024.

\bibitem{Norris_1997}
James~R. Norris.
\newblock {\em Markov Chains}.
\newblock Cambridge University Press, 1997.

\bibitem{wang2024noisygl}
Zhonghao Wang, Danyu Sun, Sheng Zhou, Haobo Wang, Jiapei Fan, Longtao Huang, and Jiajun Bu.
\newblock Noisygl: A comprehensive benchmark for graph neural networks under label noise.
\newblock {\em NeurIPS Dataset Benchmark}, 2024.

\bibitem{DeGLIf}
Pintu Kumar and Nandyala Hemachandra.
\newblock {DeGLIF} for {L}abel {N}oise {R}obust {N}ode {C}lassification using {GNNs}.
\newblock {\em arXiv}, abs/2506.00244, 2025.

\bibitem{Chen2024ADEdgeDropAE}
Zhaoliang Chen, Zhihao Wu, Ylli Sadikaj, Claudia Plant, Hong-Ning Dai, Shiping Wang, and Wenzhong Guo.
\newblock Adedgedrop: Adversarial edge dropping for robust graph neural networks.
\newblock {\em ArXiv}, abs/2403.09171, 2024.

\bibitem{Rong2019DropEdgeTD}
Yu~Rong, Wenbing Huang, Tingyang Xu, and Junzhou Huang.
\newblock Dropedge: Towards deep graph convolutional networks on node classification.
\newblock In {\em International Conference on Learning Representations}, 2019.

\bibitem{Liu2019AUF}
Xuanqing Liu, Si~Si, Xiaojin Zhu, Yang Li, and Cho-Jui Hsieh.
\newblock A unified framework for data poisoning attack to graph-based semi-supervised learning.
\newblock In {\em Neural Information Processing Systems}, 2019.

\bibitem{Zhang2020AdversarialLA}
Mengmei Zhang, Linmei Hu, Chuan Shi, and Xiao Wang.
\newblock Adversarial label-flipping attack and defense for graph neural networks.
\newblock {\em 2020 IEEE International Conference on Data Mining (ICDM)}, 2020.

\bibitem{strang2000linear}
Gilbert Strang.
\newblock Linear algebra and its applications, 2000.

\bibitem{CircDavis}
P.J. Davis.
\newblock {\em Circulant Matrices}.
\newblock Monographs and textbooks in pure and applied mathematics. Wiley, 1979.

\bibitem{Yang2016RevisitingSL}
Zhilin Yang, William~W. Cohen, and Ruslan Salakhutdinov.
\newblock Revisiting semi-supervised learning with graph embeddings.
\newblock {\em ArXiv}, abs/1603.08861, 2016.

\bibitem{shchur2018pitfalls}
Oleksandr Shchur, Maximilian Mumme, Aleksandar Bojchevski, and Stephan G{\"u}nnemann.
\newblock Pitfalls of graph neural network evaluation.
\newblock {\em arXiv preprint arXiv:1811.05868}, 2018.

\bibitem{kipf2016semi}
Thomas~N Kipf and Max Welling.
\newblock Semi-supervised classification with graph convolutional networks.
\newblock {\em arXiv preprint arXiv:1609.02907}, 2016.

\bibitem{ginXu2018HowPA}
Keyulu Xu, Weihua Hu, Jure Leskovec, and Stefanie Jegelka.
\newblock How powerful are graph neural networks?
\newblock {\em ArXiv}, abs/1810.00826, 2018.

\bibitem{sage}
William~L. Hamilton, Zhitao Ying, and Jure Leskovec.
\newblock Inductive representation learning on large graphs.
\newblock In {\em NIPS}, 2017.

\bibitem{gatVelickovic2017GraphAN}
Petar Velickovic, Guillem Cucurull, Arantxa Casanova, Adriana Romero, Pietro Lio’, and Yoshua Bengio.
\newblock Graph attention networks.
\newblock {\em ArXiv}, abs/1710.10903, 2017.

\bibitem{transforShi2020MaskedLP}
Yunsheng Shi, Zhengjie Huang, Wenjin Wang, Hui Zhong, Shikun Feng, and Yu~Sun.
\newblock Masked label prediction: Unified massage passing model for semi-supervised classification.
\newblock {\em IJCAI}, 2021.

\bibitem{ross2020introduction}
Sheldon~M. Ross.
\newblock {\em Introduction to Probability and Statistics for Engineers and Scientists}.
\newblock Academic Press, 6th edition, 2020.

\bibitem{casella2024statistical}
George Casella and Roger~L. Berger.
\newblock {\em Statistical Inference}.
\newblock Taylor \& Francis, 2nd edition, 2024.

\end{thebibliography}
\end{document}